\PassOptionsToPackage{table}{xcolor}
\documentclass[10pt,twocolumn,letterpaper]{article}
\usepackage[pagenumbers]{cvpr} %

\newcommand{\mypar}[1]{\noindent\textbf{#1}}

\usepackage{duckuments}
\usepackage{comment}
\usepackage{xspace}

\usepackage{tabularx}
\usepackage{adjustbox}
\usepackage{multicol}
\usepackage{multirow}
\usepackage{caption}

\usepackage{subcaption}

\newcommand{\classifier}{\phi}
\newcommand{\Classifier}{\Phi}

\newcommand{\paramsSp}{\omega}
\newcommand{\paramsSpT}{\paramsSp_t}
\newcommand{\paramsSpV}{\paramsSp_v}

\newcommand{\paramsSpLN}{\omega_\mathtt{LN}}
\newcommand{\paramsSpLNOptim}{\omega_\mathtt{LN}^*}
\newcommand{\paramsSpLNT}{\paramsSpLN^t}
\newcommand{\paramsSpLNV}{\paramsSpLN^v}
\newcommand{\paramsSpLNTOptim}{\paramsSpLN^{*^t}}
\newcommand{\paramsSpLNVOptim}{\paramsSpLN^{*^v}}

\newcommand{\vlm}{f}

\newcommand{\vlmV}{\vlm^v}
\newcommand{\vlmT}{\vlm^t}

\newcommand{\vlmVSP}{\vlmV_{\paramsSpV}}
\newcommand{\vlmTSP}{\vlmT_{\paramsSpT}}

\newcommand{\vlmVSPOptim}{\vlmV_{\paramsSpV^*}}
\newcommand{\vlmTSPOptim}{\vlmT_{\paramsSpT^*}}

\newcommand{\vlmVLN}{\vlmV_{\paramsSpLNV}}
\newcommand{\vlmTLN}{\vlmT_{\paramsSpLNT}}

\newcommand{\vlmVLNOptim}{\vlmV_{\paramsSpLNVOptim}}
\newcommand{\vlmTLNOptim}{\vlmT_{\paramsSpLNTOptim}}

\newcommand{\textSpace}{\mathcal{T}}
\newcommand{\imageSpace}{\mathcal{I}}
\newcommand{\embedSpace}{\mathbb{R}^d}

\newcommand{\imageIn}{x}
\newcommand{\labelIn}{y}
\newcommand{\genClass}{c}
\newcommand{\baseClass}{b}

\newcommand{\baseClasses}{\mathcal{B}}
\newcommand{\novelClasses}{\mathcal{N}}
\newcommand{\genClasses}{\mathcal{C}}

\newcommand{\dataset}{D}
\newcommand{\datasetSize}{n}
\newcommand{\niter}{m}

\newcommand{\pairDataId}[1]{(\imageIn_#1,\labelIn_#1)}

\DeclareMathOperator*{\argmax}{arg\,max}

\newcommand{\ours}{2SFS\xspace}
\newcommand{\ourslora}{\ours{}\textsubscript{\texttt{LoRA}}}
\newcommand{\oursln}{\ours{}\textsubscript{\texttt{LayerNorm}}}

\usepackage{pifont}

\usepackage[table]{xcolor}
\definecolor{lncolor}{HTML}{FEE4C4}
\definecolor{baseline}{HTML}{E0E4E8}
\definecolor{ppink}{rgb}{0.98, 0.575, 0.89}

\newcommand\blfootnote[1]{%
  \begingroup
  \renewcommand\thefootnote{}\footnote{#1}%
  \addtocounter{footnote}{-1}%
  \endgroup
}

\definecolor{cvprblue}{rgb}{0.21,0.49,0.74}
\usepackage[pagebackref,breaklinks,colorlinks,allcolors=cvprblue]{hyperref}

\def\paperID{16413} %
\def\confName{CVPR}
\def\confYear{2025}

\title{Rethinking Few-Shot Adaptation of Vision-Language Models in Two Stages}

\author{
Matteo Farina\textsuperscript{1,*}
\quad 
Massimiliano Mancini\textsuperscript{1}
\quad
Giovanni Iacca\textsuperscript{1}
\quad
Elisa Ricci\textsuperscript{1,2} \\ 
\small
\textsuperscript{1}University of Trento \quad
\textsuperscript{2}Fondazione Bruno Kessler
}

\begin{document}
\maketitle

\begin{abstract}
An old-school recipe for training a classifier is to (i) learn a good feature extractor and (ii) optimize a linear layer atop. When only a handful of samples are available per category, as in Few-Shot Adaptation (FSA), data are insufficient to fit a large number of parameters, rendering the above impractical. 
This is especially true with large pre-trained Vision-Language Models (VLMs), which motivated successful research at the intersection of Parameter-Efficient Fine-tuning (PEFT) and FSA. %
In this work, we start by analyzing the learning dynamics of PEFT techniques when trained on few-shot data from only a subset of categories, referred to as the ``base'' classes. 
We show that such dynamics naturally splits into two distinct phases: 
(i) task-level feature extraction and (ii) specialization to the available concepts.
To accommodate this dynamic, we then depart from prompt- or adapter-based methods and tackle FSA differently.
Specifically, given a fixed computational budget, we split it to (i) learn a task-specific feature extractor via PEFT and (ii) train a linear classifier on top.
We call this scheme Two-Stage Few-Shot Adaptation (\ours).
Differently from established methods, our scheme enables a novel form of selective inference at a category level, \ie, at test time, only novel categories are embedded by the adapted text encoder, while embeddings of base categories are available within the classifier.
Results with \underline{fixed} hyperparameters across two settings, three backbones, and eleven datasets, show that \ours matches or surpasses the state-of-the-art, while established methods degrade significantly across settings.
\blfootnote{\textsuperscript{*} Corresponding author: \texttt{m.farina@unitn.it}. Code at \url{https://github.com/FarinaMatteo/rethinking_fewshot_vlms}}.
\end{abstract}
    
\section{Introduction}
\label{sec:intro}
Effective visual classification requires two key components: a good feature extractor and a strong classifier operating on features. 
This is established knowledge in computer vision, and multiple influential works \cite{sharif2014cnn,yosinski2014transferable} have demonstrated that a simple linear classifier, optimized atop a frozen feature extractor pre-trained at a larger scale (\eg, on ImageNet \cite{russakovsky2015imagenet}), achieves state-of-the-art or competitive results on a variety of downstream tasks \cite{zhai2019large}.
\begin{figure}
    \centering
    \includegraphics[width=1.\columnwidth]{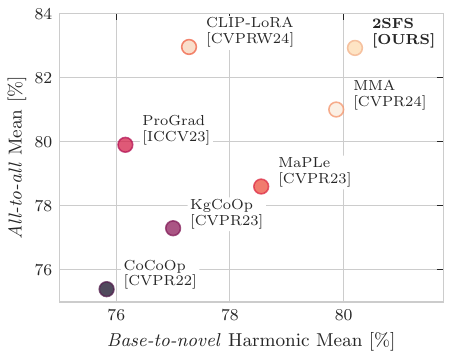}
    \caption{We present \ours{}, a technically simple revision of classifier tuning. \ours{} exhibits favorable performance both in \emph{all-to-all} FSA, where train/test categories coincide, as well as in the more challenging \emph{base-to-novel} setup, where only a subset of \emph{base} annotated categories are available, and the test suite further spans a set of unseen (\emph{novel}) classes. 
    Conversely, setting-specific SOTA approaches \cite{zanella2024cliplora, yang2024mma} degrade between settings.}
    \label{fig:teaser}
\end{figure}
Throughout adaptation, the feature extractor can also be fine-tuned together with the classifier, commonly with a lower learning rate to avoid disrupting the knowledge instilled from pretraining.
Notably, the success of this latter paradigm is linked to the curse of dimensionality and the risk of overfitting when there are too few annotated samples relative to the parameters. 

In this work, we deal with the Few-Shot Adaptation (FSA) of Vision-Language Models (VLMs), where the data-to-parameter ratio is at a critical level: there are typically hundreds of millions of parameters, but only a handful of data points are available on a per-category basis (\ie, the ``shots'').
Such a dilemma is emphasized when shots are available for a \emph{restricted} subset of categories, often referred to as the ``base'' categories, and the downstream task is assumed to span a broader range, comprising both base and ``novel'' semantic concepts.
This non-trivial challenge fueled modern research on Parameter-Efficient Fine-Tuning (PEFT) \cite{li2021prefix,hu2022lora,dettmers2024qlora}, with successful examples in the FSA of VLMs that are mostly categorized into two (non-conflicting) paradigms: (i) \emph{prompt tuning}, which optimizes a set of context vectors in either the text \cite{zhou2022coop,zhu2023prograd} or the vision encoder \cite{yao2023visual} and (ii) \emph{adapter-based} methods, where parametric functions are wrapped as external entities around (or on top of) the frozen VLM \cite{gao2024clip, zanella2024cliplora, yang2024mma}.

\emph{All} of the methods above consistently outperform full fine-tuning and linear probing.
However, this is mostly an \emph{a-posteriori} observation, and despite the abundance of literature in the field, very little is known about the \emph{learning dynamics} in FSA, \ie, about what happens throughout training with only a handful of examples.
We start this work by filling this gap, analyzing the behavior of CLIP \cite{radford2021learning} when adapting it to downstream tasks with different PEFT strategies. 
Surprisingly, we find that the learning dynamics exhibit consistent patterns across datasets when evaluated on held-out data: (i) an initial stage of \emph{joint} performance increase on both seen and unseen concepts; (ii) a \emph{breakpoint}, after which PEFT strategies still largely improve on base categories at the expense of degradation on unseen semantic concepts.
In other words, \emph{the initial stages of PEFT make for good task-level feature extraction, while the second stage specializes in the available data}.

Exploiting this finding, we introduce \textbf{Two} \textbf{S}tage \textbf{F}ew-\textbf{S}hot Adaptation (\ours), a technically simple revision of the old-school classifier tuning paradigm. 
Specifically, we split a fixed compute budget into two stages: first, we fine-tune only LayerNorm \cite{lei2016layer} instances of both modality-encoders to obtain a generalizable task-level feature extractor; second, we optimize a classifier on top, \ie, the text embeddings of base categories, to improve the discrimination ability of the model. 
In contrast to existing methods, \ours enables a novel form of selective inference at a category level, \ie, at test-time, only novel categories are embedded by the adapted text encoder, while base categories are available in $\mathcal{O}(1)$, being rows of the classifier matrix.

We thoroughly validate \ours \textit{with fixed hyperparameters} across a suite of 11 publicly available datasets, 2 different settings (\ie, base-to-novel and all-to-all), and 3 backbones, showing that our simple approach matches or surpasses setting-specific state-of-the-art methods, while they conversely degrade across settings (see Fig.~\ref{fig:teaser}). 

\begin{figure*}[t!]
     \centering
     \begin{subfigure}[b]{\linewidth}
         \centering
         \includegraphics[width=\textwidth]{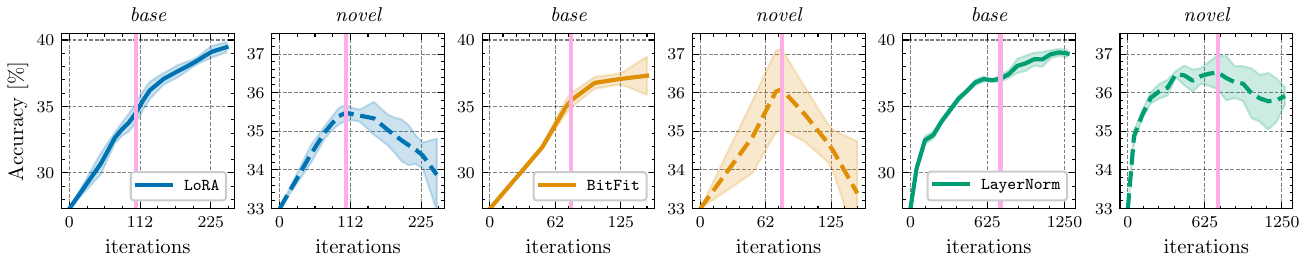}
         \caption{FGVC Aircraft}
         \label{fig:prel-fgvc}
     \end{subfigure}
     \hfill
     \begin{subfigure}[b]{\linewidth}
         \centering
         \includegraphics[width=\textwidth]{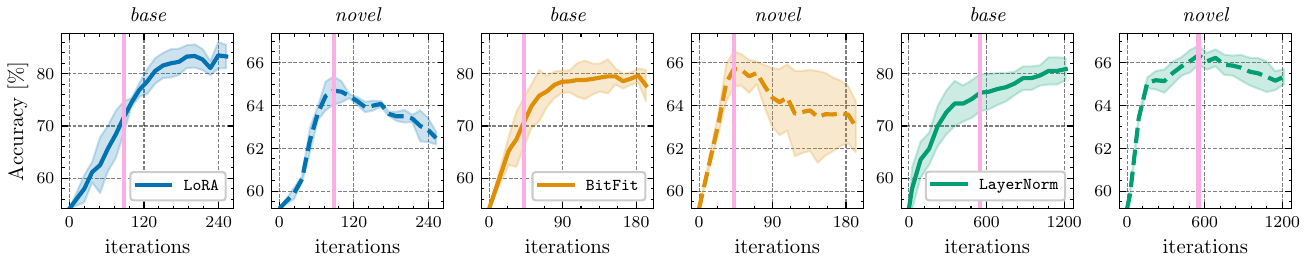}
         \caption{DTD}
         \label{fig:prel-dtd}
     \end{subfigure}
     \caption{The natural emergence of a \textcolor{ppink}{\textbf{\underline{breakpoint}}} during few-shot adaptation with different PEFT strategies. Before the breakpoint (to the left of the line), PEFT learns good task-level features, showed by \emph{joint} performance increase on both \emph{base} and \emph{novel} categories. After the breakpoint (to the right of the line), PEFT specializes in the available data, incurring unrecoverable performance degradation on novel categories accompanied by consistent improvement in base concepts. Results refer to CLIP \cite{radford2021learning} with the ViT-B/16 visual backbone \cite{dosovitskiy2020image}.}
     \label{fig:breakpoint}
\end{figure*}

\section{Related work}
\label{sec:related}
\mypar{Parameter-efficient fine-tuning} (PEFT) \cite{lialin2023scaling} adapts a large pretrained model by optimizing a small subset of parameters, while keeping the rest of the model frozen. 
One can identify three main categories among PEFT techniques: prompt tuning, selective approaches and adapter-based methods. 
Prompt tuning \cite{jia2022visual, lester2021power} adds trainable tokens either to the input or within intermediate layers. 
Initially designed for language prompts \cite{li2021prefix}, recent works have incorporated visual tokens into the recipe \cite{jia2022visual}. 
Selective methods optimize only a carefully chosen subset of \emph{existing} model parameters.
With vision architectures, popular choices are batch- or layer-normalization modules \cite{ioffe2015batch, lei2016layer}. 
Within the same categorization, BitFit (BIas-Term FIne-Tuning) \cite{zaken2022bitfit}, originally evaluated on NLP tasks, focuses on bias terms.
In contrast, adapter-based methods introduce \emph{external functions} to adapt the model.
Examples are scaling and shifting hidden features \cite{lian2022scaling}, adding non linear parametric functions on top of the frozen model outputs \cite{gao2024clip}, or side adapters, slightly refining features across layers \cite{mercea2024time}.
Recently, \emph{reparametrization-based} adapters have gained significant attention \cite{hu2022lora,kim2024hydra,dettmers2024qlora,liu2022few}, since their design allows to \emph{merge} the newly introduced parameters with the frozen model, incurring no additional inference cost. 
LoRA \cite{hu2022lora} is arguably the most popular example, optimizing a low-rank decomposition of the network's parameters.

\mypar{Few-shot adaptation of VLMs.} 
The impressive performance of modern VLMs has motivated researchers to develop adaptation techniques to expand their capabilities to specific tasks \cite{xing2024survey}. 
While some of these techniques are purely test-time methods requiring no supervision \cite{shu2022test, zanella2024test, farina2024frustratingly}, a common setup is to assume a handful of labeled examples are given.
In such a context, PEFT strategies have been designed to account for the multimodal nature of VLMs.
Following the previous categorization, the earliest approaches belong to the prompt tuning family \cite{zhou2022coop,zhou2022cocoop,zhu2023prograd}.
CoOp (Context Optimization) \cite{zhou2022coop} learns a set of soft context vectors to circumvent effortful manual prompting. 
CoCoOp (Conditional Context Optimization) \cite{zhou2022cocoop} improves CoOp, by generating an image-conditioned context. 
ProGrad (Prompt-aligned Gradient) \cite{zhu2023prograd} strives to preserve pretraining knowledge by only updating prompts when the gradient aligns with a specific direction. 
Similarly, KgCoOp (Knowledge-guided CoOp) \cite{yao2023visual} mitigates forgetting by guiding prompt learning with a hand-crafted prompt. 
PLOT (Prompt Learning with Optimal Transport) \cite{chen2022plot} and MaPLe (Multi-modal Prompt Learning) \cite{khattak2023maple} make a step further, connecting the visual and textual branches.
In PLOT \cite{chen2022plot}, class specific prompts are transported to visual features via optimal transport; MaPLe \cite{khattak2023maple} conditions deep visual prompts with their textual counterparts, with successful results.
A second category of FSA techniques relies on adapters \cite{gao2024clip,zhang2021tip,udandarao2023sus,yu2023task}. 
For example, CLIP-Adapter \cite{gao2024clip} mixes pretrained features with their non-linearly-transformed versions. 
TaskRes \cite{yu2023task} tunes a set of residual parameters instead of introducing a non-linear bottleneck.
Other methods \cite{zhang2021tip,udandarao2023sus} avoid fine-tuning and use the few-shot data as a cache to refine predictions. 

In this work, we introduce a new perspective: the benefit of a \emph{two-stage} design that adapts to the downstream task while improving generalization to unavailable categories.

\section{Preliminaries}
\label{sec:preliminaries}
In this section, we first formalize the FSA setup (Sec. \ref{sec:problem}). 
We then conduct a preliminary investigation on the impact of PEFT on the learning dynamics of FSA, with a particular emphasis on the relationship between generalization performance on base and novel categories (Sec. \ref{sec:pre-analysis}).

\subsection{Problem formulation}
\label{sec:problem}
FSA aims at adapting a model $\vlm$ to a specific downstream task given a few annotated examples per category. 
Formally, the available data in FSA are a collection of (image, category) pairs $\dataset = \{\pairDataId{1}, \cdots, \pairDataId{\datasetSize}\}$, with $x_i \in \mathcal{X}$ and $y_i \in \baseClasses$ being the available images and the %
set of \emph{base} categories, respectively.
A small number $k\in\mathbb{N}$ of available samples, referred to as \emph{shots}, is constant across categories, entailing that the available dataset has cardinality $\datasetSize=k\times|\baseClasses|$. 
In general, the downstream task spans a set $\genClasses$ of semantic categories, which is a \emph{superset} of the available annotated categories, \ie, $\baseClasses \subseteq \genClasses$, and no reliable assumptions can be made on the annotated data.
Hence, $\baseClasses$ can coincide with $\genClasses$, but also a set of \textit{novel} categories $\novelClasses \subseteq \genClasses$ may exist such that $\baseClasses \cap \novelClasses = \emptyset$ and $\baseClasses \cup \novelClasses = \genClasses$. 
One talks about \emph{all-to-all} FSA when $\novelClasses = \emptyset$ (or, equivalently, $\baseClasses = \genClasses$). 
Conversely, we talk about \emph{base-to-novel} generalization.

The function $\vlm$ is typically a contrastive VLM \cite{radford2021learning}, thus it contains a visual encoder $\vlmV:\imageSpace\rightarrow\embedSpace$ and a text encoder $\vlmT:\textSpace\rightarrow\embedSpace$ mapping images and texts in a shared $d$-dimensional Euclidean space. 
The domains $\imageSpace$ and $\textSpace$ can be thought of as the image and text spaces.
When classes are specified in natural language, \ie, each $c{\in}\genClasses$ is mapped to a string in $\textSpace$, zero-shot classification with $\vlm$ takes the form:
\begin{equation}
\label{eq:zero-shot}
    \vlm(x,\genClasses) = \argmax_{\genClass\in\genClasses} \left \{<\vlmV(\imageIn), \vlmT(\genClass)>, \forall \genClass \in \genClasses \right \}
\end{equation}
where $<\cdot, \cdot>$ denotes cosine similarity. 
Typically, the VLM $\vlm$ is parametrized by $\theta \in \mathbb{R}^N$, which largely exceed the number of available examples (\ie, $N >> \datasetSize$), rendering the optimization of $\theta$ subject to a high risk of overfitting.
Thus, a common strategy is to optimize a smaller set of task-specific parameters $\paramsSp$, according to a PEFT technique. 
Since $\paramsSp$ may influence both vision and text encoders, without loss of generality, we write it as $\paramsSp = \paramsSpV \cup \paramsSpT$.
In most cases, $\paramsSp$ are optimized through the standard softmax cross-entropy objective computed across \emph{base} categories (we omit the temperature for ease of notation):
\begin{equation}
    \label{eq:standard-loss}
    \mathcal{L}(x_i, y_i) = - \log \frac{\exp(<\vlmVSP(\imageIn_i), \vlmTSP(\labelIn_i)>)}{\sum_{\baseClass\in\baseClasses}\exp(<\vlmVSP(\imageIn_i), \vlmTSP(\baseClass)>)},
\end{equation}
where $\vlmVSP$ and $\vlmTSP$ are the visual and textual encoders modified by $\paramsSpV$ and $\paramsSpT$, respectively. 
The objective in \cref{eq:standard-loss} is minimized for a predefined number of iterations $m$.%
When optimization ends, the updated parameters $\paramsSp^* = \paramsSpV^* \cup \paramsSpT^*$ are used to parametrize $\vlm$, and \cref{eq:zero-shot} reads:
\begin{equation}
\label{eq:zero-shot-optim}
    \vlm(x,\genClasses) = \argmax_{\genClass\in\genClasses} \left \{<\vlmVSPOptim(\imageIn), \vlmTSPOptim(\genClass)>, \forall \genClass \in \genClasses \right \}.
\end{equation}

\subsection{What happens during Few-Shot Adaptation?}
\label{sec:pre-analysis}

A desideratum of FSA methods is that, by minimizing \cref{eq:standard-loss}, the optimized set of parameters $\paramsSp^*$ improves on base classes and generalizes to novel ones.
Mechanisms such as weight sharing across encoders \cite{yang2024mma} or cross-modal prompting \cite{zhou2022cocoop, khattak2023maple} have been explicitly developed to promote such behavior, suggesting that \emph{ad-hoc} designs are required to instill general task knowledge into $\vlm$.
In this section, we challenge this assumption by studying the learning dynamics of PEFT in FSA, \emph{when no explicit mechanism promoting generalization is involved in the learning recipe}.

\mypar{Preliminary PEFT Setup.} To conduct our study, we focus on three PEFT methods: LayerNorm tuning \cite{kim2022adaptVLMLayerNorm}, LoRA \cite{hu2022lora}, and BitFit \cite{zaken2022bitfit}. 
In \textbf{LayerNorm} tuning, typically used for visual adaptation \cite{de2023effectiveness}, $\paramsSp$ are the scale and bias parameters of the LayerNorm instances of the model.
\textbf{LoRA} (low-rank adapters), recently shown effective for VLMs \cite{zanella2024cliplora}, adds trainable low-rank matrices into each linear projection of the model, following a low-rank decomposition of the form $\mathbf{Wx} + \gamma\mathbf{BAx}$. 
In this case, $\paramsSp$ is the set of all low-rank matrices $\mathbf{B},\mathbf{A}$.
We adopt the design of CLIP-LoRA \cite{zanella2024cliplora}, which has extensively studied rank and placement of LoRA modules within CLIP, thereby plugging low-rank modules in all query, key, and value matrices of both encoders, with a rank $r{=}2$. 
\textbf{BitFit}, originally developed for language encoders \cite{zaken2022bitfit}, adapts the bias terms of all layers. 
In this latter case, $\paramsSp$ contains the shift vectors of all affine transformations in the network.
We apply such PEFT techniques on CLIP ViT-B/16, training with $k{=}16$ shots from both Describable Textures \cite{cimpoi14describing} and FGVC Aircraft \cite{maji13fine-grained}.
We follow the established base/novel classes split introduced in CoCoOp \cite{zhou2022cocoop} and set a budget of $m{=}8000$ iterations (\ie, gradient steps) for all PEFT techniques, which allows us to study the exact learning dynamics of the recent CLIP-LoRA \cite{zanella2024cliplora}.
See \cref{app:twostages} for the same analysis on other datasets.

\mypar{Experimental Outcome.} 
\cref{fig:breakpoint} displays the learning dynamics when the adapted model is evaluated on held-out samples.
From the outcome, it emerges that the dynamics of PEFT methods exhibit a breakpoint, \emph{naturally} separating adaptation into two distinct phases:
\begin{enumerate}
    \item \textbf{First Stage}: consistent and \emph{joint} improvement on \emph{both} base and novel categories.
    In other words, this means that PEFT is learning \emph{good task-level features} from both encoders. 
    If this were not the case, then only the performance on available categories would increase; 
    \item \textbf{Second Stage}: good task-level features are disrupted in favor of exploitation of the available categories.
    Note that, while this phenomenon looks akin to overfitting, it is not, since models are evaluated on held-out data for both $\baseClasses$ and $\novelClasses$. 
    In contrast, it has a different flavor: after the first stage, PEFT \emph{specializes} in the available categories, overriding knowledge that would have been helpful for the downstream task as a whole.
\end{enumerate}

\noindent{}We find that the pattern is consistent across datasets and PEFT techniques.
Between the two stages, the \textbf{First Stage} is the most counter-intuitive: PEFT learns good vision-language features, although \emph{no} ad-hoc designs were injected to promote such behavior.
We also observe traits characterizing PEFT methods.
For example, LoRA tends to better discriminate among base categories, while LayerNorm appears less capable in this respect. 
However, it is much more robust in the second stage and reaches the breakpoint far later than LoRA and BitFit. 
We now leverage these observations to design a simple yet effective FSA strategy.

\begin{figure*}
    \centering
    \includegraphics[width=1.\textwidth]{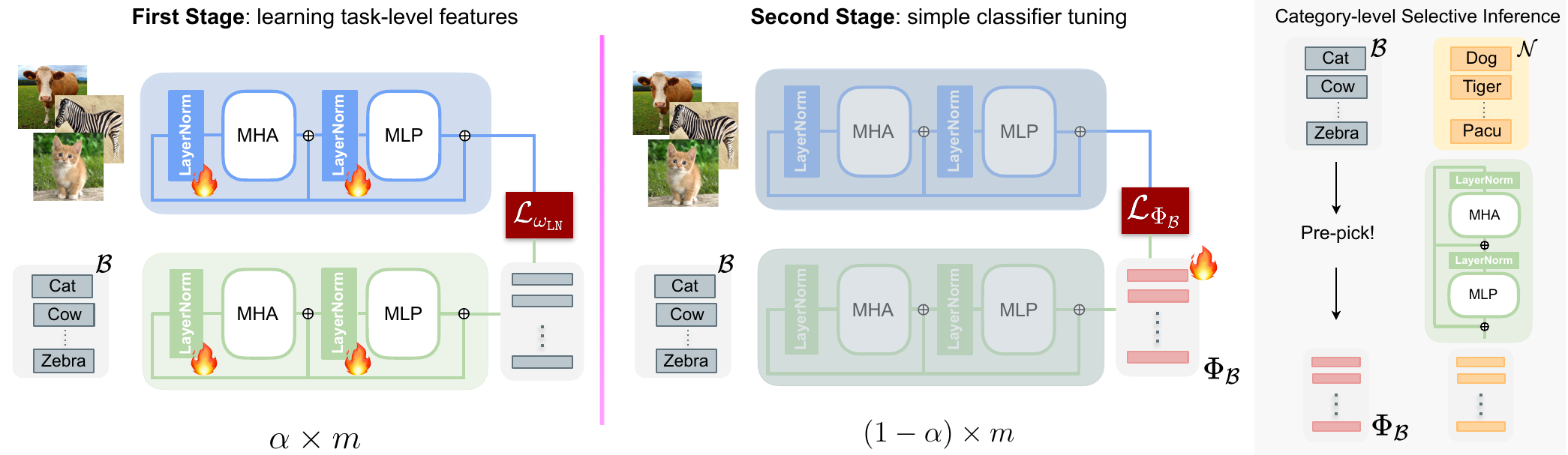}
    \caption{\textbf{\ours{}}. Given a computational budget of $\niter$ iterations, \ours{} operates in two separate stages. In the \textbf{First Stage}, \ours{} learns task-level features by tuning LayerNorm instances. In the \textbf{Second Stage}, a simple classifier initialized with the text embeddings of base categories learns to separate task-level features. 
    At inference time, \ours{} allows to selectively embed categories. Specifically, only novel categories are embedded by the adapted text encoder, while embeddings of base categories are available as rows within $\Classifier_\baseClasses$.}
    \label{fig:method}
\end{figure*}
\section{Two-Stage Few-Shot Adaptation}
\label{sec:method}
This section introduces \textbf{Two}-\textbf{S}tage \textbf{F}ew-\textbf{S}hot Adaptation (\ours), and describes how we leverage the natural emergence of two stages during FSA.
\ours{} revises the old-school paradigm of classifier tuning \cite{wortsman2022robust}, which has been set aside in favor of prompt- or adapter-based methods.
Given a fixed computational budget expressed as a maximum number of iterations $m$,  \ours{} acts in two short stages: 
\begin{enumerate}
    \item \textbf{First Stage}: initially, \ours{} allocates $\alpha \times m$ iterations to update the feature extractor via PEFT to improve base \emph{and} novel class performance, with $\alpha\in[0,1]$. 
    Motivated by the results of \cref{fig:breakpoint}, highlighting the enhanced robustness of Layer Normalization, we describe the rest of this section assuming that $\paramsSp$ are LayerNorm scale and shifts. 
    We additionally report \ours{} with LoRA in \cref{app:lora}. 
    \item \textbf{Second Stage}: for the remaining $(1-\alpha) \times m$ steps, \ours \emph{switches} parameters to avoid disrupting helpful task-level features and learns a \emph{base} classifier. 
\end{enumerate} 

At inference time \ours enables a novel form of adaptive inference on a \emph{per-category basis}, \ie, we skip the computation of the text encoder when embedding categories in $\baseClasses$, as they are readily available as rows of the classifier.
We now dive into the %
details of these components.

\subsection{First stage: learning task-level features}
In the first stage, we want to learn a good feature extractor for base categories that also generalizes to novel concepts. 
To achieve this, we exploit the first learning phase in PEFT, where the performances on both sets increase. 
In this phase, we tune all LayerNorm instances of both visual and textual encoders.
Formally, given a $d$-dimensional vector $\mathbf{a}\in \mathbb{R}^d$ of activations, LayerNorm \cite{lei2016layer} operates by scaling and shifting the standardized features in $\mathbf{a}$, \ie
\begin{equation}
    \label{eq:LN}
    \texttt{LayerNorm}(\mathbf{a}) = \gamma \odot \left (\frac{\mathbf{a} - \mu(\mathbf{a})}{\sigma(\mathbf{a})} \right ) + \beta,
\end{equation}
where $\mu(\mathbf{a}) \in \mathbb{R}$ is the mean of the vector, $\sigma(\mathbf{a}) \in \mathbb{R}$ is its standard deviation, while $\gamma$ and $\beta$ are known as the ``scale'' and ``shift'' parameters. 
Both $\gamma$ and $\beta$ are learnable $d$-dimensional vectors and are subject to fine-tuning. 

Borrowing from the notation of \cref{sec:preliminaries}, we denote as task-specific parameters the scale and shift vectors of each LayerNorm instance of the network.
Assuming a total of $L$ instances, we have $\paramsSpLN = \{(\gamma_1, \beta_1), \cdots, (\gamma_L, \beta_L)\}$, which are the union of modality-specific parameters $\paramsSpLNV$ and $\paramsSpLNT$. 
The objective function in Eq. \eqref{eq:standard-loss} now reads:
\begin{equation}
    \label{eq:1stage-loss}
    \mathcal{L}_{\omega_\mathtt{LN}}(x_i, y_i) = - \log \frac{\exp(<\vlmVLN(\imageIn_i), \vlmTLN(\labelIn_i)>)}{\sum_{\baseClass\in\baseClasses}\exp(<\vlmVLN(\imageIn_i), \vlmTLN(\baseClass)>)},
\end{equation}
which we optimize for a $\alpha \times \niter$ iterations, where $\alpha \in [0, 1]$. 
Thus, this stage only takes a \emph{fraction} of the available compute budget and ensures that no further optimization is carried out \wrt $\paramsSpLN$ after the natural breakpoint is reached (\ie, ideally, $\alpha \times \niter$ should match the pink line of \cref{fig:breakpoint}).

\subsection{Second stage: simple classifier tuning} 
After the first stage, we obtain optimized parameters $\paramsSpLN^* = \paramsSpLNVOptim \cup \paramsSpLNTOptim$ that incorporate general task knowledge.
From \cref{sec:pre-analysis}, we know that further tuning $\paramsSpLN^*$ disrupts helpful features for novel categories, hence we circumvent this issue by carrying out optimization \wrt a different set of parameters.
We employ arguably the simplest form to do so: we \emph{freeze} $\paramsSpLN^*$ and train a linear classifier on top.
Let $\classifier_\baseClass = \vlmTLNOptim(\baseClass)$ denote the embedding of the category $\baseClass$, obtained with the first-stage text encoder.
We initialize the classifier weights $\Classifier_\baseClasses = \{\classifier_\baseClass\}_{\baseClass \in \baseClasses}$ by stacking all embeddings of base categories, and optimize $\Classifier_\baseClasses$ for the remaining $(1-\alpha) \times \niter$ steps. 
At this stage, \cref{eq:standard-loss} becomes:
\begin{equation}
    \label{eq:2stage-loss}
    \mathcal{L}_{\Classifier_\baseClasses}(x_i, y_i) = - \log \frac{\exp(<\vlmVLNOptim(\imageIn_i), \classifier_{y_i}>)}{\sum_{\baseClass\in\baseClasses}\exp(<\vlmVLNOptim(\imageIn_i), \phi_\baseClass>)},
\end{equation}
where $\Classifier_\baseClasses$ learns to separate features obtained with the frozen first-stage visual parameters $\paramsSpLNVOptim$.
Along with its simplicity, this design fully exploits helpful task-level representations: $\Classifier_\baseClasses$ is initialized with first-stage text embeddings and learns to separate first-stage visual embeddings.
The optimized classifier weights are denoted as $\Classifier_\baseClasses^* = \{\classifier^*_\baseClass\}_{\baseClass \in \baseClasses}$.

\subsection{Category-level Selective Inference}
The proposed two-stage design provides unique advantages, allowing for selective inference within the text encoder.
Specifically, given an image $x$ and a set of downstream categories $\genClasses$ to discriminate, our prediction function follows:
\begin{equation}
    \label{eq:inference}
    \vlm_{\paramsSpLNOptim, \Classifier_\baseClasses^*}(\imageIn) =  \argmax_{\genClass\in\genClasses} <\vlmVLNOptim(\imageIn), \classifier^*_\genClass>
\end{equation}
with each $\classifier^*_\genClass$ being:
\begin{equation}
    \label{eq:inference-splitting}
    \classifier^*_\genClass = \begin{cases}
    \classifier^*_\baseClass & \text{if $\exists \baseClass \in \baseClasses~|~\baseClass = \genClass$}\\
    \vlmTLNOptim(\genClass) & \text{otherwise}.
    \end{cases}
\end{equation}
In essence, at inference time, the two-stage design allows to \emph{skip} computations within the text encoder, instead of embedding the categories that were available in the annotated data (which are always known in FSA), and to pick the corresponding row of the classifier in $\mathcal{O}(1)$ directly.
In contrast, images and unseen categories are always processed with the parameters obtained after the first stage.

In summary, \ours{} provides a \textit{unified recipe} for FSA: splitting into stages and optimizing a different parameter subset allows to \ding{172} preserve helpful knowledge for unseen categories and \ding{173} exploit available annotations.
The overall pipeline of \ours{} is schematized by \cref{fig:method}.

\begin{table*}[ht!]
\caption{Experiments in \emph{base-to-novel} generalization with the ViT-B/16 visual backbone. All methods use $k{=}16$ shots per base class. ``CLIP'' refers to zero-shot performance with dataset-specific templates, \eg, ``\emph{a photo of a \{\}, a type of flower}'' for Oxford Flowers.}
\label{tab:b2n}
    \def\arraystretch{1.075}
    \centering
    \scriptsize
    \begin{minipage}{0.3\textwidth}
        \centering
        \caption*{\textbf{Average across datasets.}}\vspace{-1em}
        \begin{tabular}{lcc|c}
        \toprule
         \textbf{Method} & \textbf{Base} & \textbf{Novel} & \textbf{HM} \\
         \midrule
         CLIP \cite{radford2021learning} & 69.34 & 74.22 & 71.70 \\
         CoOp \cite{zhou2022coop} & 82.69 & 63.22 & 71.66 \\
         CoCoOp \cite{zhou2022cocoop} & 80.47 & 71.69 & 75.83 \\
         MaPLe \cite{khattak2023maple} & 82.28 & 75.14 & 78.55 \\
         ProGrad \cite{zhu2023prograd} & 82.48 & 70.75 & 76.16 \\
         KgCoOp \cite{yao2023visual} & 80.73 & 73.60 & 77.00 \\
         CLIP-LoRA \cite{zanella2024cliplora} & 85.32 & 70.63 & 77.28 \\
         MMA \cite{yang2024mma} & 83.20 & 76.80 & 79.87 \\
         \cmidrule(lr){1-4}
         \rowcolor{lncolor}
         \ours & 85.55 & 75.48 & \textbf{80.20} \\
         \bottomrule
        \end{tabular}
        \vspace{0.5em}
    \end{minipage}%
    \hfill
    \centering
    \begin{minipage}{0.3\textwidth}
        \centering
        \caption*{ImageNet}\vspace{-1em}
        \begin{tabular}{lcc|c}
        \toprule
         \textbf{Method} & \textbf{Base} & \textbf{Novel} & \textbf{HM} \\
         \midrule
         CLIP \cite{radford2021learning} & 72.43 & 68.14 & 70.22 \\
         CoOp \cite{zhou2022coop} & 76.47 & 67.88 & 71.92 \\
         CoCoOp \cite{zhou2022cocoop} & 75.98 & 70.43 & 73.10 \\
         MaPLe \cite{khattak2023maple} & 76.66 & 70.54 & 73.47 \\
         ProGrad \cite{zhu2023prograd} & 77.02 & 66.66 & 71.46 \\
         KgCoOp \cite{yao2023visual} & 75.83 & 69.96 & 72.78 \\
         CLIP-LoRA \cite{zanella2024cliplora} & 77.58 & 68.76 & 72.91 \\
         MMA \cite{yang2024mma} & 77.31 & 71.00 & 74.02 \\
         \cmidrule(lr){1-4}
         \rowcolor{lncolor}
         \ours & 77.71 & 70.99 & \textbf{74.20}  \\
         \bottomrule
        \end{tabular}
        \vspace{0.5em}
    \end{minipage}%
    \hfill
    \centering
    \begin{minipage}{0.3\textwidth}
        \centering
        \caption*{Caltech101}\vspace{-1em}
        \begin{tabular}{lcc|c}
        \toprule
         \textbf{Method} & \textbf{Base} & \textbf{Novel} & \textbf{HM} \\
         \midrule
         CLIP \cite{radford2021learning} & 96.84 & 94.00 & 95.40 \\
         CoOp \cite{zhou2022coop} & 98.00 & 89.81 & 93.73 \\
         CoCoOp \cite{zhou2022cocoop} & 97.96 & 93.81 & 95.84 \\
         MaPLe \cite{khattak2023maple} & 97.74 & 94.36 & 96.02 \\
         ProGrad \cite{zhu2023prograd} & 98.02 & 93.89 & 95.91 \\
         KgCoOp \cite{yao2023visual} & 97.72 & 94.39 & 96.03 \\
         CLIP-LoRA \cite{zanella2024cliplora} & 98.19 & 93.05 & 95.55 \\
         MMA \cite{yang2024mma} & 98.40 & 94.00 & 96.15 \\
         \cmidrule(lr){1-4}
         \rowcolor{lncolor}
         \ours & 98.71 & 94.43 & \textbf{96.52} \\
         \bottomrule
        \end{tabular}
        \vspace{0.5em}
    \end{minipage}%
    \hfill
    \begin{minipage}{0.3\textwidth}
        \centering
        \caption*{Oxford Flowers}\vspace{-1em}
        \begin{tabular}{lcc|c}
        \toprule
         \textbf{Method} & \textbf{Base} & \textbf{Novel} & \textbf{HM} \\
         \midrule
         CLIP \cite{radford2021learning} & 72.08 & 77.80 & 74.83 \\
         CoOp \cite{zhou2022coop} & 97.60 & 59.67 & 74.06 \\
         CoCoOp \cite{zhou2022cocoop} & 94.87 & 71.75 & 81.71 \\
         MaPLe \cite{khattak2023maple} & 95.92 & 72.46 & 82.56 \\
         ProGrad \cite{zhu2023prograd} & 95.54 & 71.87 & 82.03 \\
         KgCoOp \cite{yao2023visual} & 95.00 & 74.73 & 83.65 \\ 
         CLIP-LoRA \cite{zanella2024cliplora} & 97.91 & 68.61 & 80.68 \\
         MMA \cite{yang2024mma} & 97.77 & 75.93 & 85.48 \\
         \cmidrule(lr){1-4}
         \rowcolor{lncolor}
         \ours & 98.29 & 76.17 & \textbf{85.83} \\
         \bottomrule
        \end{tabular}
        \vspace{0.5em}
    \end{minipage}
    \hfill
    \begin{minipage}{0.3\textwidth}
        \centering
        \caption*{Oxford Pets}\vspace{-1em}
        \begin{tabular}{lcc|c}
        \toprule
         \textbf{Method} & \textbf{Base} & \textbf{Novel} & \textbf{HM} \\
         \midrule
         CLIP \cite{radford2021learning} & 91.17 & 97.26 & 94.12 \\
         CoOp \cite{zhou2022coop} &  93.67 & 95.29 & 94.47 \\
         CoCoOp \cite{zhou2022cocoop} & 95.20 & 97.69 & 96.43 \\
         MaPLe \cite{khattak2023maple} & 95.43 & 97.76 & 96.58 \\
         ProGrad \cite{zhu2023prograd} & 95.07 & 97.63 & 96.33 \\
         KgCoOp \cite{yao2023visual} & 94.65 & 97.76 & 96.18 \\
         CLIP-LoRA \cite{zanella2024cliplora} & 94.36 & 95.71 & 95.03 \\
         MMA \cite{yang2024mma} & 95.40 & 98.07 & \textbf{96.72} \\
         \cmidrule(lr){1-4}
         \rowcolor{lncolor}
         \ours & 95.32 & 97.82 & 96.55 \\
         \bottomrule
        \end{tabular}
        \vspace{0.5em}
    \end{minipage}%
    \hfill
    \begin{minipage}{0.3\textwidth}
        \centering
        \caption*{Stanford Cars}\vspace{-1em}
        \begin{tabular}{lcc|c}
        \toprule
         \textbf{Method} & \textbf{Base} & \textbf{Novel} & \textbf{HM} \\
         \midrule
         CLIP \cite{radford2021learning} & 63.37 & 74.89 & 68.65 \\
         CoOp \cite{zhou2022coop} & 78.12 & 60.40 & 68.13 \\
         CoCoOp \cite{zhou2022cocoop} & 70.49 & 73.59 & 72.01 \\
         MaPLe \cite{khattak2023maple} & 72.94 & 74.00 & 73.47 \\
         ProGrad \cite{zhu2023prograd} & 77.68 & 68.63 & 72.88 \\
         KgCoOp \cite{yao2023visual} & 71.76 & 75.04 & 73.36 \\
         CLIP-LoRA \cite{zanella2024cliplora} & 83.93 & 65.54 & 73.60 \\
         MMA \cite{yang2024mma} & 78.50 & 73.10 & 75.70 \\
         \cmidrule(lr){1-4}
         \rowcolor{lncolor}
         \ours & 82.50 & 74.80 & \textbf{78.46} \\
         \bottomrule
        \end{tabular}
        \vspace{0.5em}
    \end{minipage}%
    \hfill
    \begin{minipage}{0.3\textwidth}
        \centering
        \caption*{Food 101}\vspace{-1em}
        \begin{tabular}{lcc|c}
        \toprule
         \textbf{Method} & \textbf{Base} & \textbf{Novel} & \textbf{HM} \\
         \midrule
         CLIP \cite{radford2021learning} & 90.10 & 91.22 & 90.66 \\
         CoOp \cite{zhou2022coop} & 88.33 & 82.26 & 85.19 \\
         CoCoOp \cite{zhou2022cocoop} & 90.70 & 91.29 & 90.99 \\
         MaPLe \cite{khattak2023maple} & 90.71 & 92.05 & \textbf{91.38} \\
         ProGrad \cite{zhu2023prograd} & 90.37 & 89.59 & 89.98 \\
         KgCoOp \cite{yao2023visual} & 90.50 & 91.70 & 91.09 \\
         CLIP-LoRA \cite{zanella2024cliplora} & 86.84 & 86.67 & 86.76 \\
         MMA \cite{yang2024mma} & 90.13 & 91.30 & 90.71 \\
         \cmidrule(lr){1-4}
         \rowcolor{lncolor}
         \ours & 89.11 & 91.34 & 90.21 \\
         \bottomrule
        \end{tabular}
        \vspace{0.5em}
    \end{minipage}
    \hfill
    \begin{minipage}{0.3\textwidth}
        \centering
        \caption*{FGVC Aircraft}\vspace{-1em}
        \begin{tabular}{lcc|c}
        \toprule
         \textbf{Method} & \textbf{Base} & \textbf{Novel} & \textbf{HM} \\
         \midrule
         CLIP \cite{radford2021learning} & 27.19 & 36.29 & 31.09 \\
         CoOp \cite{zhou2022coop} & 40.44 & 22.30 & 28.75 \\
         CoCoOp \cite{zhou2022cocoop} & 33.41 & 23.71 & 27.74 \\
         MaPLe \cite{khattak2023maple} & 37.44 & 35.61 & 36.50 \\
         ProGrad \cite{zhu2023prograd} & 40.54 & 27.57 & 32.82 \\
         KgCoOp \cite{yao2023visual} & 36.21 & 33.55 & 34.83 \\
         CLIP-LoRA \cite{zanella2024cliplora} & 50.10 & 26.03 & 34.26 \\
         MMA \cite{yang2024mma} & 40.57 & 36.33 & 38.33 \\
         \cmidrule(lr){1-4}
         \rowcolor{lncolor}
         \ours & 47.48 & 35.51 & \textbf{40.63} \\
         \bottomrule
        \end{tabular}
        \vspace{0.5em}
    \end{minipage}%
    \hfill
    \begin{minipage}{0.3\textwidth}
        \centering
        \caption*{SUN 397}\vspace{-1em}
        \begin{tabular}{lcc|c}
        \toprule
         \textbf{Method} & \textbf{Base} & \textbf{Novel} & \textbf{HM} \\
         \midrule
         CLIP \cite{radford2021learning} & 69.36 & 75.35 & 72.23 \\
         CoOp \cite{zhou2022coop} & 80.60 & 65.89 & 72.51 \\
         CoCoOp \cite{zhou2022cocoop} & 79.74 & 76.86 & 78.27 \\
         MaPLe \cite{khattak2023maple} & 80.82 & 78.70 & 79.75 \\
         ProGrad \cite{zhu2023prograd} & 81.26 & 74.17 & 77.55 \\
         KgCoOp \cite{yao2023visual} & 80.29 & 76.53 & 78.36 \\
         CLIP-LoRA \cite{zanella2024cliplora} & 81.11 & 74.53 & 77.68 \\
         MMA \cite{yang2024mma} & 82.27 & 78.57 & 80.38 \\
         \cmidrule(lr){1-4}
         \rowcolor{lncolor}
         \ours & 82.59 & 78.91 & \textbf{80.70} \\
         \bottomrule
        \end{tabular}
        \vspace{0.5em}
    \end{minipage}
    \hfill
    \begin{minipage}{0.3\textwidth}
        \centering
        \caption*{DTD}\vspace{-1em}
        \begin{tabular}{lcc|c}
        \toprule
         \textbf{Method} & \textbf{Base} & \textbf{Novel} & \textbf{HM} \\
         \midrule
         CLIP \cite{radford2021learning} & 53.24 & 59.90 & 56.37 \\
         CoOp \cite{zhou2022coop} & 79.44 & 41.18 & 54.24 \\
         CoCoOp \cite{zhou2022cocoop} & 77.01 & 56.00 & 64.85 \\
         MaPLe \cite{khattak2023maple} & 80.36 & 59.18 & 68.16 \\
         ProGrad \cite{zhu2023prograd} & 77.35 & 52.35 & 62.45 \\
         KgCoOp \cite{yao2023visual} & 77.55 & 54.99 & 64.35 \\
         CLIP-LoRA \cite{zanella2024cliplora} & 83.95 & 62.84 & 71.39 \\
         MMA \cite{yang2024mma} & 83.20 & 65.63 & 73.38 \\
         \cmidrule(lr){1-4}
         \rowcolor{lncolor}
         \ours & 84.60 & 65.01 & \textbf{73.52} \\
         \bottomrule
        \end{tabular}
    \end{minipage}%
    \hfill
    \begin{minipage}{0.3\textwidth}
        \centering
        \caption*{EuroSAT}\vspace{-1em}
        \begin{tabular}{lcc|c}
        \toprule
         \textbf{Method} & \textbf{Base} & \textbf{Novel} & \textbf{HM} \\
         \midrule
         CLIP \cite{radford2021learning} & 56.48 & 64.05 & 60.03 \\
         CoOp \cite{zhou2022coop} & 92.19 & 54.74 & 68.69\\
         CoCoOp \cite{zhou2022cocoop} & 87.49 & 60.04 & 71.21 \\
         MaPLe \cite{khattak2023maple} & 94.07 & 73.23 & 82.35 \\
         ProGrad \cite{zhu2023prograd} & 90.11 & 60.89 & 72.67 \\
         KgCoOp \cite{yao2023visual} & 85.64 & 64.34 & 73.48 \\
         CLIP-LoRA \cite{zanella2024cliplora} & 97.04 & 62.50 & 76.03 \\
         MMA \cite{yang2024mma} & 85.46 & 82.34 & \textbf{83.87} \\
         \cmidrule(lr){1-4}
         \rowcolor{lncolor}
         \ours & 96.91 & 67.09 & 79.29 \\
         \bottomrule
        \end{tabular}
    \end{minipage}
    \hfill
    \begin{minipage}{0.3\textwidth}
        \centering
        \caption*{UCF101}\vspace{-1em}
        \begin{tabular}{lcc|c}
        \toprule
         \textbf{Method} & \textbf{Base} & \textbf{Novel} & \textbf{HM} \\
         \midrule
         CLIP \cite{radford2021learning} & 70.53 & 77.50 & 73.85 \\
         CoOp \cite{zhou2022coop} & 84.69 & 56.05 & 67.46 \\
         CoCoOp \cite{zhou2022cocoop} & 82.33 & 73.45 & 77.64 \\
         MaPLe \cite{khattak2023maple} & 83.00 & 78.66 & 80.77 \\
         ProGrad \cite{zhu2023prograd} & 84.33 & 74.94 & 79.35 \\
         KgCoOp \cite{yao2023visual} & 82.89 & 76.67 & 79.65 \\
         CLIP-LoRA \cite{zanella2024cliplora} & 87.52 & 72.74 & 79.45 \\
         MMA \cite{yang2024mma} & 86.23 & 80.03 & 82.20 \\
         \cmidrule(lr){1-4}
         \rowcolor{lncolor}
         \ours & 87.85 & 78.19 & \textbf{82.74} \\
         \bottomrule
        \end{tabular}
    \end{minipage}
\end{table*}

\begin{table*}[t!]
    \def\arraystretch{1.1}
    \centering
    \footnotesize
    \caption{\emph{All-to-all} experiments, where train/test categories coincide, with the ViT-B/16 (top), ViT-B/32 (middle), and ViT-L/14 (bottom) backbones. All methods use $k=16$ shots per class. In each group, the best performer is marked by \textbf{bold text}; the second best is \underline{underlined}.}
    \begin{adjustbox}{max width=\textwidth}
    \begin{tabular}{llccccccccccc|c}
    \toprule
     \textsc{\textbf{Backbone}} &  \textsc{\textbf{Method}} &  \textsc{\textbf{ImageNet}} &  \textsc{\textbf{SUN}} &  \textsc{\textbf{AIR}} &  \textsc{\textbf{ESAT}} &  \textsc{\textbf{CARS}} &  \textsc{\textbf{FOOD}} &  \textsc{\textbf{PETS}} &  \textsc{\textbf{FLWR}} &  \textsc{\textbf{CAL}} &  \textsc{\textbf{DTD}} &  \textsc{\textbf{UCF}} &  \textsc{\textbf{Mean}} \\
    \midrule
    
    \cellcolor{gray!0} \multirow{13}{*}{\texttt{ViT-B/16}} 
& \emph{Zero-Shot} &  66.7 & 62.6 & 24.7 & 47.5 & 65.3 & 86.1 & 89.1 & 71.4 & 92.9 & 43.6 & 66.7 & 65.1 \\

& CoOp \cite{zhou2022coop} (ctx=16) &  71.9 & 74.9 & 43.2 & 85.0 & 82.9 & 84.2 & 92.0 & 96.8 & 95.8 & 69.7 & 83.1 & 80.0 \\

& CoCoOp \cite{zhou2022cocoop} &  71.1 & 72.6 & 33.3 & 73.6 & 72.3 & \textbf{87.4} & 93.4 & 89.1 & 95.1 & 63.7 & 77.2 & 75.4 \\

& TIP-Adapter-F \cite{zhang2021tip} &  73.4 & 76.0 & 44.6 & 85.9 & 82.3 & 86.8 & 92.6 & 96.2 & 95.7 & 70.8 & 83.9 & 80.7 \\

& CLIP-Adapter \cite{gao2024clip} &  69.8 & 74.2 & 34.2 & 71.4 & 74.0 & 87.1 & 92.3 & 92.9 & 94.9 & 59.4 & 80.2 & 75.5 \\

& PLOT++ \cite{chen2022plot} &  72.6 & 76.0 & 46.7 & 92.0 & 84.6 & 87.1 & 93.6 & 97.6 & 96.0 & 71.4 & 85.3 & 82.1 \\

& KgCoOp \cite{yao2023visual} &  70.4 & 73.3 & 36.5 & 76.2 & 74.8 & 87.2 & 93.2 & 93.4 & 95.2 & 68.7 & 81.7 & 77.3 \\

& TaskRes \cite{yu2023task} &  73.0 & 76.1 & 44.9 & 82.7 & 83.5 & 86.9 & 92.4 & 97.5 & 95.8 & 71.5 & 84.0 & 80.8 \\

& MaPLe \cite{khattak2023maple} &  71.9 & 74.5 & 36.8 & 87.5 & 74.3 & \textbf{87.4} & 93.2 & 94.2 & 95.4 & 68.4 & 81.4 & 78.6 \\

& ProGrad \cite{zhu2023prograd} &  72.1 & 75.1 & 43.0 & 83.6 & 82.9 & 85.8 & 92.8 & 96.6 & 95.9 & 68.8 & 82.7 & 79.9 \\

& LP++ \cite{huang2024lp++} & 73.0 & 76.0 & 42.1 & 85.5 & 80.8 & 87.2 & 92.6 & 96.3 & 95.8 & 71.9 & 83.9 & 80.5 \\

& CLIP-LoRA \cite{zanella2024cliplora} & \underline{73.6} & 76.1 & \textbf{54.7} & \underline{92.1} & \textbf{86.3} & 84.2 & 92.4 & \textbf{98.0} & \textbf{96.4} & 72.0 & \textbf{86.7} & \textbf{83.0} \\

& MMA \cite{yang2024mma} & 73.2 & \underline{76.6} & 44.7 & 85.0 & 80.2 & 87.0 & \textbf{93.9} & 96.8 & 95.8 & \underline{72.7} & 85.0 & 81.0 \\

\rowcolor{lncolor}
\cellcolor{gray!0} & \ours & \textbf{73.7} & \textbf{77.0} & \underline{50.0} & \textbf{92.4} & \underline{85.4} & 86.1 & \underline{93.7} & \underline{97.7} & \textbf{96.4} & \textbf{73.2} & \underline{86.6} & \underline{82.9} \\
 \\ [-5ex] \\ 
    \cmidrule(lr){1-14}
    
    \cellcolor{gray!0}\multirow{13}{*}{\texttt{ViT-B/32}} 
& \emph{Zero-Shot} & 61.9 & 62.0 & 19.3 & 45.1 & 60.4 & 80.5 & 87.5 & 67.0 & 91.1 & 42.6 & 62.2 & 61.8 \\

& CoOp \cite{zhou2022coop} (ctx=16) & 66.8 & 72.2 & 32.9 & 83.3 & 76.0 & 78.6 & 88.7 & 95.4 & 94.9 & 65.3 & 78.6 & 75.7 \\

& CoCoOp \cite{zhou2022cocoop} & 66.0 & 69.8 & 22.6 & 70.4 & 64.6 & \textbf{81.9} & \underline{91.0} & 82.5 & 94.3 & 59.7 & 75.3 & 70.7 \\

& TIP-Adapter-F \cite{zhang2021tip} & \textbf{68.4} & \underline{74.1} & 34.8 & 83.4 & 77.0 & 81.7 & 90.4 & 94.3 & 95.1 & 68.0 & 80.5 & 77.1 \\

& CLIP-Adapter \cite{gao2024clip} & 64.9 & 71.8 & 26.7 & 64.7 & 68.9 & \textbf{81.9} & 90.1 & 88.7 & 94.8 & 58.1 & 76.5 & 71.6 \\

& PLOT++ \cite{chen2022plot} & 67.4 & 73.4 & 36.3 & 91.1 & 77.4 & 79.7 & 89.1 & \textbf{96.3} & 94.9 & 67.0 & 81.5 & 77.6 \\

& KgCoOp \cite{yao2023visual} & 65.4 & 71.0 & 23.7 & 70.1 & 67.3 & 81.7 & 90.8 & 86.1 & 94.4 & 65.1 & 77.5 & 72.1 \\

& TaskRes  \cite{yu2023task} & 68.2 & 73.6 & 37.0 & 77.7 & 78.0 & 81.4 & 89.4 & 95.5 & 95.7 & 68.3 & 80.6 & 76.9 \\

& MaPLe \cite{khattak2023maple} & 66.7 & 72.0 & 28.0 & 83.3 & 66.9 & 82.1 & 91.7 & 89.0 & 95.1 & 63.4 & 77.3 & 74.1 \\

& ProGrad \cite{zhu2023prograd} & 66.9 & 73.2 & 33.3 & 81.0 & 76.1 & 80.1 & 89.3 & 95.1 & 95.0 & 65.8 & 79.6 & 75.9 \\

& LP++ \cite{huang2024lp++} & 68.1 & 74.0 & 34.3 & 82.8 & 75.2 & 81.8 & 90.5 & 93.9 & 95.0 & 67.8 & 80.1 & 76.7 \\

& CLIP-LoRA \cite{zanella2024cliplora} & \textbf{68.4} & 74.0 & \textbf{44.9} & \underline{91.8} & \underline{79.7} & 78.2 & 88.8 & 96.2 & 95.2 & 68.2 & \textbf{82.8} & \underline{78.9} \\

& MMA \cite{yang2024mma} & 68.0 & 74.0 & 34.0 & 80.1 & 73.5 & 81.4 & \textbf{91.5} & 94.3 & \underline{95.6} & \underline{68.9} & 81.7 & 76.7 \\

\rowcolor{lncolor}
\cellcolor{gray!0} & \ours & \textbf{68.4} & \textbf{74.8} & \underline{40.2} & \textbf{92.1} & \textbf{80.2} & 80.8 & 90.3 & \textbf{96.3} & \textbf{95.8} & \textbf{70.4} & \underline{82.3} & \textbf{79.2} \\
 \\ [-5ex] \\ 
    \cmidrule(lr){1-14}
    
    \cellcolor{gray!0}\multirow{13}{*}{\texttt{ViT-L/14}} 
& \emph{Zero-Shot} & 72.9 & 67.6 & 32.6 & 58.0 & 76.8 & 91.0 & 93.6 & 79.4 & 94.9 & 53.6 & 74.2 & 72.2 \\

& CoOp \cite{zhou2022coop} (ctx=16) & 78.2 & 77.5 & 55.2 & 88.3 & 89.0 & 89.8 & 94.6 & \textbf{99.1} & 97.2 & 74.4 & 87.3 & 84.6 \\

& CoCoOp \cite{zhou2022cocoop} & 77.8 & 76.7 & 45.2 & 79.8 & 82.7 & 91.9 & \underline{95.4} & 95.3 & 97.4 & 71.4 & 85.2 & 81.7 \\

& TIP-Adapter-F \cite{zhang2021tip} & 79.3 & 79.6 & 55.8 & 86.1 & 88.1 & 91.6 & 94.6 & 98.3 & \underline{97.5} & 74.0 & 87.4 & 84.8 \\

& CLIP-Adapter \cite{gao2024clip} & 76.4 & 78.0 & 46.4 & 75.8 & 83.8 & 91.6 & 94.3 & 97.3 & 97.3 & 71.3 & 86.1 & 81.7 \\

& PLOT++ \cite{chen2022plot} & 78.6 & 79.1 & 44.1 & 92.2 & 87.2 & 90.2 & 93.6 & 98.8 & \underline{97.5} & 75.0 & 87.1 & 83.9 \\

& KgCoOp \cite{yao2023visual} & 76.8 & 76.7 & 47.5 & 83.6 & 83.2 & 91.7 & 95.3 & 96.4 & 97.4 & 73.6 & 86.4 & 82.6 \\

& TaskRes \cite{yu2023task} & 78.1 & 76.9 & 55.0 & 84.3 & 87.6 & 91.5 & 94.7 & 97.8 & 97.3 & 74.4 & 86.6 & 84.0 \\

& MaPLe \cite{khattak2023maple} & 78.4 & 78.8 & 46.3 & 85.4 & 83.6 & \textbf{92.0} & 95.4 & 97.4 & 97.2 & 72.7 & 86.5 & 83.1 \\

& ProGrad \cite{zhu2023prograd} & 78.4 & 78.3 & 55.6 & 89.3 & 88.8 & 90.8 & 94.9 & 98.7 & \underline{97.5} & 73.7 & 87.7 & 84.9 \\

& LP++ \cite{huang2024lp++} & 79.3 & 79.7 & 54.6 & 89.3 & 87.7 & 91.7 & 94.9 & 98.5 & 97.4 & 76.1 & 88.1 & 85.2 \\

& CLIP-LoRA \cite{zanella2024cliplora} & \underline{79.6} & 79.4 & \textbf{66.2} & \textbf{93.1} & \textbf{90.9} & 89.9 & 94.3 & 99.0 & 97.3 & \underline{76.5} & \textbf{89.9} & \underline{86.9} \\

& MMA \cite{yang2024mma} & \textbf{79.9} & \underline{80.2} & 56.4 & 76.3 & 88.0 & \textbf{92.0} & \textbf{95.5} & 98.4 & \textbf{97.6} & 75.8 & 88.0 & 84.4 \\

\rowcolor{lncolor}
\cellcolor{gray!0} & \ours & 79.4 & \textbf{80.3} & \underline{64.1} & \underline{92.9} & \underline{90.3} & 91.1 & \textbf{95.5} & \textbf{99.1} & \underline{97.5} & \textbf{78.0} & \underline{89.5} & \textbf{87.1} \\
            
\bottomrule

    \end{tabular}
    \end{adjustbox}
\label{tab:all2all}
\end{table*}

\section{Experiments}
\label{sec:experiments}
This section highlights the experimental results of \ours, evaluating: (i) \emph{all-to-all} FSA, in which train/test categories coincide (\ie, $\baseClasses = \genClasses$) and (ii) \emph{base-to-novel} generalization, where each task further spans a set $\novelClasses$ of unseen categories. 

\mypar{Datasets.} For both settings, we consider a suite of 11 benchmarks.
Specifically, these include ImageNet \cite{russakovsky2015imagenet}, Caltech101 (CAL) \cite{fei2006one}, SUN397 \cite{xiao2010sun}, Describable Textures (DTD) \cite{cimpoi14describing}, FGVC Aircraft (AIR) \cite{maji13fine-grained}, Oxford Pets (PETS) \cite{parkhi12a}, Oxford Flowers 102 (FLWR) \cite{nilsback2008automated}, Stanford Cars (CARS) \cite{krause20133d}, Food-101 (FOOD) \cite{bossard2014food}, EuroSAT (ESAT) \cite{helber2019eurosat} and UCF-101 (UCF) \cite{soomro2012ucf101}. 
We use the splits of \cite{zhou2022cocoop}.

\mypar{Baselines.}
In \emph{base-to-novel} generalization, we consider a total of 8 comparative methods: CoOp \cite{zhou2022coop} and CoCoOp \cite{zhou2022cocoop} as established baselines of the field; ProGrad \cite{zhu2023prograd}, KgCoOp \cite{yao2023visual} and MaPLe \cite{khattak2023maple} as modern advancements; Multi-Modal Adapter (MMA) \cite{yang2024mma} and CLIP-LoRA \cite{zanella2024cliplora} as the strongest and most recent strategies.
Zero-shot performance using hand-crafted templates is also reported as a reference.
In the \emph{all-to-all} setup, we expand the suite to a total of 12 methods, with 5 additional strategies that are \emph{inapplicable} to unseen categories: TIP-Adapter-F \cite{zhang2021tip}, CLIP-Adapter \cite{gao2024clip}, PLOT++ \cite{chen2022plot}, TaskRes \cite{yu2023task} and LP++ \cite{huang2024lp++}.

\noindent\textbf{Implementation Details.} Following CLIP-LoRA \cite{zanella2024cliplora}, we express the number of iterations as $\niter = M \times k$, where $k$ is the number of shots and $M \in \mathbb{N}$ is a hyperparameter.
We also inherit the optimizer setup, using AdamW \cite{loshchilovdecoupled} with a learning rate of $2\times10^{-4}$, a weight decay factor of $0.01$ and a mini-batch size of 32. 
We do \emph{not} use dataset-specific templates, but we format all categories via the generic template ``\texttt{a photo of a \{\}}''.
For all settings and competitors, we report numbers from the original works or from \cite{zanella2024cliplora}. 
When unavailable, we reproduce the results using the official code.
Results are averaged over 3 different runs.

\mypar{Shots configurations.} Due to space constraints, here we only report results with $k{=}16$ available shots. 
Complementary experiments with $k\in\{4,8\}$ are in \cref{app:shots}.

\mypar{Fixed hyperparameters.}
For a realistic evaluation, we fix $M$ and $\alpha$ by tuning on ImageNet \cite{russakovsky2015imagenet} with the ViT-B/16 backbone and $k=16$ shots only. 
Search intervals are defined as $M\in\{100,300,500\}$ and $\alpha\in[0.2,0.8]$ (using a coarse step size of $0.1$).
We obtain values of $M=300$ and $\alpha=0.6$, which are always fixed unless stated otherwise.

\subsection{Base-to-novel generalization}
\label{sec:base2novel}
For this setting, we follow recent works \cite{khattak2023maple, yang2024mma} and experiment with the ViT-B/16 backbone and $k=16$ shots.
The evaluation metrics are threefold: (i) top-1 accuracy on \emph{base} categories, (ii) top-1 accuracy on \emph{novel} categories, and (iii) the harmonic mean between them.

\mypar{Results} for all datasets are reported in \cref{tab:b2n}, where (i), (ii), and (iii) are denoted as \textbf{Base}, \textbf{Novel} and \textbf{HM}, respectively.
\ours provides a greater harmonic mean than all competitors in 8 out of 11 benchmarks, and outperforms all competitors on average across datasets.
Notable success cases are Stanford Cars \cite{krause20133d} and FGVC Aircraft \cite{maji13fine-grained}, with margins in the harmonic mean of $+2.76\%$ and $+2.30\%$, while the worst case is EuroSAT.
We attribute this to the tiny size of the dataset, as there are only 5 base categories, which leads \ours to exit the first stage when $\paramsSpLNOptim$ are disrupted already.
We verify this hypothesis by simply decreasing the batch size, hence collecting gradients for fewer examples in the first stage, observing notable performance improvements (\ie, HM of $84.32\%$ with a batch size of 1).

\subsection{All-to-all Few-Shot Adaptation}
\label{sec:all2all}
We further validate \ours in the all-to-all setup, where train/test categories coincide, \ie, $\baseClasses=\genClasses$.
Intuitively, strong performance in this setting should support the hypothesis that the first stage learns good task-level features since a linear classifier is trained to separate them.
Here, note that we have $\classifier_\genClass^*$ for all categories, hence selective inference \emph{always} skips the computations within the text encoder.

\mypar{Setup details.} We follow recent work \cite{zanella2024cliplora}, and evaluate \ours across a variety of 3 different backbones: ViT-B/16, ViT-B/32 and ViT-L/14.
We keep hyperparameters fixed.

\mypar{Results} are given in \cref{tab:all2all}, confirming the hypothesis that a simple linear classifier, trained on top of frozen task-level features obtained with $\paramsSpLNOptim$, is sufficient to match or surpass state-of-the-art methods. 
The best competitor in this setting is definitely CLIP-LoRA \cite{zanella2024cliplora}, which, nevertheless, is outperformed by \ours on average across backbones.
We emphasize that \textbf{\ours exhibits stable performance across settings}, a missing trait in both CLIP-LoRA and MMA \cite{yang2024mma}.
\ours outperforms the former by $+2.9\%$ in base-to-novel settings, and surpasses the latter by $+1.9\%$, $+2.5\%$ and $+2.7\%$ in the all-to-all setup for the three backbones.

\begin{figure}[t!]
    \centering
    \includegraphics[width=\linewidth]{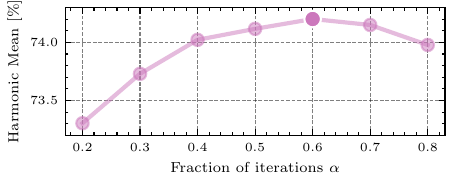}
    \caption{Hyperparameter sweep on $\alpha\in[0.2,0.8]$ with a step size of $0.1$, conducted on ImageNet \cite{russakovsky2015imagenet}. The optimal value $\alpha=0.6$ is transferred to every other experiment of this work.}
    \label{fig:alpha-sweep}
    \vspace{-1em}
\end{figure}
\begin{figure}[t!]
    \centering
    \includegraphics[width=\linewidth]{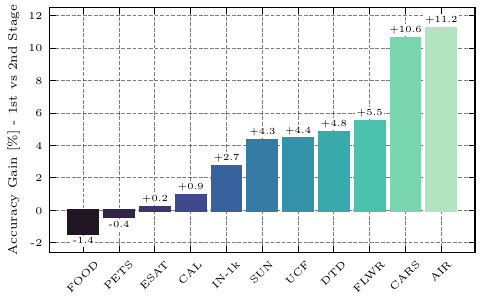}
    \caption{Visualization of the absolute accuracy improvement across the 11 benchmarks in the \emph{all-to-all} setting, relative to exiting after the first stage, \ie, relative to using $\paramsSpLNOptim$ only. }
    \label{fig:1v2ablation}
    \vspace{-1em}
\end{figure}

\subsection{Analysis}
\label{sec:post-analysis}
We conclude the experimental section by (i) reporting the sweep conducted on $\alpha$ and (ii) highlighting the benefits of \emph{switching} parameters in the second stage.

\mypar{Selection of $\alpha$.} \cref{fig:alpha-sweep} reports the outcome of the hyperparameter sweep on the fraction of iterations $\alpha$.
We conduct this search for the \emph{base-to-novel} setting using the validation set of ImageNet \cite{russakovsky2015imagenet} and report the Harmonic Mean corresponding to each $\alpha$ value in $[0.2, 0.8]$, with a step size of $0.1$. 
We use the CLIP ViT-B/16 model and $k{=}16$ shots.
In line with the evidence of \cref{sec:pre-analysis}, an optimal value $\alpha{=}0.6$ emerges, preceded and followed by decreased performance.

\mypar{The benefit of having two stages.}
\cref{fig:1v2ablation} compares \ours{} relative to only optimizing LayerNorm instances, showing that large margins are obtained on average (\eg, more than ${+}10\%$ improvement on Stanford Cars and FGVC Aircraft, and ${+}4\%$ margin for SUN397, UCF-101, DTD, and Oxford Flowers further).
We spot failure cases in Food-101 and Oxford Pets, which we examine in detail in \cref{app:twostages}.

\section{Conclusions}
\label{sec:conclusions}
In this work, we investigated the dynamics of PEFT techniques in the FSA of VLMs. 
We empirically showed that PEFT learns good task-level features, even without \textit{ad-hoc} mechanisms promoting cross-modal generalization, and that PEFT techniques exhibit different degrees of robustness to unseen categories. 
Through experiments on 11 benchmarks with fixed hyperparameters, we showed that training a linear classifier on top of frozen PEFT features, a scheme we call \ours{}, matches or outperforms the established state-of-the-art in different settings, thereby serving as a simple unified recipe. 
We hope our work will be useful for future research on VLM transfer.

\clearpage
\paragraph{Acknowledgements.} The authors acknowledge the CINECA award under the ISCRA initiative for the availability of high-performance computing resources and support. 
M.F. is supported by the PRIN B-FAIR (Prot. 2022EX F3HX) project and the PAT project ``AI@TN''. 
This work was supported by the projects EU Horizon ELIAS (No. 101120237), AI4TRUST (No.101070190), FAIR - Future AI Research (PE00000013), funded by NextGeneration EU, and carried out in the Vision and Learning joint laboratory of Fondazione Bruno Kessler and the University of Trento, Italy.

{
    \small
    \bibliographystyle{ieeenat_fullname}
    \bibliography{main}
}

\clearpage
\maketitlesupplementary
\appendix

This Supplementary Material aims to expand and complement the work's main body. 
We structure it as follows: 
\begin{itemize}
    \item \textbf{Generalization to other PEFT techniques.} \cref{app:other-peft} shows that the benefit of the two-stage design of \ours{} is not limited to Layer Normalization. 
    In \cref{app:lora} we experiment with LoRA, with particular emphasis on the relationship with CLIP-LoRA \cite{zanella2024cliplora}; \cref{app:prompt_learning} experiments with Prompt Learning techniques: CoOp \cite{zhou2022coop} and ``Independent Vision-Language Prompting'' (IVLP);
    \item \textbf{Additional visual backbones.} \cref{app:b2nbackbones} complements \cref{sec:base2novel}, by reporting results in the \emph{base-to-novel} setting also for the ViT-B/32 and ViT-L/14 backbones;
    \item \textbf{Robustness to shots availability.} \cref{app:shots} reports the results of the main paper, with varying numbers of available shots. We further experiment with $k{=}\{4,8\}$ shots; 
    \item \textbf{Hyperparameter analysis.} \cref{app:iteranalysis} analyzes the impact of the total number of allowed iterations $m$;
    \item \textbf{Extended preliminary analysis.} \cref{app:twostages} reports additional evidence for the preliminary observations of \cref{sec:pre-analysis}, with a focus on identifying the ``best'' and ``worst'' cases.
    We devote explanations for the slight failures of \cref{fig:1v2ablation} with Oxford Pets \cite{parkhi12a} and Food-101 \cite{bossard2014food};
    \item \textbf{Limitations} are discussed in \cref{app:limitations}.
\end{itemize}

\section{\ours{} with different PEFT techniques}\label{app:other-peft}
\subsection{LoRA}
\label{app:lora}
We expand \cref{sec:experiments} reporting results of \cref{tab:b2n} and \cref{tab:all2all} when LoRA \cite{hu2022lora} is applied in the first stage.
Note that \ours{}\textsubscript{\texttt{LoRA}} can be seen as a two-stage variant of the recently introduced CLIP-LoRA strategy, hence we are particularly interested in understanding the benefits, if any, relative to it.

\mypar{Implementation details.} Following \cref{sec:experiments}, we perform a sweep for $\alpha$ in $[0.2,0.8]$ with a step size of $0.1$, only for the base-to-novel setting with the ViT-B/16 backbone, validating on ImageNet.
We obtain an optimal value of $\alpha=0.3$, which aligns with the behavior highlighted by \cref{fig:breakpoint} (\ie, LoRA tends to saturate more quickly than LayerNorm). 
We transfer it to all other experiments in this Supplementary Material when \ours{}\textsubscript{\texttt{LoRA}} is specified.
We strictly follow the CLIP-LoRA recipe for plugging low-rank modules in CLIP.

\subsubsection{Base-to-novel generalization}
\begin{table*}[ht!]
\caption{Experiments in \emph{base-to-novel} generalization with the ViT-B/16 visual backbone. All methods use $k{=}16$ shots per base class. ``CLIP'' refers to zero-shot performance with dataset-specific templates, \eg, ``\emph{a photo of a \{\}, a type of flower}'' for Oxford Flowers. To highlight the benefits of the two-stage design, we report an additional line with the absolute improvement of \ours{}\textsubscript{LoRA} relative to its single-stage counterpart CLIP-LoRA \cite{zanella2024cliplora}. In each table, the best performer is \textbf{bold}, and the second best is \underline{underlined}.}
\label{tab:b2nlora}
    \def\arraystretch{1.075}
    \centering
    \scriptsize
    \begin{minipage}{0.32\textwidth}
        \centering
        \caption*{\textbf{Average across datasets.}}\vspace{-1em}
        \begin{tabular}{lcc|c}
        \toprule
         \textbf{Method} & \textbf{Base} & \textbf{Novel} & \textbf{HM} \\
         \midrule
         CLIP \cite{radford2021learning} & 69.34 & 74.22 & 71.70 \\
         CoOP \cite{zhou2022coop} & 82.69 & 63.22 & 71.66 \\
         CoCoOp \cite{zhou2022cocoop} & 80.47 & 71.69 & 75.83 \\
         MaPLe \cite{khattak2023maple} & 82.28 & 75.14 & 78.55 \\
         ProGrad \cite{zhu2023prograd} & 82.48 & 70.75 & 76.16 \\
         KgCoOp \cite{yao2023visual} & 80.73 & 73.60 & 77.00 \\
         CLIP-LoRA \cite{zanella2024cliplora} & 85.32 & 70.63 & 77.28 \\
         MMA \cite{yang2024mma} & 83.20 & 76.80 & \underline{79.87} \\
         \cmidrule(lr){1-4}
         \rowcolor{lncolor}
         \ours{}\textsubscript{\texttt{LayerNorm}} & 85.55 & 75.48 & \textbf{80.20} \\
         \rowcolor{lncolor}
         \ours{}\textsubscript{\texttt{LoRA}} & 85.97 & 73.65 & 79.33 \\
         \rowcolor{lncolor}
         & $+$0.65 & $+$3.02 & $+$2.05 \\
         \bottomrule
        \end{tabular}
        \vspace{1em}
    \end{minipage}%
    \hfill
    \centering
    \begin{minipage}{0.32\textwidth}
        \centering
        \caption*{ImageNet}\vspace{-1em}
        \begin{tabular}{lcc|c}
        \toprule
         \textbf{Method} & \textbf{Base} & \textbf{Novel} & \textbf{HM} \\
         \midrule
         CLIP \cite{radford2021learning} & 72.43 & 68.14 & 70.22 \\
         CoOp \cite{zhou2022coop} & 76.47 & 67.88 & 71.92 \\
         CoCoOp \cite{zhou2022cocoop} & 75.98 & 70.43 & 73.10 \\
         MaPLe \cite{khattak2023maple} & 76.66 & 70.54 & 73.47 \\
         ProGrad \cite{zhu2023prograd} & 77.02 & 66.66 & 71.46 \\
         KgCoOp \cite{yao2023visual} & 75.83 & 69.96 & 72.78 \\
         CLIP-LoRA \cite{zanella2024cliplora} & 77.58 & 68.76 & 72.91 \\
         MMA \cite{yang2024mma} & 77.31 & 71.00 & 74.02 \\
         \cmidrule(lr){1-4}
         \rowcolor{lncolor}
         \ours{}\textsubscript{\texttt{LayerNorm}} & 77.71 & 70.99 & \underline{74.20}  \\
         \rowcolor{lncolor}
         \ours{}\textsubscript{\texttt{LoRA}} & 77.70 & 71.60 & \textbf{74.53} \\
         \rowcolor{lncolor}
         & $+$0.12 & $+$2.84 & $+$1.62 \\
         \bottomrule
        \end{tabular}
        \vspace{1em}
    \end{minipage}%
    \hfill
    \centering
    \begin{minipage}{0.32\textwidth}
        \centering
        \caption*{Caltech101}\vspace{-1em}
        \begin{tabular}{lcc|c}
        \toprule
         \textbf{Method} & \textbf{Base} & \textbf{Novel} & \textbf{HM} \\
         \midrule
         CLIP \cite{radford2021learning} & 96.84 & 94.00 & 95.40 \\
         CoOP \cite{zhou2022coop} & 98.00 & 89.81 & 93.73 \\
         CoCoOp \cite{zhou2022cocoop} & 97.96 & 93.81 & 95.84 \\
         MaPLe \cite{khattak2023maple} & 97.74 & 94.36 & 96.02 \\
         ProGrad \cite{zhu2023prograd} & 98.02 & 93.89 & 95.91 \\
         KgCoOp \cite{yao2023visual} & 97.72 & 94.39 & 96.03 \\
         CLIP-LoRA \cite{zanella2024cliplora} & 98.19 & 93.05 & 95.55 \\
         MMA \cite{yang2024mma} & 98.40 & 94.00 & 96.15 \\
         \cmidrule(lr){1-4}
         \rowcolor{lncolor}
         \ours{}\textsubscript{\texttt{LayerNorm}} & 98.71 & 94.43 & \textbf{96.52} \\
         \rowcolor{lncolor}
         \ours{}\textsubscript{\texttt{LoRA}} & 98.45 & 94.47 & \underline{96.42} \\
         \rowcolor{lncolor}
         & $+$0.26 & $+$1.42 & $+$0.87 \\
         \bottomrule
        \end{tabular}
        \vspace{1em}
    \end{minipage}%
    \hfill
    \begin{minipage}{0.32\textwidth}
        \centering
        \caption*{Oxford Flowers}\vspace{-1em}
        \begin{tabular}{lcc|c}
        \toprule
         \textbf{Method} & \textbf{Base} & \textbf{Novel} & \textbf{HM} \\
         \midrule
         CLIP \cite{radford2021learning} & 72.08 & 77.80 & 74.83 \\
         CoOP \cite{zhou2022coop} & 97.60 & 59.67 & 74.06 \\
         CoCoOp \cite{zhou2022cocoop} & 94.87 & 71.75 & 81.71 \\
         MaPLe \cite{khattak2023maple} & 95.92 & 72.46 & 82.56 \\
         ProGrad \cite{zhu2023prograd} & 95.54 & 71.87 & 82.03 \\
         KgCoOp \cite{yao2023visual} & 95.00 & 74.73 & 83.65 \\ 
         CLIP-LoRA \cite{zanella2024cliplora} & 97.91 & 68.61 & 80.68 \\
         MMA \cite{yang2024mma} & 97.77 & 75.93 & \underline{85.48} \\
         \cmidrule(lr){1-4}
         \rowcolor{lncolor}
         \ours{}\textsubscript{\texttt{LayerNorm}} & 98.29 & 76.17 & \textbf{85.83} \\
         \rowcolor{lncolor}
         \ours{}\textsubscript{\texttt{LoRA}} & 98.04 & 70.95 & 82.32 \\
         \rowcolor{lncolor}
         & $+$0.13 & $+$2.34 & $+$1.64 \\
         \bottomrule
        \end{tabular}
        \vspace{1em}
    \end{minipage}
    \hfill
    \begin{minipage}{0.32\textwidth}
        \centering
        \caption*{Oxford Pets}\vspace{-1em}
        \begin{tabular}{lcc|c}
        \toprule
         \textbf{Method} & \textbf{Base} & \textbf{Novel} & \textbf{HM} \\
         \midrule
         CLIP \cite{radford2021learning} & 91.17 & 97.26 & 94.12 \\
         CoOP \cite{zhou2022coop} &  93.67 & 95.29 & 94.47 \\
         CoCoOp \cite{zhou2022cocoop} & 95.20 & 97.69 & 96.43 \\
         MaPLe \cite{khattak2023maple} & 95.43 & 97.76 & \underline{96.58} \\
         ProGrad \cite{zhu2023prograd} & 95.07 & 97.63 & 96.33 \\
         KgCoOp \cite{yao2023visual} & 94.65 & 97.76 & 96.18 \\
         CLIP-LoRA \cite{zanella2024cliplora} & 94.36 & 95.71 & 95.03 \\
         MMA \cite{yang2024mma} & 95.40 & 98.07 & \textbf{96.72} \\
         \cmidrule(lr){1-4}
         \rowcolor{lncolor}
         \ours{}\textsubscript{\texttt{LayerNorm}} & 95.32 & 97.82 & 96.55 \\
         \rowcolor{lncolor}
         \ours{}\textsubscript{\texttt{LoRA}} & 95.50 & 97.13 & 96.31 \\
         \rowcolor{lncolor}
         & $+$1.14 & $+$1.42 & $+$1.28 \\
         \bottomrule
        \end{tabular}
        \vspace{1em}
    \end{minipage}%
    \hfill
    \begin{minipage}{0.32\textwidth}
        \centering
        \caption*{Stanford Cars}\vspace{-1em}
        \begin{tabular}{lcc|c}
        \toprule
         \textbf{Method} & \textbf{Base} & \textbf{Novel} & \textbf{HM} \\
         \midrule
         CLIP \cite{radford2021learning} & 63.37 & 74.89 & 68.65 \\
         CoOP \cite{zhou2022coop} & 78.12 & 60.40 & 68.13 \\
         CoCoOp \cite{zhou2022cocoop} & 70.49 & 73.59 & 72.01 \\
         MaPLe \cite{khattak2023maple} & 72.94 & 74.00 & 73.47 \\
         ProGrad \cite{zhu2023prograd} & 77.68 & 68.63 & 72.88 \\
         KgCoOp \cite{yao2023visual} & 71.76 & 75.04 & 73.36 \\
         CLIP-LoRA \cite{zanella2024cliplora} & 83.93 & 65.54 & 73.60 \\
         MMA \cite{yang2024mma} & 78.50 & 73.10 & 75.70 \\
         \cmidrule(lr){1-4}
         \rowcolor{lncolor}
         \ours{}\textsubscript{\texttt{LayerNorm}} & 82.50 & 74.80 & \textbf{78.46} \\
         \rowcolor{lncolor}
         \ours{}\textsubscript{\texttt{LoRA}} & 83.87 & 70.64 & \underline{76.69} \\
         \rowcolor{lncolor}
         & $-$0.06 & $+$5.10 & $+$3.09 \\
         \bottomrule
        \end{tabular}
        \vspace{1em}
    \end{minipage}%
    \hfill
    \begin{minipage}{0.32\textwidth}
        \centering
        \caption*{Food 101}\vspace{-1em}
        \begin{tabular}{lcc|c}
        \toprule
         \textbf{Method} & \textbf{Base} & \textbf{Novel} & \textbf{HM} \\
         \midrule
         CLIP \cite{radford2021learning} & 90.10 & 91.22 & 90.66 \\
         CoOP \cite{zhou2022coop} & 88.33 & 82.26 & 85.19 \\
         CoCoOp \cite{zhou2022cocoop} & 90.70 & 91.29 & 90.99 \\
         MaPLe \cite{khattak2023maple} & 90.71 & 92.05 & \textbf{91.38} \\
         ProGrad \cite{zhu2023prograd} & 90.37 & 89.59 & 89.98 \\
         KgCoOp \cite{yao2023visual} & 90.50 & 91.70 & \underline{91.09} \\
         CLIP-LoRA \cite{zanella2024cliplora} & 86.84 & 86.67 & 86.76 \\
         MMA \cite{yang2024mma} & 90.13 & 91.30 & 90.71 \\
         \cmidrule(lr){1-4}
         \rowcolor{lncolor}
         \ours{}\textsubscript{\texttt{LayerNorm}} & 89.11 & 91.34 & 90.21 \\
         \rowcolor{lncolor}
         \ours{}\textsubscript{\texttt{LoRA}} & 88.47 & 89.96 & 89.21 \\
         \rowcolor{lncolor}
         & $+$1.61 & $+$3.29 & $+$2.45 \\ 
         \bottomrule
        \end{tabular}
        \vspace{1em}
    \end{minipage}
    \hfill
    \begin{minipage}{0.32\textwidth}
        \centering
        \caption*{FGVC Aircraft}\vspace{-1em}
        \begin{tabular}{lcc|c}
        \toprule
         \textbf{Method} & \textbf{Base} & \textbf{Novel} & \textbf{HM} \\
         \midrule
         CLIP \cite{radford2021learning} & 27.19 & 36.29 & 31.09 \\
         CoOP \cite{zhou2022coop} & 40.44 & 22.30 & 28.75 \\
         CoCoOp \cite{zhou2022cocoop} & 33.41 & 23.71 & 27.74 \\
         MaPLe \cite{khattak2023maple} & 37.44 & 35.61 & 36.50 \\
         ProGrad \cite{zhu2023prograd} & 40.54 & 27.57 & 32.82 \\
         KgCoOp \cite{yao2023visual} & 36.21 & 33.55 & 34.83 \\
         CLIP-LoRA \cite{zanella2024cliplora} & 50.10 & 26.03 & 34.26 \\
         MMA \cite{yang2024mma} & 40.57 & 36.33 & 38.33 \\
         \cmidrule(lr){1-4}
         \rowcolor{lncolor}
         \ours{}\textsubscript{\texttt{LayerNorm}} & 47.48 & 35.51 & \textbf{40.63} \\
         \rowcolor{lncolor}
         \ours{}\textsubscript{\texttt{LoRA}} & 51.00 & 31.37 & \underline{38.85} \\
         \rowcolor{lncolor}
         & $+$0.90 & $+$5.34 & $+$ 4.59 \\
         \bottomrule
        \end{tabular}
        \vspace{1em}
    \end{minipage}%
    \hfill
    \begin{minipage}{0.32\textwidth}
        \centering
        \caption*{SUN 397}\vspace{-1em}
        \begin{tabular}{lcc|c}
        \toprule
         \textbf{Method} & \textbf{Base} & \textbf{Novel} & \textbf{HM} \\
         \midrule
         CLIP \cite{radford2021learning} & 69.36 & 75.35 & 72.23 \\
         CoOP \cite{zhou2022coop} & 80.60 & 65.89 & 72.51 \\
         CoCoOp \cite{zhou2022cocoop} & 79.74 & 76.86 & 78.27 \\
         MaPLe \cite{khattak2023maple} & 80.82 & 78.70 & 79.75 \\
         ProGrad \cite{zhu2023prograd} & 81.26 & 74.17 & 77.55 \\
         KgCoOp \cite{yao2023visual} & 80.29 & 76.53 & 78.36 \\
         CLIP-LoRA \cite{zanella2024cliplora} & 81.11 & 74.53 & 77.68 \\
         MMA \cite{yang2024mma} & 82.27 & 78.57 & 80.38 \\
         \cmidrule(lr){1-4}
         \rowcolor{lncolor}
         \ours{}\textsubscript{\texttt{LayerNorm}} & 82.59 & 78.91 & \textbf{80.70} \\
         \rowcolor{lncolor}
         \ours{}\textsubscript{\texttt{LoRA}} & 82.43 & 79.05 & \textbf{80.70} \\
         \rowcolor{lncolor}
         & $+$1.32 & $+$4.52 & $+$3.02 \\
         \bottomrule
        \end{tabular}
        \vspace{1em}
    \end{minipage}
    \hfill
    \begin{minipage}{0.32\textwidth}
        \centering
        \caption*{DTD}\vspace{-1em}
        \begin{tabular}{lcc|c}
        \toprule
         \textbf{Method} & \textbf{Base} & \textbf{Novel} & \textbf{HM} \\
         \midrule
         CLIP \cite{radford2021learning} & 53.24 & 59.90 & 56.37 \\
         CoOP \cite{zhou2022coop} & 79.44 & 41.18 & 54.24 \\
         CoCoOp \cite{zhou2022cocoop} & 77.01 & 56.00 & 64.85 \\
         MaPLe \cite{khattak2023maple} & 80.36 & 59.18 & 68.16 \\
         ProGrad \cite{zhu2023prograd} & 77.35 & 52.35 & 62.45 \\
         KgCoOp \cite{yao2023visual} & 77.55 & 54.99 & 64.35 \\
         CLIP-LoRA \cite{zanella2024cliplora} & 83.95 & 62.84 & 71.39 \\
         MMA \cite{yang2024mma} & 83.20 & 65.63 & \underline{73.38} \\
         \cmidrule(lr){1-4}
         \rowcolor{lncolor}
         \ours{}\textsubscript{\texttt{LayerNorm}} & 84.60 & 65.01 & \textbf{73.52} \\
         \rowcolor{lncolor}
         \ours{}\textsubscript{\texttt{LoRA}} & 84.53 & 63.53 & 72.54 \\
         \rowcolor{lncolor}
         & $+$0.58 & $+$0.69 & $+$1.15 \\
         \bottomrule
        \end{tabular}
    \end{minipage}%
    \hfill
    \begin{minipage}{0.32\textwidth}
        \centering
        \caption*{EuroSAT}\vspace{-1em}
        \begin{tabular}{lcc|c}
        \toprule
         \textbf{Method} & \textbf{Base} & \textbf{Novel} & \textbf{HM} \\
         \midrule
         CLIP \cite{radford2021learning} & 56.48 & 64.05 & 60.03 \\
         CoOP \cite{zhou2022coop} & 92.19 & 54.74 & 68.69\\
         CoCoOp \cite{zhou2022cocoop} & 87.49 & 60.04 & 71.21 \\
         MaPLe \cite{khattak2023maple} & 94.07 & 73.23 & \underline{82.35} \\
         ProGrad \cite{zhu2023prograd} & 90.11 & 60.89 & 72.67 \\
         KgCoOp \cite{yao2023visual} & 85.64 & 64.34 & 73.48 \\
         CLIP-LoRA \cite{zanella2024cliplora} & 97.04 & 62.50 & 76.03 \\
         MMA \cite{yang2024mma} & 85.46 & 82.34 & \textbf{83.87} \\
         \cmidrule(lr){1-4}
         \rowcolor{lncolor}
         \ours{}\textsubscript{\texttt{LayerNorm}} & 96.91 & 67.09 & 79.29 \\
         \rowcolor{lncolor}
         \ours{}\textsubscript{\texttt{LoRA}} & 97.05 & 64.59 & 77.56 \\
         \rowcolor{lncolor}
         & $+$0.01 & $+$2.09 & $+$1.53 \\ 
         \bottomrule
        \end{tabular}
    \end{minipage}
    \hfill
    \begin{minipage}{0.32\textwidth}
        \centering
        \caption*{UCF101}\vspace{-1em}
        \begin{tabular}{lcc|c}
        \toprule
         \textbf{Method} & \textbf{Base} & \textbf{Novel} & \textbf{HM} \\
         \midrule
         CLIP \cite{radford2021learning} & 70.53 & 77.50 & 73.85 \\
         CoOP \cite{zhou2022coop} & 84.69 & 56.05 & 67.46 \\
         CoCoOp \cite{zhou2022cocoop} & 82.33 & 73.45 & 77.64 \\
         MaPLe \cite{khattak2023maple} & 83.00 & 78.66 & 80.77 \\
         ProGrad \cite{zhu2023prograd} & 84.33 & 74.94 & 79.35 \\
         KgCoOp \cite{yao2023visual} & 82.89 & 76.67 & 79.65 \\
         CLIP-LoRA \cite{zanella2024cliplora} & 87.52 & 72.74 & 79.45 \\
         MMA \cite{yang2024mma} & 86.23 & 80.03 & 82.20 \\
         \cmidrule(lr){1-4}
         \rowcolor{lncolor}
         \ours{}\textsubscript{\texttt{LayerNorm}} & 87.85 & 78.19 & \textbf{82.74} \\
         \rowcolor{lncolor}
         \ours{}\textsubscript{\texttt{LoRA}} & 88.59 & 76.82 & \underline{82.28} \\
         \rowcolor{lncolor}
         & $+$1.07 & $+$4.08 & $+$2.83 \\
         \bottomrule
        \end{tabular}
    \end{minipage}
\end{table*}

Extended results for the base-to-novel setup are given in \cref{tab:b2nlora}.
We observe that, \wrt CLIP-LoRA, \ourslora{} provides a notable improvement, especially pronounced in the \textbf{Novel} metric ($+3.02\%$ on average) and, more in general, it boosts performance on 32 out of 33 \{dataset, metric\} combinations. 
Overall, these results take a step further in confirming the hypothesis that the two-stage design is beneficial for other PEFT strategies, since \ourslora{} exhibits the 2nd greatest Harmonic Mean on average, only outperformed by MMA \cite{yang2024mma} among published works (excluding our own alternative \oursln).

\subsubsection{All-to-all adaptation}
\begin{table*}[t!]
    \def\arraystretch{1.1}
    \centering
    \footnotesize
    \caption{\emph{All-to-all} experiments, where train/test categories coincide, with the ViT-B/16 (top), ViT-B/32 (middle), and ViT-L/14 (bottom) backbones. All methods use $k=16$ shots per class. To highlight the benefits of the two-stage design, we report an additional line with the absolute improvement of \ours{}\textsubscript{LoRA} relative to its single-stage counterpart CLIP-LoRA \cite{zanella2024cliplora}. In each group, the best performer is marked by \textbf{bold text}; the second best is \underline{underlined}.}
    \begin{adjustbox}{max width=\textwidth}
    \begin{tabular}{llccccccccccc|c}
    \toprule
     \textsc{\textbf{Backbone}} &  \textsc{\textbf{Method}} &  \textsc{\textbf{ImageNet}} &  \textsc{\textbf{SUN}} &  \textsc{\textbf{AIR}} &  \textsc{\textbf{ESAT}} &  \textsc{\textbf{CARS}} &  \textsc{\textbf{FOOD}} &  \textsc{\textbf{PETS}} &  \textsc{\textbf{FLWR}} &  \textsc{\textbf{CAL}} &  \textsc{\textbf{DTD}} &  \textsc{\textbf{UCF}} &  \textsc{\textbf{Mean}} \\
    \midrule
    
    \cellcolor{gray!0} \multirow{13}{*}{\texttt{ViT-B/16}} 
& \emph{Zero-Shot} &  66.7 & 62.6 & 24.7 & 47.5 & 65.3 & 86.1 & 89.1 & 71.4 & 92.9 & 43.6 & 66.7 & 65.1 \\

& CoOp \cite{zhou2022coop} (ctx=16) &  71.9 & 74.9 & 43.2 & 85.0 & 82.9 & 84.2 & 92.0 & 96.8 & 95.8 & 69.7 & 83.1 & 80.0 \\

& CoCoOp \cite{zhou2022cocoop} &  71.1 & 72.6 & 33.3 & 73.6 & 72.3 & \textbf{87.4} & 93.4 & 89.1 & 95.1 & 63.7 & 77.2 & 75.4 \\

& TIP-Adapter-F \cite{zhang2021tip} &  73.4 & 76.0 & 44.6 & 85.9 & 82.3 & 86.8 & 92.6 & 96.2 & 95.7 & 70.8 & 83.9 & 80.7 \\

& CLIP-Adapter \cite{gao2024clip} &  69.8 & 74.2 & 34.2 & 71.4 & 74.0 & 87.1 & 92.3 & 92.9 & 94.9 & 59.4 & 80.2 & 75.5 \\

& PLOT++ \cite{chen2022plot} &  72.6 & 76.0 & 46.7 & 92.0 & 84.6 & 87.1 & 93.6 & 97.6 & 96.0 & 71.4 & 85.3 & 82.1 \\

& KgCoOp \cite{yao2023visual} &  70.4 & 73.3 & 36.5 & 76.2 & 74.8 & 87.2 & 93.2 & 93.4 & 95.2 & 68.7 & 81.7 & 77.3 \\

& TaskRes \cite{yu2023task} &  73.0 & 76.1 & 44.9 & 82.7 & 83.5 & 86.9 & 92.4 & 97.5 & 95.8 & 71.5 & 84.0 & 80.8 \\

& MaPLe \cite{khattak2023maple} &  71.9 & 74.5 & 36.8 & 87.5 & 74.3 & \textbf{87.4} & 93.2 & 94.2 & 95.4 & 68.4 & 81.4 & 78.6 \\

& ProGrad \cite{zhu2023prograd} &  72.1 & 75.1 & 43.0 & 83.6 & 82.9 & 85.8 & 92.8 & 96.6 & 95.9 & 68.8 & 82.7 & 79.9 \\

& LP++ \cite{huang2024lp++} & 73.0 & 76.0 & 42.1 & 85.5 & 80.8 & 87.2 & 92.6 & 96.3 & 95.8 & 71.9 & 83.9 & 80.5 \\

& CLIP-LoRA \cite{zanella2024cliplora} & 73.6 & 76.1 & \textbf{54.7} & 92.1 & \underline{86.3} & 84.2 & 92.4 & \textbf{98.0} & \textbf{96.4} & 72.0 & \underline{86.7} & \underline{83.0} \\

& MMA \cite{yang2024mma} & 73.2 & 76.6 & 44.7 & 85.0 & 80.2 & 87.0 & \textbf{93.9} & 96.8 & 95.8 & 72.7 & 85.0 & 81.0 \\

\rowcolor{lncolor}
\cellcolor{gray!0} & \ours\textsubscript{\texttt{LayerNorm}} & \underline{73.7} & \textbf{77.0} & 50.0 & \underline{92.4} & 85.4 & 86.1 & 93.7 & 97.7 & \textbf{96.4} & \underline{73.2} & 86.6 & 82.9 \\
\rowcolor{lncolor}
\cellcolor{gray!0} & \ours\textsubscript{\texttt{LoRA}} & \textbf{73.8} & \underline{76.9} & \underline{54.6} & \textbf{92.7} & \textbf{86.9} & 85.7 & \underline{93.8} & \textbf{98.0} & \textbf{96.4} & \textbf{73.5} & \textbf{87.3} & \textbf{83.6} \\
\rowcolor{lncolor}
\cellcolor{gray!0} &  & {+0.2} & {+0.8} & {-0.1} & {+0.6} & {+0.6} & +1.5& +1.4& 0.0 &0.0 & +1.5  & +0.6 & +0.6 \\
 \\ [-5ex] \\ 
    \cmidrule(lr){1-14}
    \cellcolor{gray!0}\multirow{13}{*}{\texttt{ViT-B/32}} 
& \emph{Zero-Shot} & 61.9 & 62.0 & 19.3 & 45.1 & 60.4 & 80.5 & 87.5 & 67.0 & 91.1 & 42.6 & 62.2 & 61.8 \\

& CoOp \cite{zhou2022coop} (ctx=16) & 66.8 & 72.2 & 32.9 & 83.3 & 76.0 & 78.6 & 88.7 & 95.4 & 94.9 & 65.3 & 78.6 & 75.7 \\

& CoCoOp \cite{zhou2022cocoop} & 66.0 & 69.8 & 22.6 & 70.4 & 64.6 & \textbf{81.9} & 91.0 & 82.5 & 94.3 & 59.7 & 75.3 & 70.7 \\

& TIP-Adapter-F \cite{zhang2021tip} & 68.4 & 74.1 & 34.8 & 83.4 & 77.0 & 81.7 & 90.4 & 94.3 & 95.1 & 68.0 & 80.5 & 77.1 \\

& CLIP-Adapter \cite{gao2024clip} & 64.9 & 71.8 & 26.7 & 64.7 & 68.9 & \textbf{81.9} & 90.1 & 88.7 & 94.8 & 58.1 & 76.5 & 71.6 \\

& PLOT++ \cite{chen2022plot} & 67.4 & 73.4 & 36.3 & 91.1 & 77.4 & 79.7 & 89.1 & \underline{96.3} & 94.9 & 67.0 & 81.5 & 77.6 \\

& KgCoOp \cite{yao2023visual} & 65.4 & 71.0 & 23.7 & 70.1 & 67.3 & 81.7 & 90.8 & 86.1 & 94.4 & 65.1 & 77.5 & 72.1 \\

& TaskRes  \cite{yu2023task} & 68.2 & 73.6 & 37.0 & 77.7 & 78.0 & 81.4 & 89.4 & 95.5 & \underline{95.7} & 68.3 & 80.6 & 76.9 \\

& MaPLe \cite{khattak2023maple} & 66.7 & 72.0 & 28.0 & 83.3 & 66.9 & 82.1 & \textbf{91.7} & 89.0 & 95.1 & 63.4 & 77.3 & 74.1 \\

& ProGrad \cite{zhu2023prograd} & 66.9 & 73.2 & 33.3 & 81.0 & 76.1 & 80.1 & 89.3 & 95.1 & 95.0 & 65.8 & 79.6 & 75.9 \\

& LP++ \cite{huang2024lp++} & 68.1 & 74.0 & 34.3 & 82.8 & 75.2 & 81.8 & 90.5 & 93.9 & 95.0 & 67.8 & 80.1 & 76.7 \\

& CLIP-LoRA \cite{zanella2024cliplora} & \underline{68.4} & 74.0 & \textbf{44.9} & 91.8 & 79.7 & 78.2 & 88.8 & 96.2 & 95.2 & 68.2 & \textbf{82.8} & 78.9 \\

& MMA \cite{yang2024mma} & 68.0 & 74.0 & 34.0 & 80.1 & 73.5 & 81.4 & \underline{91.5} & 94.3 & 95.6 & 68.9 & 81.7 & 76.7 \\

\rowcolor{lncolor}
\cellcolor{gray!0} & \ours\textsubscript{\texttt{LayerNorm}} & \underline{68.4} & \textbf{74.8} & 40.2 & \underline{92.1} & \underline{80.2} & 80.8 & 90.3 & \underline{96.3} & \textbf{95.8} & \textbf{70.4} & 82.3 & \underline{79.2} \\

\rowcolor{lncolor}
\cellcolor{gray!0} & \ours\textsubscript{\texttt{LoRA}} & \textbf{68.6} & \underline{74.5} & \underline{44.4} & \textbf{92.2} & \textbf{81.3} & 80.2 & 90.0 & \textbf{96.4} & \underline{95.7} & \underline{69.6} & \underline{82.5} & \textbf{79.6} \\

\rowcolor{lncolor}
\cellcolor{gray!0} &  & {+0.2} & {+0.5} & {-0.5} & {+0.4} & {+1.6} & +2.0& +1.2& +0.2 &+0.5 & +1.4  & -0.3 & +0.7 \\
 \\ [-5ex] \\ 
    \cmidrule(lr){1-14}
    \cellcolor{gray!0}\multirow{13}{*}{\texttt{ViT-L/14}} 
& \emph{Zero-Shot} & 72.9 & 67.6 & 32.6 & 58.0 & 76.8 & 91.0 & 93.6 & 79.4 & 94.9 & 53.6 & 74.2 & 72.2 \\

& CoOp \cite{zhou2022coop} (ctx=16) & 78.2 & 77.5 & 55.2 & 88.3 & 89.0 & 89.8 & 94.6 & \textbf{99.1} & 97.2 & 74.4 & 87.3 & 84.6 \\

& CoCoOp \cite{zhou2022cocoop} & 77.8 & 76.7 & 45.2 & 79.8 & 82.7 & 91.9 & 95.4 & 95.3 & 97.4 & 71.4 & 85.2 & 81.7 \\

& TIP-Adapter-F \cite{zhang2021tip} & 79.3 & 79.6 & 55.8 & 86.1 & 88.1 & 91.6 & 94.6 & 98.3 & \underline{97.5} & 74.0 & 87.4 & 84.8 \\

& CLIP-Adapter \cite{gao2024clip} & 76.4 & 78.0 & 46.4 & 75.8 & 83.8 & 91.6 & 94.3 & 97.3 & 97.3 & 71.3 & 86.1 & 81.7 \\

& PLOT++ \cite{chen2022plot} & 78.6 & 79.1 & 44.1 & 92.2 & 87.2 & 90.2 & 93.6 & 98.8 & \underline{97.5} & 75.0 & 87.1 & 83.9 \\

& KgCoOp \cite{yao2023visual} & 76.8 & 76.7 & 47.5 & 83.6 & 83.2 & 91.7 & 95.3 & 96.4 & 97.4 & 73.6 & 86.4 & 82.6 \\

& TaskRes \cite{yu2023task} & 78.1 & 76.9 & 55.0 & 84.3 & 87.6 & 91.5 & 94.7 & 97.8 & 97.3 & 74.4 & 86.6 & 84.0 \\

& MaPLe \cite{khattak2023maple} & 78.4 & 78.8 & 46.3 & 85.4 & 83.6 & \textbf{92.0} & 95.4 & 97.4 & 97.2 & 72.7 & 86.5 & 83.1 \\

& ProGrad \cite{zhu2023prograd} & 78.4 & 78.3 & 55.6 & 89.3 & 88.8 & 90.8 & 94.9 & 98.7 & \underline{97.5} & 73.7 & 87.7 & 84.9 \\

& LP++ \cite{huang2024lp++} & 79.3 & 79.7 & 54.6 & 89.3 & 87.7 & 91.7 & 94.9 & 98.5 & 97.4 & 76.1 & 88.1 & 85.2 \\

& CLIP-LoRA \cite{zanella2024cliplora} & 79.6 & 79.4 & \underline{66.2} & \underline{93.1} & \underline{90.9} & 89.9 & 94.3 & 99.0 & 97.3 & 76.5 & \underline{89.9} & 86.9 \\

& MMA \cite{yang2024mma} & \textbf{79.9} & 80.2 & 56.4 & 76.3 & 88.0 & \textbf{92.0} & \textbf{95.5} & 98.4 & \textbf{97.6} & 75.8 & 88.0 & 84.4 \\

\rowcolor{lncolor}
\cellcolor{gray!0} & \ours\textsubscript{\texttt{LayerNorm}} & 79.4 & \underline{80.3} & 64.1 & 92.9 & 90.3 & 91.1 & \textbf{95.5} & \textbf{99.1} & \underline{97.5} & \textbf{78.0} & 89.5 & \underline{87.1} \\

\rowcolor{lncolor}
\cellcolor{gray!0} & \ours\textsubscript{\texttt{LoRA}} & \underline{79.7} & \textbf{80.7} & \textbf{66.5} & \textbf{93.2} & \textbf{91.2} & 90.8 & \textbf{95.5} & 99.0 & \underline{97.5} & \underline{77.2} & \textbf{90.3} & \textbf{87.4} \\

\rowcolor{lncolor}
\cellcolor{gray!0} &  & {+0.1} & {+1.3} & {+0.3} & {+0.1} & {+0.3} & +0.9& +1.2& 0.0 &+0.2 & +0.7  & +0.4 & +0.5 \\

\bottomrule
 \\ [-5ex] \\ 

    \end{tabular}
    \end{adjustbox}
\label{tab:all2alllora}
\end{table*}

\cref{tab:all2alllora} reports results for the all-to-all scenario.
Following \cref{sec:all2all}, we experiment with ViT-B/16, ViT-B/32, and ViT-L/14, inheriting hyperparameters from the base-to-novel setup.
Here, \ourslora{} outperforms, on average, all strategies for all backbones, including \oursln{}.
This further aligns with the evidence of \cref{sec:pre-analysis}, where LoRA is shown to incorporate more helpful knowledge to discriminate among available categories. 
Compared to CLIP-LoRA, \ourslora{} outperforms it in 27 out of 33 \{dataset, backbone\} combinations, further supporting the benefits of the two-stage design.

\subsection{Prompt Learning}\label{app:prompt_learning}
Inspired by ``prefix tuning'' \cite{li2021prefix} for Language Models, Prompt Learning has become arguably the most widely adopted approach to adapt VLMs \cite{zhou2022coop, zhou2022cocoop, khattak2023maple, shu2022test, chen2022plot, jia2022visual, yao2023visual} in recent years. 
We do, hence, explore here if \ours{} also successfully integrates with Prompt Learning techniques.

\mypar{Implementation details.}
 To do so, we focus on CoOp \cite{zhou2022coop} and ``Independent Vision-Language Prompting'' (IVLP), a baseline introduced in \cite{khattak2023maple}. 
 We do not conduct any hyperparameter tuning for these PEFT methods, but we set $\alpha{=}0.3$ as used with LoRA, simply leveraging prior knowledge that excessive Prompt Learning severely harms generalization \cite{zhou2022cocoop}. 
 To exactly compare with both approaches, we use the same batch size, optimizer setup, and number of epochs suggested in the original papers.\footnote{with the only exception of ImageNet, for which we only train for 10 epochs to save computational resources.}
 When switching to the second stage of \ours{}, we train the linear classifier with the same optimizer setup described in the main paper. 

\mypar{Results} with the ViT-B/16 backbone and $k{=}16$ shots are given in \cref{tab:b2n-pl} (\emph{base-to-novel} generalization).
For all benchmarks, \emph{wrapping prompt learning approaches in the two-stage design improves the Harmonic Mean} between seen/unseen semantic categories, providing further evidence to support our findings. 
The gap is particularly evident with CoOp ($+3.56$ overall HM), although also IVLP significantly benefits from this design ($+1.23$ overall HM).
We also emphasize that the design of \ours{} is computationally friendlier than the original approaches: only the gradient \wrt the classifier is required for the second stage. 

\section{Base-to-novel generalization with different backbones}
\label{app:b2nbackbones}
This Appendix complements \cref{sec:base2novel}, where results are given for the ViT-B/16 backbone mimicking the experimental setup of \cite{khattak2023maple,yang2024mma}.
Specifically, here we focus on the comparison with the best competitor MMA \cite{yang2024mma} and expand the experimental evaluation to the ViT-B/32 and ViT-L/14 backbones further.

\begin{table*}[ht!]
\caption{Direct comparison between established prompt learning approaches (CoOp \cite{zhou2022coop} and IVLP \cite{khattak2023maple}) and their behavior when wrapped in the Two-Stage design of \ours{}. 
We examine the \emph{base-to-novel} setting with ViT-B/16 and $k{=}16$ shots per class.}
\label{tab:b2n-pl}
    \def\arraystretch{1.075}
    \centering
    \scriptsize
    \begin{minipage}{0.3\textwidth}
        \centering
        \caption*{\textbf{Average across datasets.}}\vspace{-1em}
        \begin{tabular}{lcc|c}
        \toprule
         \textbf{Method} & \textbf{Base} & \textbf{Novel} & \textbf{HM} \\
         \midrule
         CoOp \cite{zhou2022coop} & 82.69 & 63.22 & 71.66 \\
         \rowcolor{lncolor}
         \ours\textsubscript{\texttt{CoOp}} & 83.49 & 68.27 & \textbf{75.12} \\
         \cmidrule(lr){1-4}
         IVLP & 84.21 & 71.79 & 77.51 \\
         \rowcolor{lncolor}
         \ours\textsubscript{\texttt{IVLP}} & 84.53 & 73.69 & \textbf{78.74} \\
         \bottomrule
        \end{tabular}
        \vspace{0.5em}
    \end{minipage}%
    \hfill
    \centering
    \begin{minipage}{0.3\textwidth}
        \centering
        \caption*{ImageNet}\vspace{-1em}
        \begin{tabular}{lcc|c}
        \toprule
         \textbf{Method} & \textbf{Base} & \textbf{Novel} & \textbf{HM} \\
         \midrule
         CoOp \cite{zhou2022coop} & 76.47 & 67.88 & 71.92 \\
         \rowcolor{lncolor}
         \ours\textsubscript{\texttt{CoOp}} & 77.44 & 71.11 & \textbf{74.14} \\
         \cmidrule(lr){1-4}
         IVLP & 77.00 & 66.50 & 71.37 \\
         \rowcolor{lncolor}
         \ours\textsubscript{\texttt{IVLP}} & 75.85 & 71.05 & \textbf{73.37} \\
         \bottomrule
        \end{tabular}
        \vspace{0.5em}
    \end{minipage}%
    \hfill
    \centering
    \begin{minipage}{0.3\textwidth}
        \centering
        \caption*{Caltech101}\vspace{-1em}
        \begin{tabular}{lcc|c}
        \toprule
         \textbf{Method} & \textbf{Base} & \textbf{Novel} & \textbf{HM} \\
         \midrule
         CoOp \cite{zhou2022coop} & 98.00 & 89.81 & 93.73 \\
         \rowcolor{lncolor}
         \ours\textsubscript{\texttt{CoOp}} & 98.00 & 91.99 & \textbf{94.90} \\
         \cmidrule(lr){1-4}
         IVLP & 98.30 & 93.20 & 95.68 \\
         \rowcolor{lncolor}
         \ours\textsubscript{\texttt{IVLP}} & 98.45 & 94.32 & \textbf{96.34} \\
         \bottomrule
        \end{tabular}
        \vspace{0.5em}
    \end{minipage}%
    \hfill
    \begin{minipage}{0.3\textwidth}
        \centering
        \caption*{Oxford Flowers}\vspace{-1em}
        \begin{tabular}{lcc|c}
        \toprule
         \textbf{Method} & \textbf{Base} & \textbf{Novel} & \textbf{HM} \\
         \midrule
         CoOp \cite{zhou2022coop} & 97.60 & 59.67 & 74.06 \\
         \rowcolor{lncolor}
         \ours\textsubscript{\texttt{CoOp}} & 98.16 & 69.46 & \textbf{81.35} \\
         \cmidrule(lr){1-4}
         IVLP & 97.97 & 72.10 & 83.07 \\
         \rowcolor{lncolor}
         \ours\textsubscript{\texttt{IVLP}} & 98.1 & 72.93 & \textbf{83.66} \\
         \bottomrule
        \end{tabular}
        \vspace{0.5em}
    \end{minipage}
    \hfill
    \begin{minipage}{0.3\textwidth}
        \centering
        \caption*{Oxford Pets}\vspace{-1em}
        \begin{tabular}{lcc|c}
        \toprule
         \textbf{Method} & \textbf{Base} & \textbf{Novel} & \textbf{HM} \\
         \midrule
         CoOp \cite{zhou2022coop} &  93.67 & 95.29 & 94.47 \\
         \rowcolor{lncolor}
         \ours\textsubscript{\texttt{CoOp}} & 93.35 & 96.96 & \textbf{95.12} \\
         \cmidrule(lr){1-4}
         IVLP & 94.90 & 97.20 & 96.04 \\
         \rowcolor{lncolor}
         \ours\textsubscript{\texttt{IVLP}} & 95.46 & 97.61 & \textbf{96.53} \\
         \bottomrule
        \end{tabular}
        \vspace{0.5em}
    \end{minipage}%
    \hfill
    \begin{minipage}{0.3\textwidth}
        \centering
        \caption*{Stanford Cars}\vspace{-1em}
        \begin{tabular}{lcc|c}
        \toprule
         \textbf{Method} & \textbf{Base} & \textbf{Novel} & \textbf{HM} \\
         \midrule
         CoOp \cite{zhou2022coop} & 78.12 & 60.40 & 68.13 \\
         \rowcolor{lncolor}
         \ours\textsubscript{\texttt{CoOp}} & 80.15 & 67.87 & \textbf{73.50} \\
         \cmidrule(lr){1-4}
         IVLP & 79.53 & 71.47 & 75.28 \\
         \rowcolor{lncolor}
         \ours\textsubscript{\texttt{IVLP}} & 81.69 & 73.5 & \textbf{77.38} \\
         \bottomrule
        \end{tabular}
        \vspace{0.5em}
    \end{minipage}%
    \hfill
    \begin{minipage}{0.3\textwidth}
        \centering
        \caption*{Food 101}\vspace{-1em}
        \begin{tabular}{lcc|c}
        \toprule
         \textbf{Method} & \textbf{Base} & \textbf{Novel} & \textbf{HM} \\
         \midrule
         CoOp \cite{zhou2022coop} & 88.33 & 82.26 & 85.19 \\
         \rowcolor{lncolor}
         \ours\textsubscript{\texttt{CoOp}} & 88.06 & 88.68 & \textbf{88.37} \\
         \cmidrule(lr){1-4}
         IVLP & 89.37 & 90.30 & 89.83 \\
         \rowcolor{lncolor}
         \ours\textsubscript{\texttt{IVLP}} & 89.45 & 91.51 & \textbf{90.47} \\
         \bottomrule
        \end{tabular}
        \vspace{0.5em}
    \end{minipage}
    \hfill
    \begin{minipage}{0.3\textwidth}
        \centering
        \caption*{FGVC Aircraft}\vspace{-1em}
        \begin{tabular}{lcc|c}
        \toprule
         \textbf{Method} & \textbf{Base} & \textbf{Novel} & \textbf{HM} \\
         \midrule
         CoOp \cite{zhou2022coop} & 40.44 & 22.30 & 28.75 \\
         \rowcolor{lncolor}
         \ours\textsubscript{\texttt{CoOp}} & 44.60 & 29.91 & \textbf{35.81} \\
         \cmidrule(lr){1-4}
         IVLP & 42.60 & 25.23 & 31.69 \\
         \rowcolor{lncolor}
         \ours\textsubscript{\texttt{IVLP}} & 44.78 & 25.93 & \textbf{32.85} \\
         \bottomrule
        \end{tabular}
        \vspace{0.5em}
    \end{minipage}%
    \hfill
    \begin{minipage}{0.3\textwidth}
        \centering
        \caption*{SUN 397}\vspace{-1em}
        \begin{tabular}{lcc|c}
        \toprule
         \textbf{Method} & \textbf{Base} & \textbf{Novel} & \textbf{HM} \\
         \midrule
         CoOp \cite{zhou2022coop} & 80.60 & 65.89 & 72.51 \\
         \rowcolor{lncolor}
         \ours\textsubscript{\texttt{CoOp}} & 79.16 & 70.32 & \textbf{74.48} \\
         \cmidrule(lr){1-4}
         IVLP & 81.60 & 75.50 & 78.43 \\
         \rowcolor{lncolor}
         \ours\textsubscript{\texttt{IVLP}} & 81.31 & 78.27 & \textbf{79.76} \\
         \bottomrule
        \end{tabular}
        \vspace{0.5em}
    \end{minipage}
    \hfill
    \begin{minipage}{0.3\textwidth}
        \centering
        \caption*{DTD}\vspace{-1em}
        \begin{tabular}{lcc|c}
        \toprule
         \textbf{Method} & \textbf{Base} & \textbf{Novel} & \textbf{HM} \\
         \midrule
         CoOp \cite{zhou2022coop} & 79.44 & 41.18 & 54.24 \\
         \rowcolor{lncolor}
         \ours\textsubscript{\texttt{CoOp}} & 81.40 & 49.11 & \textbf{61.27} \\
         \cmidrule(lr){1-4}
         IVLP & 82.40 & 56.20 & \textbf{66.82} \\
         \rowcolor{lncolor}
         \ours\textsubscript{\texttt{IVLP}} & 83.45 & 53.66 & 65.32 \\
         \bottomrule
        \end{tabular}
    \end{minipage}%
    \hfill
    \begin{minipage}{0.3\textwidth}
        \centering
        \caption*{EuroSAT}\vspace{-1em}
        \begin{tabular}{lcc|c}
        \toprule
         \textbf{Method} & \textbf{Base} & \textbf{Novel} & \textbf{HM} \\
         \midrule
         CoOp \cite{zhou2022coop} & 92.19 & 54.74 & \textbf{68.69} \\
         \rowcolor{lncolor}
         \ours\textsubscript{\texttt{CoOp}} & 92.94 & 50.76 & 65.66 \\
         \cmidrule(lr){1-4}
         IVLP & 96.73 & 67.83 & 79.74 \\
         \rowcolor{lncolor}
         \ours\textsubscript{\texttt{IVLP}} & 95.01 & 74.51 & \textbf{83.52} \\
         \bottomrule
        \end{tabular}
    \end{minipage}
    \hfill
    \begin{minipage}{0.3\textwidth}
        \centering
        \caption*{UCF101}\vspace{-1em}
        \begin{tabular}{lcc|c}
        \toprule
         \textbf{Method} & \textbf{Base} & \textbf{Novel} & \textbf{HM} \\
         \midrule
         CoOp \cite{zhou2022coop} & 84.69 & 56.05 & 67.46 \\
         \rowcolor{lncolor}
         \ours\textsubscript{\texttt{CoOp}} & 85.04 & 64.67 & \textbf{73.47} \\
         \cmidrule(lr){1-4}
         IVLP & 85.93 & 74.17 & 79.62 \\
         \rowcolor{lncolor}
         \ours\textsubscript{\texttt{IVLP}} & 86.28 & 77.27 & \textbf{81.53} \\
         \bottomrule
        \end{tabular}
    \end{minipage}
\end{table*}

\begin{table*}
\caption{Experiments in \emph{base-to-novel} generalization, with the ViT-B/32 visual backbone and $k=16$ shots per base category, focusing on the comparison with MultiModal Adapter (MMA) \cite{yang2024mma}. ``CLIP'' refers to zero-shot performance with dataset-specific templates, \eg, ``\emph{a photo of a \{\}, a type of flower}'' for Oxford Flowers. Formatting follows \cref{tab:b2n}.}
\label{tab:b2nvitb32}
    \def\arraystretch{1.075}
    \centering
    \scriptsize
    \begin{minipage}{0.32\textwidth}
        \centering
        \caption*{\textbf{Average across datasets.}}\vspace{-1em}
        \begin{tabular}{lcc|c}
        \toprule
         \textbf{Method} & \textbf{Base} & \textbf{Novel} & \textbf{HM} \\
         \midrule
         CLIP \cite{radford2021learning} & 67.27 & 71.68 & 69.41 \\
         MMA \cite{yang2024mma} & 78.69 & 71.04 & 74.67 \\
         \cmidrule(lr){1-4}
         \rowcolor{lncolor}
         \ours{}  & 82.32 & 71.23 & \textbf{76.37} \\
         \bottomrule
        \end{tabular}
        \vspace{0.5em}
    \end{minipage}%
    \hfill
    \centering
    \begin{minipage}{0.32\textwidth}
        \centering
        \caption*{ImageNet}\vspace{-1em}
        \begin{tabular}{lcc|c}
        \toprule
         \textbf{Method} & \textbf{Base} & \textbf{Novel} & \textbf{HM} \\
         \midrule
          CLIP \cite{radford2021learning} & 67.49 & 64.06 & 65.73 \\
         MMA \cite{yang2024mma} & 72.53 & 65.77 & 68.98 \\
         \cmidrule(lr){1-4}
         \rowcolor{lncolor}
         \ours{} & 72.52 & 66.62 & \textbf{69.44} \\
         \bottomrule
        \end{tabular}
        \vspace{0.5em}
    \end{minipage}%
    \hfill
    \centering
    \begin{minipage}{0.32\textwidth}
        \centering
        \caption*{Caltech101}\vspace{-1em}
        \begin{tabular}{lcc|c}
        \toprule
         \textbf{Method} & \textbf{Base} & \textbf{Novel} & \textbf{HM} \\
         \midrule
          CLIP \cite{radford2021learning} & 94.06 & 94.00 & 94.03 \\
         MMA \cite{yang2024mma} & 97.20 & 92.63 & 94.86 \\
         \cmidrule(lr){1-4}
         \rowcolor{lncolor}
         \ours{}  & 97.83 & 93.30 & \textbf{95.51} \\
         \bottomrule
        \end{tabular}
        \vspace{0.5em}
    \end{minipage}%
    \hfill
    \begin{minipage}{0.32\textwidth}
        \centering
        \caption*{Oxford Flowers}\vspace{-1em}
        \begin{tabular}{lcc|c}
        \toprule
         \textbf{Method} & \textbf{Base} & \textbf{Novel} & \textbf{HM} \\
         \midrule
          CLIP \cite{radford2021learning} & 72.36 & 73.69 & 73.02 \\
         MMA \cite{yang2024mma} & 95.50 & 71.57 & \textbf{81.82} \\
         \cmidrule(lr){1-4}
         \rowcolor{lncolor}
         \ours{}  & 96.64 & 70.02 & 81.20 \\
         \bottomrule
        \end{tabular}
        \vspace{0.5em}
    \end{minipage}
    \hfill
    \begin{minipage}{0.32\textwidth}
        \centering
        \caption*{Oxford Pets}\vspace{-1em}
        \begin{tabular}{lcc|c}
        \toprule
         \textbf{Method} & \textbf{Base} & \textbf{Novel} & \textbf{HM} \\
         \midrule
          CLIP \cite{radford2021learning} & 90.64 & 96.87 & 93.65 \\
         MMA \cite{yang2024mma} & 93.77 & 96.30 & \textbf{95.02} \\
         \cmidrule(lr){1-4}
         \rowcolor{lncolor}
         \ours{}  & 93.18 & 95.56 & 94.35 \\
         \bottomrule
        \end{tabular}
        \vspace{0.5em}
    \end{minipage}%
    \hfill
    \begin{minipage}{0.32\textwidth}
        \centering
        \caption*{Stanford Cars}\vspace{-1em}
        \begin{tabular}{lcc|c}
        \toprule
         \textbf{Method} & \textbf{Base} & \textbf{Novel} & \textbf{HM} \\
         \midrule
          CLIP \cite{radford2021learning} & 60.72 & 69.74 & 64.92 \\
         MMA \cite{yang2024mma} & 73.73 & 69.27 & 71.43 \\
         \cmidrule(lr){1-4}
         \rowcolor{lncolor}
         \ours{}  & 78.21 & 70.30 & \textbf{74.04} \\
         \bottomrule
        \end{tabular}
        \vspace{0.5em}
    \end{minipage}%
    \hfill
    \begin{minipage}{0.32\textwidth}
        \centering
        \caption*{Food 101}\vspace{-1em}
        \begin{tabular}{lcc|c}
        \toprule
         \textbf{Method} & \textbf{Base} & \textbf{Novel} & \textbf{HM} \\
         \midrule
        CLIP \cite{radford2021learning} & 85.30 & 86.89 & 86.09 \\
         MMA \cite{yang2024mma} & 85.77 & 87.13 & \textbf{86.44} \\
         \cmidrule(lr){1-4}
         \rowcolor{lncolor}
         \ours{}  & 84.75 & 87.37 & 86.04 \\ 
         \bottomrule
        \end{tabular}
        \vspace{0.5em}
    \end{minipage}
    \hfill
    \begin{minipage}{0.32\textwidth}
        \centering
        \caption*{FGVC Aircraft}\vspace{-1em}
        \begin{tabular}{lcc|c}
        \toprule
         \textbf{Method} & \textbf{Base} & \textbf{Novel} & \textbf{HM} \\
         \midrule
          CLIP \cite{radford2021learning} & 21.25 & 29.27 & 24.62 \\
         MMA \cite{yang2024mma} & 31.77 & 28.73 & 30.17 \\
         \cmidrule(lr){1-4}
         \rowcolor{lncolor}
         \ours{}  & 39.12 & 30.85 & \textbf{34.50} \\
         \bottomrule
        \end{tabular}
        \vspace{0.5em}
    \end{minipage}%
    \hfill
    \begin{minipage}{0.32\textwidth}
        \centering
        \caption*{SUN 397}\vspace{-1em}
        \begin{tabular}{lcc|c}
        \toprule
         \textbf{Method} & \textbf{Base} & \textbf{Novel} & \textbf{HM} \\
         \midrule
          CLIP \cite{radford2021learning} & 69.80 & 73.01 & 71.37 \\
         MMA \cite{yang2024mma} & 80.27 & 76.57 & 78.38 \\
         \cmidrule(lr){1-4}
         \rowcolor{lncolor}
         \ours{}  & 81.11 & 78.02 & \textbf{79.53} \\
         \bottomrule
        \end{tabular}
        \vspace{0.5em}
    \end{minipage}
    \hfill
    \begin{minipage}{0.32\textwidth}
        \centering
        \caption*{DTD}\vspace{-1em}
        \begin{tabular}{lcc|c}
        \toprule
         \textbf{Method} & \textbf{Base} & \textbf{Novel} & \textbf{HM} \\
         \midrule
        CLIP \cite{radford2021learning} & 54.17 & 58.21 & 56.12 \\
         MMA \cite{yang2024mma} & 79.50 & 57.00 & \textbf{66.40} \\
         \cmidrule(lr){1-4}
         \rowcolor{lncolor}
         \ours{}  & 80.09 & 54.63 & 64.95 \\
         \bottomrule
        \end{tabular}
    \end{minipage}%
    \hfill
    \begin{minipage}{0.32\textwidth}
        \centering
        \caption*{EuroSAT}\vspace{-1em}
        \begin{tabular}{lcc|c}
        \toprule
         \textbf{Method} & \textbf{Base} & \textbf{Novel} & \textbf{HM} \\
         \midrule
        CLIP \cite{radford2021learning} & 55.14 & 69.77 & 61.60 \\
         MMA \cite{yang2024mma} & 71.83 & 62.97 & 67.11 \\
         \cmidrule(lr){1-4}
         \rowcolor{lncolor}
         \ours{}  & 96.80 & 62.68 & \textbf{76.09} \\
         \bottomrule
        \end{tabular}
    \end{minipage}
    \hfill
    \begin{minipage}{0.32\textwidth}
        \centering
        \caption*{UCF101}\vspace{-1em}
        \begin{tabular}{lcc|c}
        \toprule
         \textbf{Method} & \textbf{Base} & \textbf{Novel} & \textbf{HM} \\
         \midrule
        CLIP \cite{radford2021learning} & 69.08 & 72.96 & 70.97 \\
         MMA \cite{yang2024mma} & 83.77 & 73.47 & 78.28 \\
         \cmidrule(lr){1-4}
         \rowcolor{lncolor}
         \ours{}  & 85.25 & 74.15 & \textbf{79.31} \\
         \bottomrule
        \end{tabular}
    \end{minipage}
\end{table*}

\begin{table*}
\caption{Experiments in \emph{base-to-novel} generalization, with the ViT-L/14 visual backbone and $k=16$ shots per base category, focusing on the comparison with MultiModal Adapter (MMA) \cite{yang2024mma}. ``CLIP'' refers to zero-shot performance with dataset-specific templates, \eg, ``\emph{a photo of a \{\}, a type of flower}'' for Oxford Flowers. Formatting follows \cref{tab:b2n}.}
\label{tab:b2nvitl14}
    \def\arraystretch{1.075}
    \centering
    \scriptsize
    \begin{minipage}{0.32\textwidth}
        \centering
        \caption*{\textbf{Average across datasets.}}\vspace{-1em}
        \begin{tabular}{lcc|c}
        \toprule
         \textbf{Method} & \textbf{Base} & \textbf{Novel} & \textbf{HM} \\
         \midrule
         CLIP \cite{radford2021learning} & 76.18 & 80.08 & 78.08 \\
         MMA \cite{yang2024mma} & 85.70 & 79.06 & 82.25 \\
         \cmidrule(lr){1-4}
         \rowcolor{lncolor}
         \ours{}  & 89.05 & 79.64 & \textbf{84.08} \\
         \bottomrule
        \end{tabular}
        \vspace{0.5em}
    \end{minipage}%
    \hfill
    \centering
    \begin{minipage}{0.32\textwidth}
        \centering
        \caption*{ImageNet}\vspace{-1em}
        \begin{tabular}{lcc|c}
        \toprule
         \textbf{Method} & \textbf{Base} & \textbf{Novel} & \textbf{HM} \\
         \midrule
          CLIP \cite{radford2021learning} & 79.18	& 74.04 & 76.53 \\
         MMA \cite{yang2024mma} & 83.17 & 76.73 & 79.82 \\
         \cmidrule(lr){1-4}
         \rowcolor{lncolor}
         \ours{} & 83.11 & 76.98 & \textbf{79.93} \\
         \bottomrule
        \end{tabular}
        \vspace{0.5em}
    \end{minipage}%
    \hfill
    \centering
    \begin{minipage}{0.32\textwidth}
        \centering
        \caption*{Caltech101}\vspace{-1em}
        \begin{tabular}{lcc|c}
        \toprule
         \textbf{Method} & \textbf{Base} & \textbf{Novel} & \textbf{HM} \\
         \midrule
          CLIP \cite{radford2021learning} & 95.61 & 95.41 & 95.51 \\
         MMA \cite{yang2024mma} & 98.60 & 95.97 & 97.27 \\
         \cmidrule(lr){1-4}
         \rowcolor{lncolor}
         \ours{}  & 98.82 & 96.69 & \textbf{97.74} \\
         \bottomrule
        \end{tabular}
        \vspace{0.5em}
    \end{minipage}%
    \hfill
    \begin{minipage}{0.32\textwidth}
        \centering
        \caption*{Oxford Flowers}\vspace{-1em}
        \begin{tabular}{lcc|c}
        \toprule
         \textbf{Method} & \textbf{Base} & \textbf{Novel} & \textbf{HM} \\
          CLIP \cite{radford2021learning} & 80.34 & 83.05 & 81.67 \\
         MMA \cite{yang2024mma} & 99.00 & 80.20 & 88.61 \\
         \cmidrule(lr){1-4}
         \rowcolor{lncolor}
         \ours{}  & 98.99 & 80.73 & \textbf{88.93} \\
         \bottomrule
        \end{tabular}
        \vspace{0.5em}
    \end{minipage}
    \hfill
    \begin{minipage}{0.32\textwidth}
        \centering
        \caption*{Oxford Pets}\vspace{-1em}
        \begin{tabular}{lcc|c}
        \toprule
         \textbf{Method} & \textbf{Base} & \textbf{Novel} & \textbf{HM} \\
         \midrule
          CLIP \cite{radford2021learning} & 93.78 & 96.53 & 95.14 \\
         MMA \cite{yang2024mma} & 96.23 & 98.70 & 97.45 \\
         \cmidrule(lr){1-4}
         \rowcolor{lncolor}
         \ours{}  & 96.74 & 98.64 & \textbf{97.68} \\
         \bottomrule
        \end{tabular}
        \vspace{0.5em}
    \end{minipage}%
    \hfill
    \begin{minipage}{0.32\textwidth}
        \centering
        \caption*{Stanford Cars}\vspace{-1em}
        \begin{tabular}{lcc|c}
        \toprule
         \textbf{Method} & \textbf{Base} & \textbf{Novel} & \textbf{HM} \\
         \midrule
          CLIP \cite{radford2021learning} & 74.56 & 84.65 & 79.29 \\
         MMA \cite{yang2024mma} & 85.27 & 83.80 & 84.53 \\
         \cmidrule(lr){1-4}
         \rowcolor{lncolor}
         \ours{}  & 87.46 & 84.56 & \textbf{85.99} \\
         \bottomrule
        \end{tabular}
        \vspace{0.5em}
    \end{minipage}%
    \hfill
    \begin{minipage}{0.32\textwidth}
        \centering
        \caption*{Food 101}\vspace{-1em}
        \begin{tabular}{lcc|c}
        \toprule
         \textbf{Method} & \textbf{Base} & \textbf{Novel} & \textbf{HM} \\
         \midrule
        CLIP \cite{radford2021learning} & 93.75 & 94.82 & 94.28 \\
         MMA \cite{yang2024mma} & 94.23 & 95.10 & \textbf{94.66} \\
         \cmidrule(lr){1-4}
         \rowcolor{lncolor}
         \ours{}  & 93.59 & 94.93 & 94.26 \\ 
         \bottomrule
        \end{tabular}
        \vspace{0.5em}
    \end{minipage}
    \hfill
    \begin{minipage}{0.32\textwidth}
        \centering
        \caption*{FGVC Aircraft}\vspace{-1em}
        \begin{tabular}{lcc|c}
        \toprule
         \textbf{Method} & \textbf{Base} & \textbf{Novel} & \textbf{HM} \\
         \midrule
          CLIP \cite{radford2021learning} & 37.52 & 44.21 & 40.59 \\
         MMA \cite{yang2024mma} & 50.00 & 42.47 & 45.93 \\
         \cmidrule(lr){1-4}
         \rowcolor{lncolor}
         \ours{}  & 59.76 & 43.59 & \textbf{50.41} \\
         \bottomrule
        \end{tabular}
        \vspace{0.5em}
    \end{minipage}%
    \hfill
    \begin{minipage}{0.32\textwidth}
        \centering
        \caption*{SUN 397}\vspace{-1em}
        \begin{tabular}{lcc|c}
        \toprule
         \textbf{Method} & \textbf{Base} & \textbf{Novel} & \textbf{HM} \\
         \midrule
          CLIP \cite{radford2021learning} & 73.23 & 77.71 & 75.40 \\
         MMA \cite{yang2024mma} & 85.03 & 81.77 & 83.37 \\
         \cmidrule(lr){1-4}
         \rowcolor{lncolor}
         \ours{}  & 85.57 & 82.24 & \textbf{83.87} \\
         \bottomrule
        \end{tabular}
        \vspace{0.5em}
    \end{minipage}
    \hfill
    \begin{minipage}{0.32\textwidth}
        \centering
        \caption*{DTD}\vspace{-1em}
        \begin{tabular}{lcc|c}
        \toprule
         \textbf{Method} & \textbf{Base} & \textbf{Novel} & \textbf{HM} \\
         \midrule
        CLIP \cite{radford2021learning} & 59.14 & 67.87 & 63.21 \\
         MMA \cite{yang2024mma} & 85.23 & 70.77 & 77.33 \\
         \cmidrule(lr){1-4}
         \rowcolor{lncolor}
         \ours{}  & 87.35 & 70.73 & \textbf{78.17} \\
         \bottomrule
        \end{tabular}
    \end{minipage}%
    \hfill
    \begin{minipage}{0.32\textwidth}
        \centering
        \caption*{EuroSAT}\vspace{-1em}
        \begin{tabular}{lcc|c}
        \toprule
         \textbf{Method} & \textbf{Base} & \textbf{Novel} & \textbf{HM} \\
         \midrule
        CLIP \cite{radford2021learning} & 70.93	& 82.90 & 76.45 \\
         MMA \cite{yang2024mma} & 77.33 & 62.77 & 69.29 \\
         \cmidrule(lr){1-4}
         \rowcolor{lncolor}
         \ours{} & 98.41 & 64.69 & \textbf{78.06} \\
         \bottomrule
        \end{tabular}
    \end{minipage}
    \hfill
    \begin{minipage}{0.32\textwidth}
        \centering
        \caption*{UCF101}\vspace{-1em}
        \begin{tabular}{lcc|c}
        \toprule
         \textbf{Method} & \textbf{Base} & \textbf{Novel} & \textbf{HM} \\
         \midrule
        CLIP \cite{radford2021learning} & 79.94 & 79.66 & 79.80 \\
         MMA \cite{yang2024mma} & 88.60 & 81.37 & 84.83 \\
         \cmidrule(lr){1-4}
         \rowcolor{lncolor}
         \ours{}  & 89.80 & 82.24 & \textbf{85.85} \\
         \bottomrule
        \end{tabular}
    \end{minipage}
\end{table*}

\mypar{Implementation Details.} To align with \cref{sec:base2novel}, we use layer normalization in the first stage as in the main body of the paper, and make no hyperparameter changes. 
Results for these backbones are not available in the official article of \cite{yang2024mma}, hence we used the open-source implementation of the authors with no modifications (the repository already integrates with different CLIP variants) as done for the all-to-all experiments of \cref{sec:all2all}.

\mypar{Results} are given in \cref{tab:b2nvitb32} for ViT-B/32 and in \cref{tab:b2nvitl14} for ViT-L/14.
For both backbones, \ours{} largely outperforms MMA on average, exhibiting larger improvements than those emerging with the ViT-B/16 visual encoder ($+1.70\%$ and $+1.83\%$, respectively), confirming that the effectiveness of \ours{} does not depend on a specific backbone. 

\section{Varying Shots}
\label{app:shots}
In \cref{sec:experiments} of the main body, results are given for the most popular FSA scenario in which $k{=}16$ shots are available per category.
Here, we test the robustness of \ours{} in extreme data scarcity, working with both $k{=}4$ and $k{=}8$ shots for both all-to-all and base-to-novel cases. 
The results are discussed below.

\mypar{Base-to-novel generalization.}
In line with \cref{app:b2nbackbones}, we focus on the comparison with MMA \cite{yang2024mma}.
We experiment with all backbones, and report results in \cref{tab:b2n_vitb16_shots4} and \cref{tab:b2n_vitb16_shots8} for ViT-B/16, \cref{tab:b2n_vitb32_shots4} and \cref{tab:b2n_vitb32_shots8} for ViT-B/32, and \cref{tab:b2n_vitl14_shots4} and \cref{tab:b2n_vitl14_shots8} for ViT-L/14, with 4 and 8 shots, respectively.\footnote{Please note that results for CoOp, CoCoOp, ProGrad, and KgCoOp with the ViT-B/16 backbone and $k{\in}\{4,8\}$ are given in the supplementary material of \cite{yao2023visual}, which we omit to avoid excessively dense tables. \ours{} largely outperforms all methods with available results.}

On average, \emph{\ours{} outperforms MMA for all backbones and all shots setups}.
Importantly, we observe that the performance gap increases as the shots decrease, up to large gaps such as $+3.98\%$ and $+4.45\%$ HM with ViT-B/16 and ViT-B/32 using 4 shots. 
We speculate this behavior stems from the reduced amount of learnable parameters of \ours{}, which better accommodates a smaller number of examples. 
To ground the discussion in some numbers: summing up LayerNorm instances totals around 61k parameters for ViT-B backbones, while MMA introduces 674k new parameters.

\begin{table}[t!]
    \centering
    \begin{tabular}{ccc|c}
    \toprule
     M & \textbf{Base} & \textbf{Novel} & \textbf{HM}  \\
     \midrule
     $M=100$ & 77.55 & 70.50 & 73.86 \\
     $M=300$ & 77.71 & 70.99 & 74.20 \\
     $M=500$ & 77.35 & 71.16 & 74.12 \\
     \bottomrule
    \end{tabular}
    \caption{Sweep on $M\in\{100,300,500\}$ when $\alpha=0.6$ on the ImageNet validation set and CLIP ViT-B/16.}
    \label{tab:sweep-on-M}
\end{table}

\mypar{All-to-all adaptation.}
\begin{table*}
    \def\arraystretch{1.1}
    \centering
    \small
    \caption{\emph{All-to-all} experiments with $\mathbf{k=4}$ shots, using ViT-B/16, ViT-B/32, and ViT-L/14. Formatting follows \cref{tab:all2all}.}
    \begin{adjustbox}{max width=\textwidth}
    \begin{tabular}{llccccccccccc|c}
    \toprule
     \textsc{\textbf{Backbone}} &  \textsc{\textbf{Method}} &  \textsc{\textbf{ImageNet}} &  \textsc{\textbf{SUN}} &  \textsc{\textbf{AIR}} &  \textsc{\textbf{ESAT}} &  \textsc{\textbf{CARS}} &  \textsc{\textbf{FOOD}} &  \textsc{\textbf{PETS}} &  \textsc{\textbf{FLWR}} &  \textsc{\textbf{CAL}} &  \textsc{\textbf{DTD}} &  \textsc{\textbf{UCF}} &  \textsc{\textbf{Mean}} \\
    \midrule
    
    \cellcolor{gray!0} \multirow{13}{*}{\texttt{ViT-B/16}} & \emph{Zero-Shot} \cite{radford2021learning} & 66.7 & 62.6 & 24.7 & 47.5 & 65.3 & 86.1 & 89.1 & 71.4 & 92.9 & 43.6 & 66.7 & 65.1 \\
 & CoOp \cite{zhou2022coop} (ctx=16) & 68.8 & 69.7 & 30.9 & 69.7 & 74.4 & 84.5 & 92.5 & 92.2 & 94.5 & 59.5 & 77.6 & 74.0 \\
 & CoCoOp \cite{zhou2022cocoop} & 70.6 & 70.4 & 30.6 & 61.7 & 69.5 & 86.3 & 92.7 & 81.5 & 94.8 & 55.7 & 75.3 & 71.7 \\
 & TIP-Adapter-F \cite{zhang2021tip} & 70.7 & 70.8 & 35.7 & 76.8 & 74.1 & 86.5 & 91.9 & 92.1 & 94.8 & 59.8 & 78.1 & 75.6 \\
 & CLIP-Adapter \cite{gao2024clip} & 68.6 & 68.0 & 27.9 & 51.2 & 67.5 & 86.5 & 90.8 & 73.1 & 94.0 & 46.1 & 70.6 & 67.7 \\
 & PLOT++ \cite{chen2022plot} & 70.4 & 71.7 & 35.3 & 83.2 & 76.3 & 86.5 & 92.6 & 92.9 & 95.1 & 62.4 & 79.8 & 76.9 \\
 & KgCoOp \cite{yao2023visual} & 69.9 & 71.5 & 32.2 & 71.8 & 69.5 & \textbf{86.9} & 92.6 & 87.0 & 95.0 & 58.7 & 77.6 & 73.9 \\
 & TaskRes \cite{yu2023task} & 71.0 & 72.7 & 33.4 & 74.2 & 76.0 & 86.0 & 91.9 & 85.0 & 95.0 & 60.1 & 76.2 & 74.7 \\
 & MaPLe \cite{khattak2023maple} & 70.6 & 71.4 & 30.1 & 69.9 & 70.1 & \underline{86.7} & \textbf{93.3} & 84.9 & 95.0 & 59.0 & 77.1 & 73.5 \\
 & ProGrad \cite{zhu2023prograd} & 70.2 & 71.7 & 34.1 & 69.6 & 75.0 & 85.4 & 92.1 & 91.1 & 94.4 & 59.7 & 77.9 & 74.7 \\
 & LP++ \cite{huang2024lp++} & 70.8 & \underline{73.2} & 34.0 & 73.6 & 74.0 & 85.9 & 90.9 & 93.0 & 95.1 & 62.4 & 79.2 & 75.6 \\
 & CLIP-LoRA \cite{zanella2024cliplora} & \textbf{71.4} & 72.8 & \underline{37.9} & \underline{84.9} & \underline{77.4} & 82.7 & 91.0 & \underline{93.7} & \underline{95.2} & \underline{63.8} & \underline{81.1} & \underline{77.4} \\
 & MMA \cite{yang2024mma} & 70.5 & 72.9 & 35.0 & 42.4 & 73.3 & 86.0 & \underline{92.9} & 91.3 & 94.5 & 60.1 & 79.0 & 72.5 \\

 \rowcolor{lncolor}
\cellcolor{gray!0} & \ours{} & \underline{71.1} & \textbf{73.7} & \textbf{39.8} & \textbf{85.5} & \textbf{77.5} & 85.9 & 92.6 & \textbf{94.0} & \textbf{95.4} & \textbf{66.0} & \textbf{82.0} & \textbf{78.5} \\
 \\ [-5ex] \\ 
    \cmidrule(lr){1-14}
    \cellcolor{gray!0} \multirow{13}{*}{\texttt{ViT-B/32}} & \emph{Zero-Shot} \cite{radford2021learning} & 61.9 & 62.0 & 19.3 & 45.1 & 60.4 & 80.5 & 87.5 & 67.0 & 91.1 & 42.6 & 62.2 & 61.8 \\
 & CoOp \cite{zhou2022coop} (ctx=16) & 63.2 & 67.1 & 24.0 & 68.7 & 66.2 & 75.6 & 88.8 & 87.9 & 93.0 & 55.3 & 75.0 & 69.5 \\
 & CoCoOp \cite{zhou2022cocoop} & 65.2 & 67.8 & 17.3 & 58.5 & 62.0 & 81.1 & 89.8 & 74.6 & 93.2 & 52.3 & 71.6 & 66.7 \\
 & TIP-Adapter-F \cite{zhang2021tip} & 65.8 & 68.3 & \underline{28.8} & 71.5 & 67.6 & 80.9 & 88.6 & 88.9 & \underline{94.6} & 58.0 & 75.1 & 71.6 \\
 & CLIP-Adapter \cite{gao2024clip} & 63.7 & 65.6 & 21.3 & 49.9 & 62.2 & \underline{81.3} & 88.4 & 68.3 & 92.0 & 47.2 & 67.3 & 64.3 \\
 & PLOT++ \cite{chen2022plot} & 64.6 & 69.2 & 26.2 & 81.6 & \underline{68.5} & 77.8 & 89.1 & \underline{90.2} & 93.9 & 57.2 & 75.6 & 72.2 \\
 & KgCoOp \cite{yao2023visual} & 64.7 & 69.2 & 22.6 & 64.9 & 63.2 & 81.2 & 89.5 & 76.8 & 93.8 & 55.1 & 71.6 & 68.4 \\
 & TaskRes \cite{yu2023task} & \underline{66.1} & 66.7 & 23.1 & 70.7 & 66.7 & 76.7 & 86.7 & 79.0 & 90.6 & 57.0 & 68.2 & 68.3 \\
 & MaPLe \cite{khattak2023maple} & 65.6 & 69.4 & 23.4 & 64.7 & 62.2 & \textbf{81.4} & \underline{90.5} & 78.1 & 94.0 & 55.0 & 70.9 & 68.7 \\
 & ProGrad \cite{zhu2023prograd} & 65.2 & 69.6 & 24.8 & 63.7 & 66.4 & 79.2 & 89.4 & 87.5 & 93.2 & 55.9 & 73.4 & 69.8 \\
 & LP++ \cite{huang2024lp++} & \underline{66.1} & \underline{70.5} & 26.0 & 73.5 & 67.3 & 80.0 & 88.9 & \underline{90.2} & 94.0 & 59.3 & 74.8 & 71.9 \\
 & CLIP-LoRA \cite{zanella2024cliplora} & \textbf{66.5} & 70.3 & 27.7 & \textbf{85.6} & 68.3 & 75.6 & 86.3 & 90.1 & 94.3 & \underline{60.3} & \underline{76.5} & \underline{72.9} \\
 & MMA \cite{yang2024mma} & 64.7 & 70.4 & 25.6 & 36.0 & 66.3 & 80.5 & \textbf{90.7} & 86.1 & 94.0 & 55.6 & 74.6 & 67.7 \\

\rowcolor{lncolor}
\cellcolor{gray!0} & \ours{} & 66.0 & \textbf{71.4} & \textbf{30.6} & \underline{82.6} & \textbf{70.4} & 80.2 & 89.4 & \textbf{91.0} & \textbf{95.1} & \textbf{63.1} & \textbf{77.4} & \textbf{74.3} \\
 \\ [-5ex] \\ 
    \cmidrule(lr){1-14}
    \cellcolor{gray!0} \multirow{13}{*}{\texttt{ViT-L/14}} & \emph{Zero-Shot} \cite{radford2021learning} & 72.9 & 67.6 & 32.6 & 58.0 & 76.8 & 91.0 & 93.6 & 79.4 & 94.9 & 53.6 & 74.2 & 72.2 \\
 & CoOp \cite{zhou2022coop} (ctx=16) & 74.9 & 73.1 & 43.6 & 75.9 & 83.3 & 88.7 & 94.6 & 95.9 & 96.5 & 63.9 & 82.8 & 79.4 \\
 & CoCoOp \cite{zhou2022cocoop} & 77.0 & 74.7 & 41.0 & 74.7 & 79.7 & 91.3 & 94.9 & 89.8 & 97.1 & 64.9 & 82.6 & 78.9 \\
 & TIP-Adapter-F \cite{zhang2021tip} & 77.1 & 74.1 & 47.4 & 81.4 & 82.3 & 91.2 & 94.0 & 95.5 & 96.5 & 64.4 & 83.9 & 80.7 \\
 & CLIP-Adapter \cite{gao2024clip} & 75.2 & 72.1 & 35.8 & 61.3 & 78.8 & 91.2 & 93.7 & 81.7 & 95.6 & 57.9 & 77.9 & 74.7 \\
 & PLOT++ \cite{zhang2021tip} & 76.4 & 75.2 & 43.2 & 81.3 & 82.6 & 87.7 & 94.2 & 95.9 & 96.9 & 66.8 & 83.8 & 80.4 \\
 & KgCoOp \cite{yao2023visual} & 76.4 & 75.2 & 40.6 & 79.5 & 80.0 & \textbf{91.5} & 94.4 & 90.2 & 96.9 & 66.3 & 83.4 & 79.5 \\
 & TaskRes \cite{yu2023task} & 77.1 & 74.9 & 42.5 & 76.6 & 83.6 & 90.7 & 94.4 & 90.3 & 96.5 & 65.4 & 80.1 & 79.3 \\
 & MaPLe \cite{khattak2023maple} & 77.2 & 76.0 & 40.4 & 74.6 & 80.3 & \textbf{91.5} & \textbf{95.0} & 93.2 & 97.0 & 64.5 & 82.8 & 79.3 \\
 & ProGrad \cite{zhu2023prograd} & 76.5 & 75.0 & 44.6 & 79.3 & 83.8 & 90.6 & 94.8 & 95.6 & 96.8 & 66.3 & 83.6 & 80.6 \\
 & LP++ \cite{huang2024lp++} & 77.4 & 76.9 & 45.9 & 83.1 & 82.7 & 91.0 & 93.8 & 97.2 & \textbf{97.4} & 68.3 & 85.3 & 81.7 \\
 & CLIP-LoRA \cite{zanella2024cliplora} & \textbf{77.9} & 76.7 & \underline{48.9} & \underline{86.4} & \textbf{85.2} & 89.6 & 93.9 & \underline{97.4} & 97.2 & \underline{70.4} & \underline{86.4} & \underline{82.7} \\
 & MMA \cite{yang2024mma} & \underline{77.7} & \underline{77.1} & 45.2 & 55.3 & 83.3 & 91.4 & 94.3 & 95.1 & 97.0 & 63.8 & 83.2 & 78.5 \\
\rowcolor{lncolor}
\cellcolor{gray!0} & \ours{} & 77.3 & \textbf{77.5} & \textbf{52.0} & \textbf{86.7} & \underline{84.9} & 90.9 & \textbf{95.0} & \textbf{97.5} & \textbf{97.4} & \textbf{71.1} & \textbf{86.9} & \textbf{83.4} \\

\bottomrule
 \\ [-5ex] \\ 

    \end{tabular}
    \end{adjustbox}
\label{tab:all2all-4shots}
\end{table*}

\begin{table*}
    \def\arraystretch{1.1}
    \centering
    \small
    \caption{\emph{All-to-all} experiments with $\mathbf{k=8}$ shots, using ViT-B/16, ViT-B/32, and ViT-L/14. Formatting follows \cref{tab:all2all}.}
    \begin{adjustbox}{max width=\textwidth}
    \begin{tabular}{llccccccccccc|c}
    \toprule
     \textsc{\textbf{Backbone}} &  \textsc{\textbf{Method}} &  \textsc{\textbf{ImageNet}} &  \textsc{\textbf{SUN}} &  \textsc{\textbf{AIR}} &  \textsc{\textbf{ESAT}} &  \textsc{\textbf{CARS}} &  \textsc{\textbf{FOOD}} &  \textsc{\textbf{PETS}} &  \textsc{\textbf{FLWR}} &  \textsc{\textbf{CAL}} &  \textsc{\textbf{DTD}} &  \textsc{\textbf{UCF}} &  \textsc{\textbf{Mean}} \\
    \midrule
    
    \cellcolor{gray!0} \multirow{13}{*}{\texttt{ViT-B/16}} & \emph{Zero-Shot} \cite{radford2021learning} & 66.7 & 62.6 & 24.7 & 47.5 & 65.3 & 86.1 & 89.1 & 71.4 & 92.9 & 43.6 & 66.7 & 65.1 \\
 & CoOp \cite{zhou2022coop} (ctx=16) & 70.6 & 71.9 & 38.5 & 77.1 & 79.0 & 82.7 & 91.3 & 94.9 & 94.5 & 64.8 & 80.0 & 76.8 \\
 & CoCoOp \cite{zhou2022cocoop} & 70.8 & 71.5 & 32.4 & 69.1 & 70.4 & \underline{87.0} & \textbf{93.3} & 86.3 & 94.9 & 60.1 & 75.9 & 73.8 \\
 & TIP-Adapter-F \cite{zhang2021tip} & 71.7 & 73.5 & 39.5 & 81.3 & 78.3 & 86.9 & 91.8 & 94.3 & 95.2 & 66.7 & 82.0 & 78.3 \\
 & CLIP-Adapter \cite{gao2024clip} & 69.1 & 71.7 & 30.5 & 61.6 & 70.7 & 86.9 & 91.9 & 83.3 & 94.5 & 50.5 & 76.2 & 71.5 \\
 & PLOT++ \cite{chen2022plot} & 71.3 & 73.9 & 41.4 & 88.4 & 81.3 & 86.6 & 93.0 & 95.4 & 95.5 & 66.5 & 82.8 & 79.6 \\
 & KgCoOp \cite{yao2023visual} & 70.2 & 72.6 & 34.8 & 73.9 & 72.8 & \underline{87.0} & 93.0 & 91.5 & 95.1 & 65.6 & 80.0 & 76.0 \\
 & TaskRes \cite{yu2023task} & \underline{72.3} & 74.6 & 40.3 & 77.5 & 79.6 & 86.4 & 92.0 & \underline{96.0} & 95.3 & 66.7 & 81.6 & 78.4 \\
 & MaPLe \cite{khattak2023maple} & 71.3 & 73.2 & 33.8 & 82.8 & 71.3 & \textbf{87.2} & \underline{93.1} & 90.5 & 95.1 & 63.0 & 79.5 & 76.4 \\
 & ProGrad \cite{zhu2023prograd} & 71.3 & 73.0 & 37.7 & 77.8 & 78.7 & 86.1 & 92.2 & 95.0 & 94.8 & 63.9 & 80.5 & 77.4 \\
 & LP++ \cite{huang2024lp++} & 72.1 & \underline{75.1} & 39.0 & 78.2 & 76.4 & 86.8 & 91.8 & 95.2 & 95.5 & \underline{67.7} & 81.9 & 78.2 \\
 & CLIP-LoRA \cite{zanella2024cliplora} & \underline{72.3} & 74.7 & \textbf{45.7} & \textbf{89.7} & \textbf{82.1} & 83.1 & 91.7 & \textbf{96.3} & \underline{95.6} & 67.5 & \underline{84.1} & \underline{80.3} \\
 & MMA \cite{yang2024mma} & 71.9 & 74.7 & 38.9 & 69.7 & 76.8 & 86.4 & 92.9 & 94.6 & \underline{95.6} & 66.9 & 82.9 & 77.4 \\
 \rowcolor{lncolor}
\cellcolor{gray!0} & \ours{} & \textbf{72.5} & \textbf{75.5} & \underline{44.3} & \underline{89.1} & \underline{81.9} & 86.1 & 92.9 & 95.9 & \textbf{96.1} & \textbf{68.7} & \textbf{84.4} & \textbf{80.7} \\
 \\ [-5ex] \\ 
    \cmidrule(lr){1-14}
    \cellcolor{gray!0} \multirow{13}{*}{\texttt{ViT-B/32}} & \emph{Zero-Shot} \cite{radford2021learning} & 61.9 & 62.0 & 19.3 & 45.1 & 60.4 & 80.5 & 87.5 & 67.0 & 91.1 & 42.6 & 62.2 & 61.8 \\
 & CoOp \cite{zhou2022coop} (ctx=16) & 65.5 & 69.2 & 29.1 & 76.4 & 71.3 & 76.3 & 87.4 & 92.7 & 93.8 & 61.7 & 76.5 & 72.7 \\
 & CoCoOp \cite{zhou2022cocoop} & 65.8 & 68.9 & 20.3 & 58.1 & 63.4 & 81.6 & 90.1 & 77.3 & 93.8 & 57.4 & 72.4 & 68.1 \\
 & TIP-Adapter-F \cite{zhang2021tip} & 66.8 & 71.2 & 32.1 & 75.0 & 72.6 & 81.3 & 89.8 & 90.4 & 94.5 & 63.6 & 78.0 & 74.1 \\
 & CLIP-Adapter \cite{gao2024clip} & 64.2 & 69.3 & 23.5 & 55.2 & 65.4 & 81.5 & 89.3 & 78.0 & 93.9 & 50.8 & 73.0 & 67.6 \\
 & PLOT++ \cite{chen2022plot} & 66.2 & 71.0 & 31.7 & 87.1 & 73.5 & 78.2 & 88.4 & \textbf{93.8} & 94.4 & 62.9 & 79.1 & 75.1 \\
 & KgCoOp \cite{yao2023visual} & 65.1 & 69.5 & 24.7 & 66.2 & 65.0 & \underline{81.7} & 90.3 & 83.1 & 94.5 & 61.1 & 74.7 & 70.5 \\
 & TaskRes \cite{yu2023task} & \textbf{67.4} & 71.9 & 31.9 & 74.9 & 73.8 & 80.6 & 89.1 & \underline{93.5} & \underline{94.8} & \underline{64.5} & 78.4 & 74.6 \\
 & MaPLe \cite{khattak2023maple} & 66.3 & 70.3 & 25.4 & 79.0 & 63.7 & \textbf{81.9} & \underline{90.9} & 81.1 & 94.4 & 59.8 & 75.0 & 71.6 \\
 & ProGrad \cite{zhu2023prograd} & 66.1 & 71.1 & 29.0 & 73.5 & 71.8 & 80.0 & 89.1 & 92.1 & 94.2 & 62.3 & 75.7 & 73.2 \\
 & LP++ \cite{huang2024lp++} & 67.1 & \underline{72.2} & 30.3 & 78.8 & 71.2 & 81.5 & 89.3 & 92.4 & 94.6 & 64.2 & 78.4 & 74.5 \\
 & CLIP-LoRA \cite{zanella2024cliplora} & \underline{67.2} & 72.1 & \textbf{36.1} & \textbf{88.8} & \underline{74.4} & 76.7 & 87.7 & 92.4 & \underline{94.8} & 63.7 & \underline{80.1} & \underline{75.8} \\
 & MMA \cite{yang2024mma} & 66.7 & \underline{72.2} & 29.6 & 56.2 & 70.4 & 81.0 & \textbf{91.0} & 90.7 & 94.6 & 64.4 & 78.7 & 72.3 \\

\rowcolor{lncolor}
\cellcolor{gray!0} & \ours{} & \underline{67.2} & \textbf{73.1} & \underline{35.2} & \underline{88.7} & \textbf{75.4} & 80.4 & 90.4 & 93.4 & \textbf{95.4} & \textbf{65.9} & \textbf{80.2} & \textbf{76.8} \\
 \\ [-5ex] \\ 
    \cmidrule(lr){1-14}
    \cellcolor{gray!0} \multirow{13}{*}{\texttt{ViT-L/14}} & \emph{Zero-Shot} \cite{radford2021learning} & 72.9 & 67.6 & 32.6 & 58.0 & 76.8 & 91.0 & 93.6 & 79.4 & 94.9 & 53.6 & 74.2 & 72.2 \\
 & CoOp \cite{zhou2022coop} (ctx=16) & 76.8 & 75.0 & 51.2 & 82.8 & 86.4 & 88.6 & 94.0 & \underline{98.0} & 96.7 & 69.4 & 85.1 & 82.2 \\
 & CoCoOp \cite{zhou2022cocoop} & 77.4 & 75.6 & 43.3 & 77.0 & 81.4 & \textbf{91.6} & \textbf{95.3} & 93.0 & 97.0 & 67.9 & 84.5 & 80.4 \\
 & TIP-Adapter-F \cite{zhang2021tip} & 77.8 & 76.7 & 50.4 & 84.9 & 85.9 & 91.4 & 94.1 & 97.3 & 96.9 & 71.2 & 86.2 & 83.0 \\
 & CLIP-Adapter \cite{gao2024clip} & 75.7 & 75.9 & 40.7 & 67.9 & 81.6 & 91.4 & 94.3 & 92.3 & 96.8 & 63.8 & 82.8 & 78.5 \\
 & PLOT++ \cite{chen2022plot} & 77.8 & 77.0 & 43.2 & 87.0 & 84.6 & 89.6 & 93.3 & 96.3 & 96.8 & 69.5 & 84.8 & 81.8 \\
 & KgCoOp \cite{yao2023visual} & 76.7 & 76.2 & 45.9 & 82.1 & 82.3 & \textbf{91.6} & 95.1 & 95.2 & \underline{97.3} & 70.8 & 85.7 & 81.7 \\
 & TaskRes \cite{yu2023task} & 77.9 & 76.0 & 51.1 & 81.1 & 85.7 & 91.1 & 94.5 & 96.7 & 96.9 & 69.4 & 85.6 & 82.4 \\
 & MaPLe \cite{khattak2023maple} & 78.0 & 77.2 & 42.9 & 80.7 & 81.8 & 90.1 & 95.0 & 95.8 & 96.8 & 69.5 & 85.1 & 81.2 \\
 & ProGrad \cite{zhu2023prograd} & 77.7 & 76.1 & 49.9 & 83.6 & 86.2 & 90.8 & 95.1 & 97.8 & 96.7 & 69.9 & 85.4 & 82.7 \\
 & LP++ \cite{huang2024lp++} & 78.4 & 78.4 & 50.8 & 85.0 & 85.2 & 91.4 & 94.4 & 97.9 & \textbf{97.6} & 72.1 & 86.0 & 83.4 \\
 & CLIP-LoRA \cite{zanella2024cliplora} & 78.5 & 78.0 & \underline{57.5} & \textbf{90.0} & \textbf{88.7} & 89.7 & 94.2 & \underline{98.0} & 97.0 & \underline{72.2} & \underline{88.3} & \underline{84.7} \\
 & MMA \cite{yang2024mma} & \textbf{78.6} & \underline{78.8} & 50.9 & 61.4 & 85.8 & 91.5 & 95.1 & 97.7 & 97.1 & 71.9 & 86.2 & 81.4 \\

 \rowcolor{lncolor}
\cellcolor{gray!0} & \ours{} & \textbf{78.6} & \textbf{79.2} & \textbf{57.6} & \underline{89.8} & \underline{88.2} & 91.4 & \underline{95.2} & \textbf{98.3} & 97.2 & \textbf{74.2} & \textbf{88.4} & \textbf{85.3} \\

\bottomrule
 \\ [-5ex] \\ 

    \end{tabular}
    \end{adjustbox}
\label{tab:all2all-8shots}
\end{table*}

Results for the all-to-all setup are given in \cref{tab:all2all-4shots} and \cref{tab:all2all-8shots} for 4 and 8 shots. 
We include numbers from all 11 competitors of \cref{sec:all2all}, following the reported results of \cite{zanella2024cliplora}, and reproducing when unavailable.
Also in this case, \emph{\ours{} outperforms all competitors on average for all \{backbones, shots\} combinations}.

In summary, looking at both scenarios, \ours{} appears to be a stronger approach \wrt to the comparison suite, regardless of how many shots are available. 
Importantly, it does so by (i) not employing any external source of knowledge (such as LLMs to generate descriptions or Image Generators to craft new examples \cite{da2023diversified}), (ii) avoiding the usage of well-engineered templates for each dataset, which are likely to be unavailable in practice, and (iii) only leveraging a single template ``\emph{a photo of a \{\}}'', in contrast to an ensemble of templates \cite{khattak2023self}.
We speculate, however, that such orthogonal techniques may further improve \ours{}.

\section{Total gradient steps allowed}
\label{app:iteranalysis}
This Appendix briefly analyzes the impact of increasing or reducing the total number of iterations $m$.
For simplicity, we stick to the ViT-B/16 backbone and the ImageNet validation set.
Recall that in \cref{sec:experiments}, the total number of iterations is defined as $m = M \times k$, where, in our case, $M=300$ and $k$ is the number of shots. 
$M$ was chosen so to match the number of gradient steps performed on ImageNet with $\approx$ 10 epochs (constant mini-batch size of 32, 16 shots for all categories).
In essence, this means that the budget is expressed in terms of a constant number of gradient steps rather than epochs, following \cite{zanella2024cliplora}.
Here, we analyze the impact of varying $M$ when $\alpha{=}0.6$ as in \cref{sec:experiments}.
Results are given in \cref{tab:sweep-on-M}.
We observe that $M=100$ likely allocates insufficient compute for learning a good feature extractor in the first stage (lowest ``Novel'' metric).
In contrast, $M=300$ and $M=500$ exhibit more comparable behaviors, which leads to choosing $M=300$ considering the reduced overall runtime.  

\section{Extended preliminary analysis}
\label{app:twostages}
\begin{figure*}
     \centering
     \begin{subfigure}[b]{\linewidth}
         \centering
         \includegraphics[width=\textwidth]{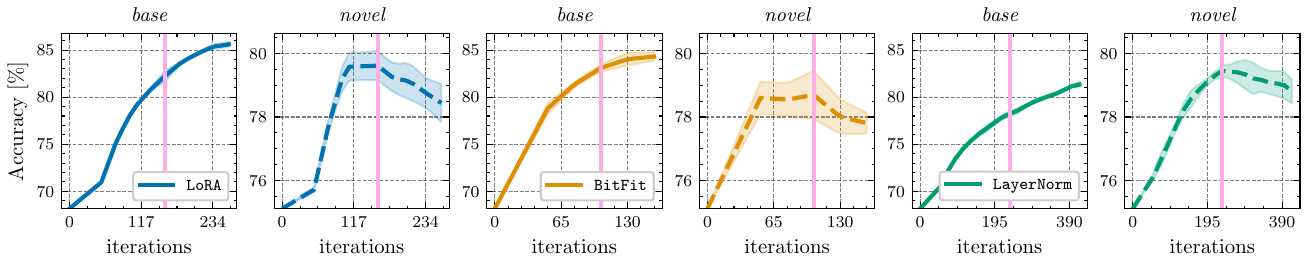}
         \caption{UCF-101}
         \label{fig:prel-ucf101}
     \end{subfigure}
     \hfill
     \begin{subfigure}[b]{\linewidth}
         \centering
         \includegraphics[width=\textwidth]{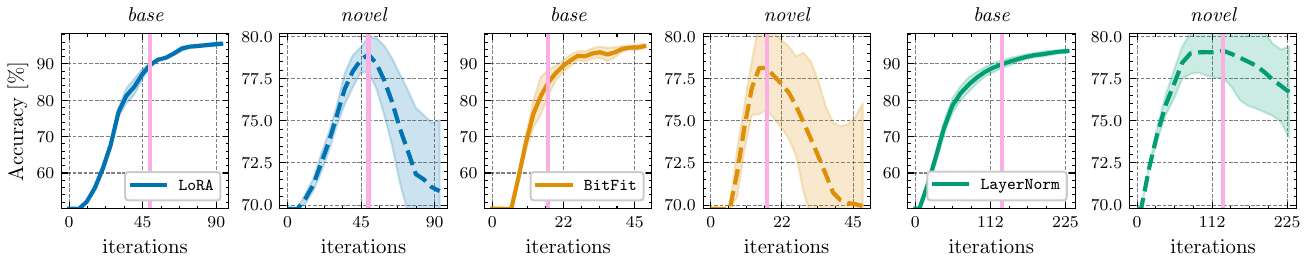}
         \caption{EuroSAT}
         \label{fig:prel-eurosat}
     \end{subfigure}
     \caption{Breakpoints consistently emerging for UCF-101 \cite{soomro2012ucf101} and EuroSAT \cite{helber2019eurosat}, regardless of the PEFT technique used in our study. The pattern appears particularly evident with EuroSAT (bottom).}
     \label{fig:breakpoint-confirmed}
\end{figure*}

\begin{figure*}
     \centering
     \begin{subfigure}[b]{\linewidth}
         \centering
         \includegraphics[width=\textwidth]{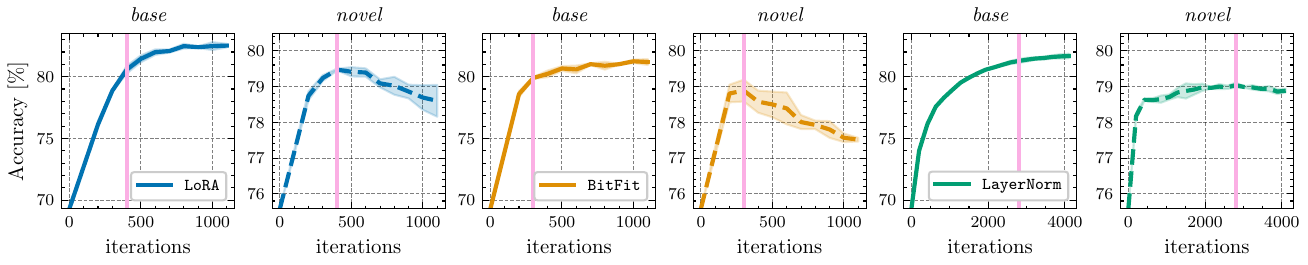}
         \caption{SUN397}
         \label{fig:prel-sun397}
     \end{subfigure}
     \hfill
     \begin{subfigure}[b]{\linewidth}
         \centering
         \includegraphics[width=\textwidth]{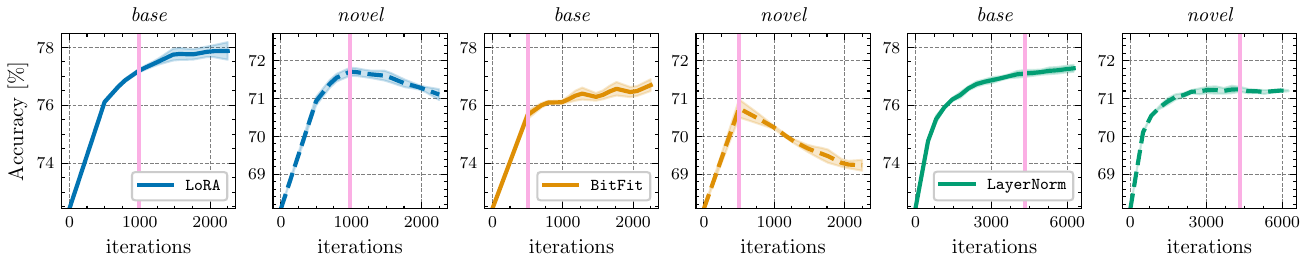}
         \caption{ImageNet}
         \label{fig:prel-imagenet}
     \end{subfigure}
     \caption{Breakpoint further confirmed for both LoRA \cite{hu2022lora} and BitFit \cite{zaken2022bitfit} on ImageNet \cite{russakovsky2015imagenet} and SUN397 \cite{xiao2010sun}. For Layer Normalization, we speculate that the more balanced data-to-parameter ratio, given the larger number of examples in these datasets and the smaller number of parameters of LayerNorm, has a regularizing effect, which avoids breaking and leads to saturation.}
     \label{fig:breakpoint-ln-saturates}
\end{figure*}

\begin{figure*}
     \centering
     \begin{subfigure}[b]{\linewidth}
         \centering
         \includegraphics[width=\textwidth]{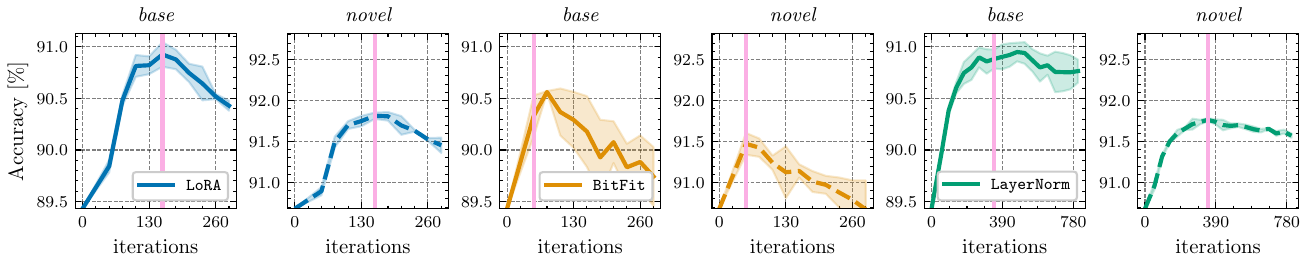}
         \caption{Food-101}
         \label{fig:prel-food101}
     \end{subfigure}
     \hfill
     \begin{subfigure}[b]{\linewidth}
         \centering
         \includegraphics[width=\textwidth]{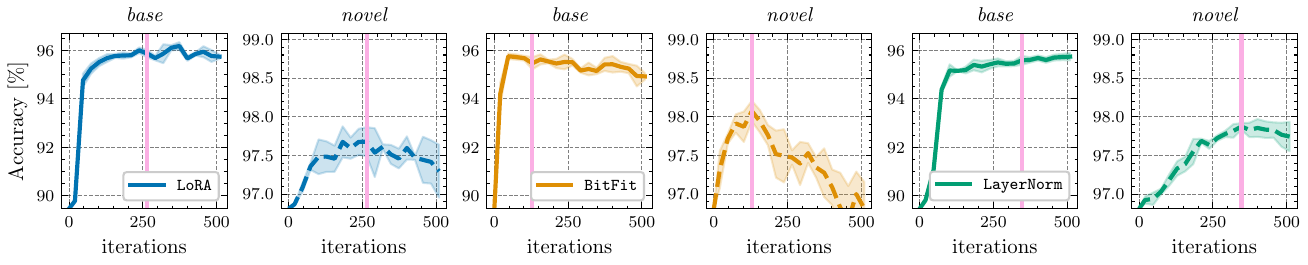}
         \caption{Oxford Pets}
         \label{fig:prel-pets}
     \end{subfigure}
     \caption{Understanding the failure cases of \cref{sec:post-analysis} through the lens of breakpoints. On Oxford Pets \cite{parkhi12a} and Food-101 \cite{bossard2014food}, base accuracy overfits or saturates right after degradation on novel accuracy, which leads the second stage of \ours{} to train a classifier on disrupted base features since $\alpha$ is fixed. These visualizations suggest that $\alpha$ and $M$ should be tuned explicitly for these benchmarks, which we avoid to strive for an evaluation as realistic as possible.}
     \label{fig:breakpoint-failures}
\end{figure*}

Here, we aim to enrich the preliminary analysis conducted in \cref{sec:pre-analysis}.
Recall that \cref{sec:pre-analysis} introduces the natural emergence of two distinct stages when training CLIP ViT-B/16 with three different PEFT techniques in the low-data regime of FSA, and does so by visualizing the learning dynamics on DTD \cite{cimpoi14describing} and FGVC Aircraft \cite{maji13fine-grained}.
First, we show that such a dynamic is not limited to those datasets.
Second, we identify a \emph{saturating} behavior of Layer Normalization, which we link to the data-to-parameter ratio.
Finally, we focus on Oxford Pets \cite{parkhi12a} and Food-101 \cite{bossard2014food}, which were the only datasets (out of 11) leading to a slight performance degradation during the ablation study of \cref{sec:post-analysis}.

\mypar{Consistent behaviors.}
\cref{fig:breakpoint-confirmed} shows that analogous and consistent patterns emerge also for UCF-101 \cite{soomro2012ucf101} and EuroSAT \cite{helber2019eurosat} for all the PEFT techniques of our study.
Particularly with EuroSAT, this behavior emerges to the extreme, with sharp breakpoints.
In line with \cref{sec:pre-analysis}, BitFit tends to ``break'' earlier than both LoRA and LayerNorm.

\mypar{Saturating behaviors.} \cref{fig:breakpoint-ln-saturates} shows consistent breakpoints for LoRA and BitFit further, displaying SUN397 \cite{xiao2010sun} and ImageNet \cite{russakovsky2015imagenet}.
These two datasets have a trait in common \wrt the rest of the evaluation suite: a much larger label space.
In FSA, where samples are constant per category, this inevitably entails a larger amount of examples.
In parallel, LayerNorm instances total a reduced number of parameters \wrt to LoRA and BitFit (61k, 184k, 125k, respectively).
We speculate that the more balanced data-to-parameter ratio of LayerNorm for these larger datasets has a regularizing effect, which avoids breaking and reaches a behavior similar to saturation, where the novel class accuracy remains constant.

\mypar{Unexpected behaviors.} \cref{fig:breakpoint-failures} depicts the learning dynamics on Food-101 \cite{bossard2014food} and Oxford Pets \cite{parkhi12a}.
These were the only two datasets where including a second stage did not appear beneficial in \cref{fig:1v2ablation} of the main body.
From the dynamics, the reason is evident: base and novel accuracy break together.
For both datasets, \emph{base} accuracy either decreases or saturates right after the breakpoint (pink line), implying
overfitting since training data are available for base categories only.
This suggests that $\alpha$ and $M$ should be tailored to these datasets, to avoid training a classifier on overfitted features.
However, we consider it fairer to transfer hyperparameters across datasets since, in practice, no annotated data except for the shots should be available in FSA, which raises concerns about the feasibility of tuning hyperparameters per dataset.

\section{Limitations}
\label{app:limitations}
In this work, we build on the finding that PEFT techniques learn good task-level features to design a simple and effective strategy for few-shot adaptation. 
For completeness, we identify and report three limitations of our work, which we hope can help construct future works.

\mypar{Evaluating outside of our suite.} 
While we successfully experiment with a variety of backbones (\ie, ViT-B/16, ViT-B/32, ViT-L/14), datasets (\ie, 11 different benchmarks), settings (\ie, base-to-novel, all-to-all), PEFT techniques (\ie, LayerNorm tuning and LoRA), and data availability conditions (\ie, 4, 8, and 16 shots), as per most empirical observations, our results might not extend when tested with other (or future) PEFT strategies and on different benchmarks or additional models. 

\mypar{Expanding the variety of tasks.} Our work focuses on downstream classification, following the established and recent field literature ~\cite{zhou2022coop, zhou2022cocoop, khattak2023maple, yao2023visual, gao2024clip, zanella2024cliplora, zhu2023prograd, yu2023task, chen2022plot, yang2024mma, zhang2021tip}. 
However, an additional intriguing direction to pursue is represented by tasks focusing on different challenges (\eg, the spatial ones of semantic segmentation, and the temporal one of action recognition), which may require different adaptation strategies. 

\mypar{Validation-free stopping criterion.} Finally, a core hyperparameter of our approach is $\alpha$, regulating when to stop with the feature extractor training (\ie, the first stage) and start with the second one (\ie, classifier learning). 
As we have shown empirically with Oxford Pets \cite{parkhi12a} and Food-101 \cite{bossard2014food} a single $\alpha$, tuned on a given dataset, may not be ideal for others. 
To this aim, future works may integrate (or investigate) stopping criteria not requiring a validation set~\cite{mahsereci2017early}, to dynamically understand or approximate, in an unsupervised manner, when to switch between the two stages. 

\begin{table*}
\caption{Experiments in \emph{base-to-novel} generalization with the ViT-B/16 visual backbone $\mathbf{k{=}4}$ shots per base class.}
\label{tab:b2n_vitb16_shots4}
\def\arraystretch{1.075}
\centering
\scriptsize
\begin{minipage}{0.32\textwidth}
\centering 
\caption*{\textbf{Average across datasets.}}\vspace{-1em} 
\begin{tabular}{lcc|c} 
\toprule 
\textbf{Method} & \textbf{Base} & \textbf{Novel} & \textbf{HM} \\ 
\midrule 
CLIP \cite{radford2021learning} & 69.34 & 74.22 & 71.70 \\ 
MMA \cite{yang2024mma} & 80.13 & 78.57 & 74.90 \\ 
\cmidrule(lr){1-4} 
\rowcolor{lncolor}
\ours{} & 84.64 & 78.53 & \textbf{78.88} \\ 
\bottomrule 
\end{tabular} 
\vspace{0.5em} 
\end{minipage} 
\hfill
\begin{minipage}{0.32\textwidth}
\centering 
\caption*{ImageNet}\vspace{-1em} 
\begin{tabular}{lcc|c} 
\toprule 
\textbf{Method} & \textbf{Base} & \textbf{Novel} & \textbf{HM} \\ 
\midrule 
CLIP \cite{radford2021learning} & 72.43 & 68.14 & 70.22 \\ 
MMA \cite{yang2024mma} & 75.37 & 70.10 & 72.64 \\ 
\cmidrule(lr){1-4} 
\rowcolor{lncolor}
\ours{} & 75.68 & 70.27 & \textbf{72.87} \\ 
\bottomrule 
\end{tabular} 
\vspace{0.5em} 
\end{minipage} 
\hfill
\begin{minipage}{0.32\textwidth}
\centering 
\caption*{Caltech101}\vspace{-1em} 
\begin{tabular}{lcc|c} 
\toprule 
\textbf{Method} & \textbf{Base} & \textbf{Novel} & \textbf{HM} \\ 
\midrule 
CLIP \cite{radford2021learning} & 96.84 & 94.00 & 93.73 \\ 
MMA \cite{yang2024mma} & 97.33 & 94.57 & 95.93 \\ 
\cmidrule(lr){1-4} 
\rowcolor{lncolor}
\ours{} & 98.13 & 94.21 & \textbf{96.13} \\ 
\bottomrule 
\end{tabular} 
\vspace{0.5em} 
\end{minipage} 
\hfill
\begin{minipage}{0.32\textwidth}
\centering 
\caption*{Oxford Flowers}\vspace{-1em} 
\begin{tabular}{lcc|c} 
\toprule 
\textbf{Method} & \textbf{Base} & \textbf{Novel} & \textbf{HM} \\ 
\midrule 
CLIP \cite{radford2021learning} & 72.08 & 77.80 & 74.83 \\ 
MMA \cite{yang2024mma} & 91.07 & 75.07 & 82.30 \\ 
\cmidrule(lr){1-4} 
\rowcolor{lncolor}
\ours{} & 94.94 & 76.26 & \textbf{84.58} \\ 
\bottomrule 
\end{tabular} 
\vspace{0.5em} 
\end{minipage} 
\hfill
\begin{minipage}{0.32\textwidth}
\centering 
\caption*{Oxford Pets}\vspace{-1em} 
\begin{tabular}{lcc|c} 
\toprule 
\textbf{Method} & \textbf{Base} & \textbf{Novel} & \textbf{HM} \\ 
\midrule 
CLIP \cite{radford2021learning} & 91.17 & 97.26 & 94.12 \\ 
MMA \cite{yang2024mma} & 91.30 & 97.07 & 94.10 \\ 
\cmidrule(lr){1-4} 
\rowcolor{lncolor}
\ours{} & 94.58 & 97.48 & \textbf{96.01} \\ 
\bottomrule 
\end{tabular} 
\vspace{0.5em} 
\end{minipage} 
\hfill
\begin{minipage}{0.32\textwidth}
\centering 
\caption*{Stanford Cars}\vspace{-1em} 
\begin{tabular}{lcc|c} 
\toprule 
\textbf{Method} & \textbf{Base} & \textbf{Novel} & \textbf{HM} \\ 
\midrule 
CLIP \cite{radford2021learning} & 63.37 & 74.89 & 68.65 \\ 
MMA \cite{yang2024mma} & 71.30 & 74.07 & 72.66 \\ 
\cmidrule(lr){1-4} 
\rowcolor{lncolor}
\ours{} & 74.45 & 75.98 & \textbf{75.21} \\ 
\bottomrule 
\end{tabular} 
\vspace{0.5em} 
\end{minipage} 
\hfill
\begin{minipage}{0.32\textwidth}
\centering 
\caption*{Food101}\vspace{-1em} 
\begin{tabular}{lcc|c} 
\toprule 
\textbf{Method} & \textbf{Base} & \textbf{Novel} & \textbf{HM} \\ 
\midrule 
CLIP \cite{radford2021learning} & 90.10 & 91.22 & \textbf{90.66} \\ 
MMA \cite{yang2024mma} & 89.77 & 91.10 & 90.43 \\ 
\cmidrule(lr){1-4} 
\rowcolor{lncolor}
\ours{} & 88.91 & 91.40 & 90.14 \\ 
\bottomrule 
\end{tabular} 
\vspace{0.5em} 
\end{minipage} 
\hfill
\begin{minipage}{0.32\textwidth}
\centering 
\caption*{FGVC Aircraft}\vspace{-1em} 
\begin{tabular}{lcc|c} 
\toprule 
\textbf{Method} & \textbf{Base} & \textbf{Novel} & \textbf{HM} \\ 
\midrule 
CLIP \cite{radford2021learning} & 27.19 & 36.29 & 31.09 \\ 
MMA \cite{yang2024mma} & 31.97 & 34.03 & 32.97 \\ 
\cmidrule(lr){1-4} 
\rowcolor{lncolor}
\ours{} & 39.36 & 35.67 & \textbf{37.42} \\ 
\bottomrule 
\end{tabular} 
\vspace{0.5em} 
\end{minipage} 
\hfill
\begin{minipage}{0.32\textwidth}
\centering 
\caption*{SUN397}\vspace{-1em} 
\begin{tabular}{lcc|c} 
\toprule 
\textbf{Method} & \textbf{Base} & \textbf{Novel} & \textbf{HM} \\ 
\midrule 
CLIP \cite{radford2021learning} & 69.36 & 75.35 & 72.23 \\ 
MMA \cite{yang2024mma} & 79.37 & 78.53 & 78.95 \\ 
\cmidrule(lr){1-4} 
\rowcolor{lncolor}
\ours{} & 80.23 & 78.46 & \textbf{79.33} \\ 
\bottomrule 
\end{tabular} 
\vspace{0.5em} 
\end{minipage} 
\hfill
\begin{minipage}{0.32\textwidth}
\centering 
\caption*{DTD}\vspace{-1em} 
\begin{tabular}{lcc|c} 
\toprule 
\textbf{Method} & \textbf{Base} & \textbf{Novel} & \textbf{HM} \\ 
\midrule 
CLIP \cite{radford2021learning} & 53.24 & 59.90 & 56.37 \\ 
MMA \cite{yang2024mma} & 63.50 & 63.57 & 63.53 \\ 
\cmidrule(lr){1-4} 
\rowcolor{lncolor}
\ours{} & 76.81 & 65.02 & \textbf{70.42} \\ 
\bottomrule 
\end{tabular} 
\vspace{0.5em} 
\end{minipage} 
\hfill
\begin{minipage}{0.32\textwidth}
\centering 
\caption*{EuroSAT}\vspace{-1em} 
\begin{tabular}{lcc|c} 
\toprule 
\textbf{Method} & \textbf{Base} & \textbf{Novel} & \textbf{HM} \\ 
\midrule 
CLIP \cite{radford2021learning} & 56.48 & 64.05 & 60.03 \\ 
MMA \cite{yang2024mma} & 50.13 & 69.93 & 58.40 \\ 
\cmidrule(lr){1-4} 
\rowcolor{lncolor}
\ours{} & 91.29 & 75.10 & \textbf{82.41} \\ 
\bottomrule 
\end{tabular} 
\vspace{0.5em} 
\end{minipage} 
\hfill
\begin{minipage}{0.32\textwidth}
\centering 
\caption*{UCF101}\vspace{-1em} 
\begin{tabular}{lcc|c} 
\toprule 
\textbf{Method} & \textbf{Base} & \textbf{Novel} & \textbf{HM} \\ 
\midrule 
CLIP \cite{radford2021learning} & 70.53 & 77.50 & 73.85 \\ 
MMA \cite{yang2024mma} & 80.13 & 78.57 & 79.34 \\ 
\cmidrule(lr){1-4} 
\rowcolor{lncolor}
\ours{} & 84.64 & 78.53 & \textbf{81.47} \\ 
\bottomrule 
\end{tabular} 
\vspace{0.5em} 
\end{minipage} 
\hfill
\end{table*}

\begin{table*}
\caption{Experiments in \emph{base-to-novel} generalization with the ViT-B/16 visual backbone $\mathbf{k{=}8}$ shots per base class.}
\label{tab:b2n_vitb16_shots8}
\def\arraystretch{1.075}
\centering
\scriptsize
\begin{minipage}{0.32\textwidth}
\centering 
\caption*{\textbf{Average across datasets.}}\vspace{-1em} 
\begin{tabular}{lcc|c} 
\toprule 
\textbf{Method} & \textbf{Base} & \textbf{Novel} & \textbf{HM} \\ 
\midrule 
CLIP \cite{radford2021learning} & 69.34 & 74.22 & 71.70 \\ 
MMA \cite{yang2024mma} & 84.80 & 78.10 & 76.68 \\ 
\cmidrule(lr){1-4} 
\rowcolor{lncolor}
\ours{} & 86.37 & 78.58 & \textbf{79.74} \\ 
\bottomrule 
\end{tabular} 
\vspace{0.5em} 
\end{minipage} 
\hfill
\begin{minipage}{0.32\textwidth}
\centering 
\caption*{ImageNet}\vspace{-1em} 
\begin{tabular}{lcc|c} 
\toprule 
\textbf{Method} & \textbf{Base} & \textbf{Novel} & \textbf{HM} \\ 
\midrule 
CLIP \cite{radford2021learning} & 72.43 & 68.14 & 70.22 \\ 
MMA \cite{yang2024mma} & 76.43 & 70.07 & 73.11 \\ 
\cmidrule(lr){1-4} 
\rowcolor{lncolor}
\ours{} & 76.97 & 70.67 & \textbf{73.69} \\ 
\bottomrule 
\end{tabular} 
\vspace{0.5em} 
\end{minipage} 
\hfill
\begin{minipage}{0.32\textwidth}
\centering 
\caption*{Caltech101}\vspace{-1em} 
\begin{tabular}{lcc|c} 
\toprule 
\textbf{Method} & \textbf{Base} & \textbf{Novel} & \textbf{HM} \\ 
\midrule 
CLIP \cite{radford2021learning} & 96.84 & 94.00 & 93.73 \\ 
MMA \cite{yang2024mma} & 97.77 & 93.87 & 95.78 \\ 
\cmidrule(lr){1-4} 
\rowcolor{lncolor}
\ours{} & 98.13 & 94.18 & \textbf{96.11} \\ 
\bottomrule 
\end{tabular} 
\vspace{0.5em} 
\end{minipage} 
\hfill
\begin{minipage}{0.32\textwidth}
\centering 
\caption*{Oxford Flowers}\vspace{-1em} 
\begin{tabular}{lcc|c} 
\toprule 
\textbf{Method} & \textbf{Base} & \textbf{Novel} & \textbf{HM} \\ 
\midrule 
CLIP \cite{radford2021learning} & 72.08 & 77.80 & 74.83 \\ 
MMA \cite{yang2024mma} & 95.37 & 75.43 & 84.24 \\ 
\cmidrule(lr){1-4} 
\rowcolor{lncolor}
\ours{} & 96.68 & 76.36 & \textbf{85.33} \\ 
\bottomrule 
\end{tabular} 
\vspace{0.5em} 
\end{minipage} 
\hfill
\begin{minipage}{0.32\textwidth}
\centering 
\caption*{Oxford Pets}\vspace{-1em} 
\begin{tabular}{lcc|c} 
\toprule 
\textbf{Method} & \textbf{Base} & \textbf{Novel} & \textbf{HM} \\ 
\midrule 
CLIP \cite{radford2021learning} & 91.17 & 97.26 & 94.12 \\ 
MMA \cite{yang2024mma} & 94.90 & 97.50 & 96.18 \\ 
\cmidrule(lr){1-4} 
\rowcolor{lncolor}
\ours{} & 94.95 & 97.80 & \textbf{96.35} \\ 
\bottomrule 
\end{tabular} 
\vspace{0.5em} 
\end{minipage} 
\hfill
\begin{minipage}{0.32\textwidth}
\centering 
\caption*{Stanford Cars}\vspace{-1em} 
\begin{tabular}{lcc|c} 
\toprule 
\textbf{Method} & \textbf{Base} & \textbf{Novel} & \textbf{HM} \\ 
\midrule 
CLIP \cite{radford2021learning} & 63.37 & 74.89 & 68.65 \\ 
MMA \cite{yang2024mma} & 75.00 & 74.80 & 74.90 \\ 
\cmidrule(lr){1-4} 
\rowcolor{lncolor}
\ours{} & 78.99 & 75.45 & \textbf{77.18} \\ 
\bottomrule 
\end{tabular} 
\vspace{0.5em} 
\end{minipage} 
\hfill
\begin{minipage}{0.32\textwidth}
\centering 
\caption*{Food101}\vspace{-1em} 
\begin{tabular}{lcc|c} 
\toprule 
\textbf{Method} & \textbf{Base} & \textbf{Novel} & \textbf{HM} \\ 
\midrule 
CLIP \cite{radford2021learning} & 90.10 & 91.22 & \textbf{90.66} \\ 
MMA \cite{yang2024mma} & 89.53 & 91.07 & 90.29 \\ 
\cmidrule(lr){1-4} 
\rowcolor{lncolor}
\ours{} & 89.02 & 91.36 & 90.17 \\ 
\bottomrule 
\end{tabular} 
\vspace{0.5em} 
\end{minipage} 
\hfill
\begin{minipage}{0.32\textwidth}
\centering 
\caption*{FGVC Aircraft}\vspace{-1em} 
\begin{tabular}{lcc|c} 
\toprule 
\textbf{Method} & \textbf{Base} & \textbf{Novel} & \textbf{HM} \\ 
\midrule 
CLIP \cite{radford2021learning} & 27.19 & 36.29 & 31.09 \\ 
MMA \cite{yang2024mma} & 37.53 & 34.57 & 35.99 \\ 
\cmidrule(lr){1-4} 
\rowcolor{lncolor}
\ours{} & 42.96 & 36.39 & \textbf{39.40} \\ 
\bottomrule 
\end{tabular} 
\vspace{0.5em} 
\end{minipage} 
\hfill
\begin{minipage}{0.32\textwidth}
\centering 
\caption*{SUN397}\vspace{-1em} 
\begin{tabular}{lcc|c} 
\toprule 
\textbf{Method} & \textbf{Base} & \textbf{Novel} & \textbf{HM} \\ 
\midrule 
CLIP \cite{radford2021learning} & 69.36 & 75.35 & 72.23 \\ 
MMA \cite{yang2024mma} & 80.73 & 78.33 & 79.51 \\ 
\cmidrule(lr){1-4} 
\rowcolor{lncolor}
\ours{} & 81.25 & 78.76 & \textbf{79.99} \\ 
\bottomrule 
\end{tabular} 
\vspace{0.5em} 
\end{minipage} 
\hfill
\begin{minipage}{0.32\textwidth}
\centering 
\caption*{DTD}\vspace{-1em} 
\begin{tabular}{lcc|c} 
\toprule 
\textbf{Method} & \textbf{Base} & \textbf{Novel} & \textbf{HM} \\ 
\midrule 
CLIP \cite{radford2021learning} & 53.24 & 59.90 & 56.37 \\ 
MMA \cite{yang2024mma} & 77.07 & 64.47 & 70.21 \\ 
\cmidrule(lr){1-4} 
\rowcolor{lncolor}
\ours{} & 80.17 & 64.45 & \textbf{71.46} \\ 
\bottomrule 
\end{tabular} 
\vspace{0.5em} 
\end{minipage} 
\hfill
\begin{minipage}{0.32\textwidth}
\centering 
\caption*{EuroSAT}\vspace{-1em} 
\begin{tabular}{lcc|c} 
\toprule 
\textbf{Method} & \textbf{Base} & \textbf{Novel} & \textbf{HM} \\ 
\midrule 
CLIP \cite{radford2021learning} & 56.48 & 64.05 & 60.03 \\ 
MMA \cite{yang2024mma} & 50.30 & 69.97 & 58.53 \\ 
\cmidrule(lr){1-4} 
\rowcolor{lncolor}
\ours{} & 93.33 & 75.09 & \textbf{83.23} \\ 
\bottomrule 
\end{tabular} 
\vspace{0.5em} 
\end{minipage} 
\hfill
\begin{minipage}{0.32\textwidth}
\centering 
\caption*{UCF101}\vspace{-1em} 
\begin{tabular}{lcc|c} 
\toprule 
\textbf{Method} & \textbf{Base} & \textbf{Novel} & \textbf{HM} \\ 
\midrule 
CLIP \cite{radford2021learning} & 70.53 & 77.50 & 73.85 \\ 
MMA \cite{yang2024mma} & 84.80 & 78.10 & 81.31 \\ 
\cmidrule(lr){1-4} 
\rowcolor{lncolor}
\ours{} & 86.37 & 78.58 & \textbf{82.29} \\ 
\bottomrule 
\end{tabular} 
\vspace{0.5em} 
\end{minipage} 
\hfill
\end{table*}

\begin{table*}
\caption{Experiments in \emph{base-to-novel} generalization with the ViT-B/32 visual backbone $\mathbf{k{=}4}$ shots per base class.}
\label{tab:b2n_vitb32_shots4}
\def\arraystretch{1.075}
\centering
\scriptsize
\begin{minipage}{0.32\textwidth}
\centering 
\caption*{\textbf{Average across datasets.}}\vspace{-1em} 
\begin{tabular}{lcc|c} 
\toprule 
\textbf{Method} & \textbf{Base} & \textbf{Novel} & \textbf{HM} \\ 
\midrule 
CLIP \cite{radford2021learning} & 67.27 & 71.68 & 69.41 \\ 
MMA \cite{yang2024mma} & 77.20 & 74.43 & 70.55 \\ 
\cmidrule(lr){1-4} 
\rowcolor{lncolor}
\ours{} & 82.04 & 74.44 & \textbf{75.00} \\ 
\bottomrule 
\end{tabular} 
\vspace{0.5em} 
\end{minipage} 
\hfill
\begin{minipage}{0.32\textwidth}
\centering 
\caption*{ImageNet}\vspace{-1em} 
\begin{tabular}{lcc|c} 
\toprule 
\textbf{Method} & \textbf{Base} & \textbf{Novel} & \textbf{HM} \\ 
\midrule 
CLIP \cite{radford2021learning} & 67.49 & 64.06 & 65.73 \\ 
MMA \cite{yang2024mma} & 69.77 & 65.63 & 67.64 \\ 
\cmidrule(lr){1-4} 
\rowcolor{lncolor}
\ours{} & 70.51 & 65.91 & \textbf{68.13} \\ 
\bottomrule 
\end{tabular} 
\vspace{0.5em} 
\end{minipage} 
\hfill
\begin{minipage}{0.32\textwidth}
\centering 
\caption*{Caltech101}\vspace{-1em} 
\begin{tabular}{lcc|c} 
\toprule 
\textbf{Method} & \textbf{Base} & \textbf{Novel} & \textbf{HM} \\ 
\midrule 
CLIP \cite{radford2021learning} & 94.06 & 94.00 & 94.03 \\ 
MMA \cite{yang2024mma} & 96.90 & 93.03 & 94.93 \\ 
\cmidrule(lr){1-4} 
\rowcolor{lncolor}
\ours{} & 97.16 & 93.52 & \textbf{95.31} \\ 
\bottomrule 
\end{tabular} 
\vspace{0.5em} 
\end{minipage} 
\hfill
\begin{minipage}{0.32\textwidth}
\centering 
\caption*{Oxford Flowers}\vspace{-1em} 
\begin{tabular}{lcc|c} 
\toprule 
\textbf{Method} & \textbf{Base} & \textbf{Novel} & \textbf{HM} \\ 
\midrule 
CLIP \cite{radford2021learning} & 72.36 & 73.69 & 73.02 \\ 
MMA \cite{yang2024mma} & 87.03 & 70.83 & 78.10 \\ 
\cmidrule(lr){1-4} 
\rowcolor{lncolor}
\ours{} & 93.99 & 71.70 & \textbf{81.35} \\ 
\bottomrule 
\end{tabular} 
\vspace{0.5em} 
\end{minipage} 
\hfill
\begin{minipage}{0.32\textwidth}
\centering 
\caption*{Oxford Pets}\vspace{-1em} 
\begin{tabular}{lcc|c} 
\toprule 
\textbf{Method} & \textbf{Base} & \textbf{Novel} & \textbf{HM} \\ 
\midrule 
CLIP \cite{radford2021learning} & 90.64 & 96.87 & 93.65 \\ 
MMA \cite{yang2024mma} & 88.43 & 96.33 & 92.21 \\ 
\cmidrule(lr){1-4} 
\rowcolor{lncolor}
\ours{} & 92.50 & 95.25 & \textbf{93.86} \\ 
\bottomrule 
\end{tabular} 
\vspace{0.5em} 
\end{minipage} 
\hfill
\begin{minipage}{0.32\textwidth}
\centering 
\caption*{Stanford Cars}\vspace{-1em} 
\begin{tabular}{lcc|c} 
\toprule 
\textbf{Method} & \textbf{Base} & \textbf{Novel} & \textbf{HM} \\ 
\midrule 
CLIP \cite{radford2021learning} & 60.72 & 69.74 & 64.92 \\ 
MMA \cite{yang2024mma} & 66.23 & 69.17 & 67.67 \\ 
\cmidrule(lr){1-4} 
\rowcolor{lncolor}
\ours{} & 69.22 & 70.84 & \textbf{70.02} \\ 
\bottomrule 
\end{tabular} 
\vspace{0.5em} 
\end{minipage} 
\hfill
\begin{minipage}{0.32\textwidth}
\centering 
\caption*{Food101}\vspace{-1em} 
\begin{tabular}{lcc|c} 
\toprule 
\textbf{Method} & \textbf{Base} & \textbf{Novel} & \textbf{HM} \\ 
\midrule 
CLIP \cite{radford2021learning} & 85.30 & 86.89 & \textbf{86.09} \\ 
MMA \cite{yang2024mma} & 85.03 & 86.40 & 85.71 \\ 
\cmidrule(lr){1-4} 
\rowcolor{lncolor}
\ours{} & 84.04 & 86.95 & 85.47 \\ 
\bottomrule 
\end{tabular} 
\vspace{0.5em} 
\end{minipage} 
\hfill
\begin{minipage}{0.32\textwidth}
\centering 
\caption*{FGVC Aircraft}\vspace{-1em} 
\begin{tabular}{lcc|c} 
\toprule 
\textbf{Method} & \textbf{Base} & \textbf{Novel} & \textbf{HM} \\ 
\midrule 
CLIP \cite{radford2021learning} & 21.25 & 29.27 & 24.62 \\ 
MMA \cite{yang2024mma} & 25.27 & 27.90 & 26.52 \\ 
\cmidrule(lr){1-4} 
\rowcolor{lncolor}
\ours{} & 33.11 & 30.23 & \textbf{31.61} \\ 
\bottomrule 
\end{tabular} 
\vspace{0.5em} 
\end{minipage} 
\hfill
\begin{minipage}{0.32\textwidth}
\centering 
\caption*{SUN397}\vspace{-1em} 
\begin{tabular}{lcc|c} 
\toprule 
\textbf{Method} & \textbf{Base} & \textbf{Novel} & \textbf{HM} \\ 
\midrule 
CLIP \cite{radford2021learning} & 69.80 & 73.01 & 71.37 \\ 
MMA \cite{yang2024mma} & 77.37 & 76.40 & 76.88 \\ 
\cmidrule(lr){1-4} 
\rowcolor{lncolor}
\ours{} & 78.23 & 76.34 & \textbf{77.27} \\ 
\bottomrule 
\end{tabular} 
\vspace{0.5em} 
\end{minipage} 
\hfill
\begin{minipage}{0.32\textwidth}
\centering 
\caption*{DTD}\vspace{-1em} 
\begin{tabular}{lcc|c} 
\toprule 
\textbf{Method} & \textbf{Base} & \textbf{Novel} & \textbf{HM} \\ 
\midrule 
CLIP \cite{radford2021learning} & 54.17 & 58.21 & 56.12 \\ 
MMA \cite{yang2024mma} & 62.13 & 57.70 & 59.83 \\ 
\cmidrule(lr){1-4} 
\rowcolor{lncolor}
\ours{} & 74.23 & 56.76 & \textbf{64.33} \\ 
\bottomrule 
\end{tabular} 
\vspace{0.5em} 
\end{minipage} 
\hfill
\begin{minipage}{0.32\textwidth}
\centering 
\caption*{EuroSAT}\vspace{-1em} 
\begin{tabular}{lcc|c} 
\toprule 
\textbf{Method} & \textbf{Base} & \textbf{Novel} & \textbf{HM} \\ 
\midrule 
CLIP \cite{radford2021learning} & 55.14 & 69.77 & 61.60 \\ 
MMA \cite{yang2024mma} & 41.73 & 57.17 & 48.24 \\ 
\cmidrule(lr){1-4} 
\rowcolor{lncolor}
\ours{} & 87.91 & 68.35 & \textbf{76.91} \\ 
\bottomrule 
\end{tabular} 
\vspace{0.5em} 
\end{minipage} 
\hfill
\begin{minipage}{0.32\textwidth}
\centering 
\caption*{UCF101}\vspace{-1em} 
\begin{tabular}{lcc|c} 
\toprule 
\textbf{Method} & \textbf{Base} & \textbf{Novel} & \textbf{HM} \\ 
\midrule 
CLIP \cite{radford2021learning} & 69.08 & 72.96 & 70.97 \\ 
MMA \cite{yang2024mma} & 77.20 & 74.43 & 75.79 \\ 
\cmidrule(lr){1-4} 
\rowcolor{lncolor}
\ours{} & 82.04 & 74.44 & \textbf{78.05} \\ 
\bottomrule 
\end{tabular} 
\vspace{0.5em} 
\end{minipage} 
\hfill
\end{table*}

\begin{table*}
\caption{Experiments in \emph{base-to-novel} generalization with the ViT-B/32 visual backbone $\mathbf{k{=}8}$ shots per base class.}
\label{tab:b2n_vitb32_shots8}
\def\arraystretch{1.075}
\centering
\scriptsize
\begin{minipage}{0.32\textwidth}
\centering 
\caption*{\textbf{Average across datasets.}}\vspace{-1em} 
\begin{tabular}{lcc|c} 
\toprule 
\textbf{Method} & \textbf{Base} & \textbf{Novel} & \textbf{HM} \\ 
\midrule 
CLIP \cite{radford2021learning} & 67.27 & 71.68 & 69.41 \\ 
MMA \cite{yang2024mma} & 81.67 & 73.83 & 72.06 \\ 
\cmidrule(lr){1-4} 
\rowcolor{lncolor}
\ours{} & 84.21 & 74.62 & \textbf{75.88} \\ 
\bottomrule 
\end{tabular} 
\vspace{0.5em} 
\end{minipage} 
\hfill
\begin{minipage}{0.32\textwidth}
\centering 
\caption*{ImageNet}\vspace{-1em} 
\begin{tabular}{lcc|c} 
\toprule 
\textbf{Method} & \textbf{Base} & \textbf{Novel} & \textbf{HM} \\ 
\midrule 
CLIP \cite{radford2021learning} & 67.49 & 64.06 & 65.73 \\ 
MMA \cite{yang2024mma} & 71.03 & 65.17 & 67.97 \\ 
\cmidrule(lr){1-4} 
\rowcolor{lncolor}
\ours{} & 71.39 & 66.24 & \textbf{68.72} \\ 
\bottomrule 
\end{tabular} 
\vspace{0.5em} 
\end{minipage} 
\hfill
\begin{minipage}{0.32\textwidth}
\centering 
\caption*{Caltech101}\vspace{-1em} 
\begin{tabular}{lcc|c} 
\toprule 
\textbf{Method} & \textbf{Base} & \textbf{Novel} & \textbf{HM} \\ 
\midrule 
CLIP \cite{radford2021learning} & 94.06 & 94.00 & 94.03 \\ 
MMA \cite{yang2024mma} & 97.13 & 92.57 & 94.80 \\ 
\cmidrule(lr){1-4} 
\rowcolor{lncolor}
\ours{} & 97.61 & 93.56 & \textbf{95.54} \\ 
\bottomrule 
\end{tabular} 
\vspace{0.5em} 
\end{minipage} 
\hfill
\begin{minipage}{0.32\textwidth}
\centering 
\caption*{Oxford Flowers}\vspace{-1em} 
\begin{tabular}{lcc|c} 
\toprule 
\textbf{Method} & \textbf{Base} & \textbf{Novel} & \textbf{HM} \\ 
\midrule 
CLIP \cite{radford2021learning} & 72.36 & 73.69 & 73.02 \\ 
MMA \cite{yang2024mma} & 92.97 & 71.63 & 80.92 \\ 
\cmidrule(lr){1-4} 
\rowcolor{lncolor}
\ours{} & 95.79 & 71.35 & \textbf{81.78} \\ 
\bottomrule 
\end{tabular} 
\vspace{0.5em} 
\end{minipage} 
\hfill
\begin{minipage}{0.32\textwidth}
\centering 
\caption*{Oxford Pets}\vspace{-1em} 
\begin{tabular}{lcc|c} 
\toprule 
\textbf{Method} & \textbf{Base} & \textbf{Novel} & \textbf{HM} \\ 
\midrule 
CLIP \cite{radford2021learning} & 90.64 & 96.87 & 93.65 \\ 
MMA \cite{yang2024mma} & 93.57 & 95.57 & \textbf{94.56} \\ 
\cmidrule(lr){1-4} 
\rowcolor{lncolor}
\ours{} & 92.79 & 95.58 & 94.16 \\ 
\bottomrule 
\end{tabular} 
\vspace{0.5em} 
\end{minipage} 
\hfill
\begin{minipage}{0.32\textwidth}
\centering 
\caption*{Stanford Cars}\vspace{-1em} 
\begin{tabular}{lcc|c} 
\toprule 
\textbf{Method} & \textbf{Base} & \textbf{Novel} & \textbf{HM} \\ 
\midrule 
CLIP \cite{radford2021learning} & 60.72 & 69.74 & 64.92 \\ 
MMA \cite{yang2024mma} & 69.83 & 69.80 & 69.81 \\ 
\cmidrule(lr){1-4} 
\rowcolor{lncolor}
\ours{} & 73.48 & 70.71 & \textbf{72.07} \\ 
\bottomrule 
\end{tabular} 
\vspace{0.5em} 
\end{minipage} 
\hfill
\begin{minipage}{0.32\textwidth}
\centering 
\caption*{Food101}\vspace{-1em} 
\begin{tabular}{lcc|c} 
\toprule 
\textbf{Method} & \textbf{Base} & \textbf{Novel} & \textbf{HM} \\ 
\midrule 
CLIP \cite{radford2021learning} & 85.30 & 86.89 & \textbf{86.09} \\ 
MMA \cite{yang2024mma} & 85.03 & 86.53 & 85.77 \\ 
\cmidrule(lr){1-4} 
\rowcolor{lncolor}
\ours{} & 84.15 & 87.33 & 85.71 \\ 
\bottomrule 
\end{tabular} 
\vspace{0.5em} 
\end{minipage} 
\hfill
\begin{minipage}{0.32\textwidth}
\centering 
\caption*{FGVC Aircraft}\vspace{-1em} 
\begin{tabular}{lcc|c} 
\toprule 
\textbf{Method} & \textbf{Base} & \textbf{Novel} & \textbf{HM} \\ 
\midrule 
CLIP \cite{radford2021learning} & 21.25 & 29.27 & 24.62 \\ 
MMA \cite{yang2024mma} & 28.63 & 27.67 & 28.14 \\ 
\cmidrule(lr){1-4} 
\rowcolor{lncolor}
\ours{} & 35.47 & 30.35 & \textbf{32.71} \\ 
\bottomrule 
\end{tabular} 
\vspace{0.5em} 
\end{minipage} 
\hfill
\begin{minipage}{0.32\textwidth}
\centering 
\caption*{SUN397}\vspace{-1em} 
\begin{tabular}{lcc|c} 
\toprule 
\textbf{Method} & \textbf{Base} & \textbf{Novel} & \textbf{HM} \\ 
\midrule 
CLIP \cite{radford2021learning} & 69.80 & 73.01 & 71.37 \\ 
MMA \cite{yang2024mma} & 78.83 & 76.10 & 77.44 \\ 
\cmidrule(lr){1-4} 
\rowcolor{lncolor}
\ours{} & 79.49 & 77.13 & \textbf{78.29} \\ 
\bottomrule 
\end{tabular} 
\vspace{0.5em} 
\end{minipage} 
\hfill
\begin{minipage}{0.32\textwidth}
\centering 
\caption*{DTD}\vspace{-1em} 
\begin{tabular}{lcc|c} 
\toprule 
\textbf{Method} & \textbf{Base} & \textbf{Novel} & \textbf{HM} \\ 
\midrule 
CLIP \cite{radford2021learning} & 54.17 & 58.21 & 56.12 \\ 
MMA \cite{yang2024mma} & 73.70 & 56.07 & 63.69 \\ 
\cmidrule(lr){1-4} 
\rowcolor{lncolor}
\ours{} & 75.85 & 55.23 & \textbf{63.92} \\ 
\bottomrule 
\end{tabular} 
\vspace{0.5em} 
\end{minipage} 
\hfill
\begin{minipage}{0.32\textwidth}
\centering 
\caption*{EuroSAT}\vspace{-1em} 
\begin{tabular}{lcc|c} 
\toprule 
\textbf{Method} & \textbf{Base} & \textbf{Novel} & \textbf{HM} \\ 
\midrule 
CLIP \cite{radford2021learning} & 55.14 & 69.77 & 61.60 \\ 
MMA \cite{yang2024mma} & 41.80 & 57.20 & 48.30 \\ 
\cmidrule(lr){1-4} 
\rowcolor{lncolor}
\ours{} & 94.18 & 68.24 & \textbf{79.14} \\ 
\bottomrule 
\end{tabular} 
\vspace{0.5em} 
\end{minipage} 
\hfill
\begin{minipage}{0.32\textwidth}
\centering 
\caption*{UCF101}\vspace{-1em} 
\begin{tabular}{lcc|c} 
\toprule 
\textbf{Method} & \textbf{Base} & \textbf{Novel} & \textbf{HM} \\ 
\midrule 
CLIP \cite{radford2021learning} & 69.08 & 72.96 & 70.97 \\ 
MMA \cite{yang2024mma} & 81.67 & 73.83 & 77.55 \\ 
\cmidrule(lr){1-4} 
\rowcolor{lncolor}
\ours{} & 84.21 & 74.62 & \textbf{79.12} \\ 
\bottomrule 
\end{tabular} 
\vspace{0.5em} 
\end{minipage} 
\hfill
\end{table*}

\begin{table*}
\caption{Experiments in \emph{base-to-novel} generalization with the ViT-L/14 visual backbone $\mathbf{k{=}4}$ shots per base class.}
\label{tab:b2n_vitl14_shots4}
\def\arraystretch{1.075}
\centering
\scriptsize
\begin{minipage}{0.32\textwidth}
\centering 
\caption*{\textbf{Average across datasets.}}\vspace{-1em} 
\begin{tabular}{lcc|c} 
\toprule 
\textbf{Method} & \textbf{Base} & \textbf{Novel} & \textbf{HM} \\ 
\midrule 
CLIP \cite{radford2021learning} & 76.18 & 80.08 & 78.08 \\ 
MMA \cite{yang2024mma} & 82.70 & 81.60 & 80.25 \\ 
\cmidrule(lr){1-4} 
\rowcolor{lncolor}
\ours{} & 88.11 & 82.15 & \textbf{82.82} \\ 
\bottomrule 
\end{tabular} 
\vspace{0.5em} 
\end{minipage} 
\hfill
\begin{minipage}{0.32\textwidth}
\centering 
\caption*{ImageNet}\vspace{-1em} 
\begin{tabular}{lcc|c} 
\toprule 
\textbf{Method} & \textbf{Base} & \textbf{Novel} & \textbf{HM} \\ 
\midrule 
CLIP \cite{radford2021learning} & 79.18 & 74.04 & 76.53 \\ 
MMA \cite{yang2024mma} & 82.00 & 76.67 & \textbf{79.25} \\ 
\cmidrule(lr){1-4} 
\rowcolor{lncolor}
\ours{} & 81.35 & 75.90 & 78.53 \\ 
\bottomrule 
\end{tabular} 
\vspace{0.5em} 
\end{minipage} 
\hfill
\begin{minipage}{0.32\textwidth}
\centering 
\caption*{Caltech101}\vspace{-1em} 
\begin{tabular}{lcc|c} 
\toprule 
\textbf{Method} & \textbf{Base} & \textbf{Novel} & \textbf{HM} \\ 
\midrule 
CLIP \cite{radford2021learning} & 95.61 & 95.41 & 95.51 \\ 
MMA \cite{yang2024mma} & 97.30 & 97.30 & 97.30 \\ 
\cmidrule(lr){1-4} 
\rowcolor{lncolor}
\ours{} & 98.36 & 97.09 & \textbf{97.72} \\ 
\bottomrule 
\end{tabular} 
\vspace{0.5em} 
\end{minipage} 
\hfill
\begin{minipage}{0.32\textwidth}
\centering 
\caption*{Oxford Flowers}\vspace{-1em} 
\begin{tabular}{lcc|c} 
\toprule 
\textbf{Method} & \textbf{Base} & \textbf{Novel} & \textbf{HM} \\ 
\midrule 
CLIP \cite{radford2021learning} & 80.34 & 83.05 & 81.67 \\ 
MMA \cite{yang2024mma} & 92.93 & 81.87 & 87.05 \\ 
\cmidrule(lr){1-4} 
\rowcolor{lncolor}
\ours{} & 97.94 & 81.77 & \textbf{89.13} \\ 
\bottomrule 
\end{tabular} 
\vspace{0.5em} 
\end{minipage} 
\hfill
\begin{minipage}{0.32\textwidth}
\centering 
\caption*{Oxford Pets}\vspace{-1em} 
\begin{tabular}{lcc|c} 
\toprule 
\textbf{Method} & \textbf{Base} & \textbf{Novel} & \textbf{HM} \\ 
\midrule 
CLIP \cite{radford2021learning} & 93.78 & 96.53 & 95.14 \\ 
MMA \cite{yang2024mma} & 94.93 & 98.47 & 96.67 \\ 
\cmidrule(lr){1-4} 
\rowcolor{lncolor}
\ours{} & 96.46 & 98.56 & \textbf{97.50} \\ 
\bottomrule 
\end{tabular} 
\vspace{0.5em} 
\end{minipage} 
\hfill
\begin{minipage}{0.32\textwidth}
\centering 
\caption*{Stanford Cars}\vspace{-1em} 
\begin{tabular}{lcc|c} 
\toprule 
\textbf{Method} & \textbf{Base} & \textbf{Novel} & \textbf{HM} \\ 
\midrule 
CLIP \cite{radford2021learning} & 74.56 & 84.65 & 79.29 \\ 
MMA \cite{yang2024mma} & 79.83 & 85.03 & 82.35 \\ 
\cmidrule(lr){1-4} 
\rowcolor{lncolor}
\ours{} & 82.50 & 85.11 & \textbf{83.79} \\ 
\bottomrule 
\end{tabular} 
\vspace{0.5em} 
\end{minipage} 
\hfill
\begin{minipage}{0.32\textwidth}
\centering 
\caption*{Food101}\vspace{-1em} 
\begin{tabular}{lcc|c} 
\toprule 
\textbf{Method} & \textbf{Base} & \textbf{Novel} & \textbf{HM} \\ 
\midrule 
CLIP \cite{radford2021learning} & 93.75 & 94.82 & \textbf{94.28} \\ 
MMA \cite{yang2024mma} & 93.70 & 94.57 & 94.13 \\ 
\cmidrule(lr){1-4} 
\rowcolor{lncolor}
\ours{} & 93.11 & 94.76 & 93.93 \\ 
\bottomrule 
\end{tabular} 
\vspace{0.5em} 
\end{minipage} 
\hfill
\begin{minipage}{0.32\textwidth}
\centering 
\caption*{FGVC Aircraft}\vspace{-1em} 
\begin{tabular}{lcc|c} 
\toprule 
\textbf{Method} & \textbf{Base} & \textbf{Novel} & \textbf{HM} \\ 
\midrule 
CLIP \cite{radford2021learning} & 37.52 & 44.21 & 40.59 \\ 
MMA \cite{yang2024mma} & 42.57 & 42.40 & 42.48 \\ 
\cmidrule(lr){1-4} 
\rowcolor{lncolor}
\ours{} & 51.58 & 44.57 & \textbf{47.82} \\ 
\bottomrule 
\end{tabular} 
\vspace{0.5em} 
\end{minipage} 
\hfill
\begin{minipage}{0.32\textwidth}
\centering 
\caption*{SUN397}\vspace{-1em} 
\begin{tabular}{lcc|c} 
\toprule 
\textbf{Method} & \textbf{Base} & \textbf{Novel} & \textbf{HM} \\ 
\midrule 
CLIP \cite{radford2021learning} & 73.23 & 77.71 & 75.40 \\ 
MMA \cite{yang2024mma} & 82.17 & 81.80 & 81.98 \\ 
\cmidrule(lr){1-4} 
\rowcolor{lncolor}
\ours{} & 82.91 & 81.20 & \textbf{82.05} \\ 
\bottomrule 
\end{tabular} 
\vspace{0.5em} 
\end{minipage} 
\hfill
\begin{minipage}{0.32\textwidth}
\centering 
\caption*{DTD}\vspace{-1em} 
\begin{tabular}{lcc|c} 
\toprule 
\textbf{Method} & \textbf{Base} & \textbf{Novel} & \textbf{HM} \\ 
\midrule 
CLIP \cite{radford2021learning} & 59.14 & 67.87 & 63.21 \\ 
MMA \cite{yang2024mma} & 65.90 & 67.00 & 66.45 \\ 
\cmidrule(lr){1-4} 
\rowcolor{lncolor}
\ours{} & 80.86 & 70.29 & \textbf{75.21} \\ 
\bottomrule 
\end{tabular} 
\vspace{0.5em} 
\end{minipage} 
\hfill
\begin{minipage}{0.32\textwidth}
\centering 
\caption*{EuroSAT}\vspace{-1em} 
\begin{tabular}{lcc|c} 
\toprule 
\textbf{Method} & \textbf{Base} & \textbf{Novel} & \textbf{HM} \\ 
\midrule 
CLIP \cite{radford2021learning} & 70.93 & 82.90 & 76.45 \\ 
MMA \cite{yang2024mma} & 72.50 & 72.20 & 72.35 \\ 
\cmidrule(lr){1-4} 
\rowcolor{lncolor}
\ours{} & 92.92 & 67.15 & \textbf{77.96} \\ 
\bottomrule 
\end{tabular} 
\vspace{0.5em} 
\end{minipage} 
\hfill
\begin{minipage}{0.32\textwidth}
\centering 
\caption*{UCF101}\vspace{-1em} 
\begin{tabular}{lcc|c} 
\toprule 
\textbf{Method} & \textbf{Base} & \textbf{Novel} & \textbf{HM} \\ 
\midrule 
CLIP \cite{radford2021learning} & 79.94 & 79.66 & 79.80 \\ 
MMA \cite{yang2024mma} & 82.70 & 81.60 & 82.15 \\ 
\cmidrule(lr){1-4} 
\rowcolor{lncolor}
\ours{} & 88.11 & 82.15 & \textbf{85.03} \\ 
\bottomrule 
\end{tabular} 
\vspace{0.5em} 
\end{minipage} 
\hfill
\end{table*}

\begin{table*}
\caption{Experiments in \emph{base-to-novel} generalization with the ViT-L/14 visual backbone $\mathbf{k{=}8}$ shots per base class.}
\label{tab:b2n_vitl14_shots8}
\def\arraystretch{1.075}
\centering
\scriptsize
\begin{minipage}{0.32\textwidth}
\centering 
\caption*{\textbf{Average across datasets.}}\vspace{-1em} 
\begin{tabular}{lcc|c} 
\toprule 
\textbf{Method} & \textbf{Base} & \textbf{Novel} & \textbf{HM} \\ 
\midrule 
CLIP \cite{radford2021learning} & 76.18 & 80.08 & 78.08 \\ 
MMA \cite{yang2024mma} & 86.30 & 80.73 & 81.54 \\ 
\cmidrule(lr){1-4} 
\rowcolor{lncolor}
\ours{} & 88.28 & 82.24 & \textbf{83.66} \\ 
\bottomrule 
\end{tabular} 
\vspace{0.5em} 
\end{minipage} 
\hfill
\begin{minipage}{0.32\textwidth}
\centering 
\caption*{ImageNet}\vspace{-1em} 
\begin{tabular}{lcc|c} 
\toprule 
\textbf{Method} & \textbf{Base} & \textbf{Novel} & \textbf{HM} \\ 
\midrule 
CLIP \cite{radford2021learning} & 79.18 & 74.04 & 76.53 \\ 
MMA \cite{yang2024mma} & 82.63 & 76.80 & \textbf{79.61} \\ 
\cmidrule(lr){1-4} 
\rowcolor{lncolor}
\ours{} & 82.43 & 76.46 & 79.34 \\ 
\bottomrule 
\end{tabular} 
\vspace{0.5em} 
\end{minipage} 
\hfill
\begin{minipage}{0.32\textwidth}
\centering 
\caption*{Caltech101}\vspace{-1em} 
\begin{tabular}{lcc|c} 
\toprule 
\textbf{Method} & \textbf{Base} & \textbf{Novel} & \textbf{HM} \\ 
\midrule 
CLIP \cite{radford2021learning} & 95.61 & 95.41 & 95.51 \\ 
MMA \cite{yang2024mma} & 98.30 & 96.60 & 97.44 \\ 
\cmidrule(lr){1-4} 
\rowcolor{lncolor}
\ours{} & 98.52 & 96.62 & \textbf{97.56} \\ 
\bottomrule 
\end{tabular} 
\vspace{0.5em} 
\end{minipage} 
\hfill
\begin{minipage}{0.32\textwidth}
\centering 
\caption*{Oxford Flowers}\vspace{-1em} 
\begin{tabular}{lcc|c} 
\toprule 
\textbf{Method} & \textbf{Base} & \textbf{Novel} & \textbf{HM} \\ 
\midrule 
CLIP \cite{radford2021learning} & 80.34 & 83.05 & 81.67 \\ 
MMA \cite{yang2024mma} & 97.97 & 80.30 & 88.26 \\ 
\cmidrule(lr){1-4} 
\rowcolor{lncolor}
\ours{} & 98.67 & 81.21 & \textbf{89.09} \\ 
\bottomrule 
\end{tabular} 
\vspace{0.5em} 
\end{minipage} 
\hfill
\begin{minipage}{0.32\textwidth}
\centering 
\caption*{Oxford Pets}\vspace{-1em} 
\begin{tabular}{lcc|c} 
\toprule 
\textbf{Method} & \textbf{Base} & \textbf{Novel} & \textbf{HM} \\ 
\midrule 
CLIP \cite{radford2021learning} & 93.78 & 96.53 & 95.14 \\ 
MMA \cite{yang2024mma} & 95.77 & 98.33 & 97.03 \\ 
\cmidrule(lr){1-4} 
\rowcolor{lncolor}
\ours{} & 96.15 & 98.47 & \textbf{97.30} \\ 
\bottomrule 
\end{tabular} 
\vspace{0.5em} 
\end{minipage} 
\hfill
\begin{minipage}{0.32\textwidth}
\centering 
\caption*{Stanford Cars}\vspace{-1em} 
\begin{tabular}{lcc|c} 
\toprule 
\textbf{Method} & \textbf{Base} & \textbf{Novel} & \textbf{HM} \\ 
\midrule 
CLIP \cite{radford2021learning} & 74.56 & 84.65 & 79.29 \\ 
MMA \cite{yang2024mma} & 82.63 & 84.20 & 83.41 \\ 
\cmidrule(lr){1-4} 
\rowcolor{lncolor}
\ours{} & 85.51 & 84.97 & \textbf{85.24} \\ 
\bottomrule 
\end{tabular} 
\vspace{0.5em} 
\end{minipage} 
\hfill
\begin{minipage}{0.32\textwidth}
\centering 
\caption*{Food101}\vspace{-1em} 
\begin{tabular}{lcc|c} 
\toprule 
\textbf{Method} & \textbf{Base} & \textbf{Novel} & \textbf{HM} \\ 
\midrule 
CLIP \cite{radford2021learning} & 93.75 & 94.82 & 94.28 \\ 
MMA \cite{yang2024mma} & 93.87 & 94.87 & \textbf{94.37} \\ 
\cmidrule(lr){1-4} 
\rowcolor{lncolor}
\ours{} & 93.81 & 94.76 & 94.28 \\ 
\bottomrule 
\end{tabular} 
\vspace{0.5em} 
\end{minipage} 
\hfill
\begin{minipage}{0.32\textwidth}
\centering 
\caption*{FGVC Aircraft}\vspace{-1em} 
\begin{tabular}{lcc|c} 
\toprule 
\textbf{Method} & \textbf{Base} & \textbf{Novel} & \textbf{HM} \\ 
\midrule 
CLIP \cite{radford2021learning} & 37.52 & 44.21 & 40.59 \\ 
MMA \cite{yang2024mma} & 46.50 & 41.23 & 43.71 \\ 
\cmidrule(lr){1-4} 
\rowcolor{lncolor}
\ours{} & 55.00 & 44.49 & \textbf{49.19} \\ 
\bottomrule 
\end{tabular} 
\vspace{0.5em} 
\end{minipage} 
\hfill
\begin{minipage}{0.32\textwidth}
\centering 
\caption*{SUN397}\vspace{-1em} 
\begin{tabular}{lcc|c} 
\toprule 
\textbf{Method} & \textbf{Base} & \textbf{Novel} & \textbf{HM} \\ 
\midrule 
CLIP \cite{radford2021learning} & 73.23 & 77.71 & 75.40 \\ 
MMA \cite{yang2024mma} & 83.67 & 81.40 & 82.52 \\ 
\cmidrule(lr){1-4} 
\rowcolor{lncolor}
\ours{} & 84.25 & 81.84 & \textbf{83.03} \\ 
\bottomrule 
\end{tabular} 
\vspace{0.5em} 
\end{minipage} 
\hfill
\begin{minipage}{0.32\textwidth}
\centering 
\caption*{DTD}\vspace{-1em} 
\begin{tabular}{lcc|c} 
\toprule 
\textbf{Method} & \textbf{Base} & \textbf{Novel} & \textbf{HM} \\ 
\midrule 
CLIP \cite{radford2021learning} & 59.14 & 67.87 & 63.21 \\ 
MMA \cite{yang2024mma} & 78.13 & 69.90 & 73.79 \\ 
\cmidrule(lr){1-4} 
\rowcolor{lncolor}
\ours{} & 83.91 & 70.01 & \textbf{76.33} \\ 
\bottomrule 
\end{tabular} 
\vspace{0.5em} 
\end{minipage} 
\hfill
\begin{minipage}{0.32\textwidth}
\centering 
\caption*{EuroSAT}\vspace{-1em} 
\begin{tabular}{lcc|c} 
\toprule 
\textbf{Method} & \textbf{Base} & \textbf{Novel} & \textbf{HM} \\ 
\midrule 
CLIP \cite{radford2021learning} & 70.93 & 82.90 & 76.45 \\ 
MMA \cite{yang2024mma} & 72.73 & 72.00 & 72.36 \\ 
\cmidrule(lr){1-4} 
\rowcolor{lncolor}
\ours{} & 94.50 & 71.68 & \textbf{81.53} \\ 
\bottomrule 
\end{tabular} 
\vspace{0.5em} 
\end{minipage} 
\hfill
\begin{minipage}{0.32\textwidth}
\centering 
\caption*{UCF101}\vspace{-1em} 
\begin{tabular}{lcc|c} 
\toprule 
\textbf{Method} & \textbf{Base} & \textbf{Novel} & \textbf{HM} \\ 
\midrule 
CLIP \cite{radford2021learning} & 79.94 & 79.66 & 79.80 \\ 
MMA \cite{yang2024mma} & 86.30 & 80.73 & 83.42 \\ 
\cmidrule(lr){1-4} 
\rowcolor{lncolor}
\ours{} & 88.28 & 82.24 & \textbf{85.15} \\ 
\bottomrule 
\end{tabular} 
\vspace{0.5em} 
\end{minipage} 
\hfill
\end{table*}

\end{document}